\documentclass[accepted]{uai2023} % after acceptance, for a revised
\usepackage[american]{babel}
\usepackage{natbib} % has a nice set of citation styles and commands
\usepackage{mathtools} % amsmath with fixes and additions
\usepackage{booktabs} % commands to create good-looking tables
\usepackage{tikz} % nice language for creating drawings and diagrams

\makeatletter
\newcommand{\biggg}{\bBigg@{1.1}}  
\def\bigggl{\mathopen\biggg}

\makeatother

\usepackage{amsmath}
\usepackage{amssymb}
\usepackage{mathtools}
\usepackage{amsthm}
\usepackage{multirow}
\usepackage{microtype}
\usepackage{graphicx}
\usepackage{subfigure}
\usepackage{booktabs} % for professional tables
\usepackage{color}
\usepackage{hyperref}
\hypersetup{
	colorlinks=true,
	linkcolor=blue,
	citecolor=blue,
	urlcolor=blue
}

\title{Ultra-High-Definition Image Deblurring via Multi-scale Cubic-Mixer}

\author[1]{Xingchi Chen}
\author[2]{Xiuyi Jia}
\author[1]{Zhuoran Zheng$^{*}$}
\affil[1]{%
    School of Cyber Science and Technology\\ Shenzhen Campus\\ Sun Yat-sen University
}
\affil[2]{%
    School of Computer Science and Engineering\\ Nanjing University of Science and Technology
}
\begin{document}
\maketitle

\begin{abstract}
Currently, transformer-based algorithms are making a splash in the domain of image deblurring.
Their achievement depends on the self-attention mechanism with CNN stem to model long range dependencies between tokens.
Unfortunately, this ear-pleasing pipeline introduces high computational complexity and makes it difficult to run an ultra-high-definition image on a single GPU in real time.
To trade-off accuracy and efficiency, the input degraded image is computed cyclically over three dimensional ($C$, $W$, and $H$) signals without a self-attention mechanism.
We term this deep network as Multi-scale Cubic-Mixer, which is acted on both the real and imaginary components after fast Fourier transform to estimate the Fourier coefficients and thus obtain a deblurred image. 
Furthermore, we combine the multi-scale cubic-mixer with a slicing strategy to generate high-quality results at a much lower computational cost. 
Experimental results demonstrate that the proposed algorithm performs favorably against the state-of-the-art deblurring approaches on the several benchmarks and a new ultra-high-definition dataset in terms of accuracy and speed.
The URL address for the code is at \url{https://github.com/zzr-idam/deblur}.
%The harsh challenges of full-screen devices are position a camera behind a screen, especially for UAVs.
 
\end{abstract}

\section{Introduction}\label{sec:intro}
%This work focuses on the challenge of motion blur in a single ultra-high-definition image.
%	Many mobile device manufacturers have released new devices (e.g., smartphones and tablets) with 4K support.
%However, motion blur is often seen in photos taken with hand-held mobile devices or in low-frame-rate videos that contain moving objects.
%In particular, UHD images magnify blurred areas compared to HD images or low-resolution images.
Blur is ubiquitous in photography, which degrades the quality of human perception and causes serious disturbances to subsequent computer vision analysis.
To address the problem, traditional methods~\citep  {ChenFWZ19,DongZSW11,KrishnanTF11,LiPLGS018,PerroneF14,SunCWH13,XuZJ13} use statistical constraints for image deblurring.
However, most of these methods cannot generalize to dynamic scenes by using handcrafted priors.
%KrishnanTF11 XuZJ13 SunCWH13
%Chakrabarti16 XuRLJ14 DelbracioS15

Recently, deep learning-based methods have been proposed to solve image deblurring.
Numerous networks and functional units for image deblurring have been developed, including vanilla convoluted pipeline \citep  {gong2017motion,KaufmanF20}, multi-scale networks \citep  {NahKL17,zamir2021multi,ZhangDLK19}, residual learning \citep  {chen2022simple,SuinPR20}, recurrent neural network \citep  {TaoGSWJ18,ZhangPRSBL018}, attention mechanisms \citep  {chen2021pre,liang2021swinir,tu2022maxim,wang2021uformer,Wu0LCW20,zamir2021restormer}, and GAN-based approaches \citep  {KupynBMMM18,KupynMWW19,sui2021mri}. 
%Most existing methods~\citep  {gong2017motion, KupynMWW19, NahKL17, SuinPR20,TaoGSWJ18, ZhangDLK19} recover a sharp image from a blurry input with CNNs in the spatial domain. 
%Although these approaches achieve state-of-the-art results, these networks usually involve many stacked convolutional kernels to capture the global information of heterogeneous blurs in the spatial domain and consume massive computing resources.
%
Although these approaches achieve state-of-the-art results, these networks usually involve many stacked convolutional kernels or deeply aligned self-attention units to capture the global information of heterogeneous blurs in the spatial domain and consume massive computing resources.
%

%\textcolor{red}{Add some description about frequency domain based methods and their shortcomings.}
Another research line~\citep  {Chakrabarti16,DelbracioS15_cvpr,DelbracioS15,qin2021blind,Residual_f} tries to remove blurs based on frequency domain information. These methods obtain Fourier coefficients by fast Fourier transform (FFT) to generate a sharp image with the deep network.
However, there are two limitations to these frequency-based methods.
%
%First, only the real component of the Fourier coefficient is exploited in these networks. Few efforts have been made to bring the imaginary component to image deblurring since the deep network weights are all real numbers and cannot be combined with complex numbers for dot product operations.
%
%Second, 
First, these methods still need to use large/numerous convolutional kernels to establish long-range dependence of Fourier coefficients and usually under-performing.
Second, these methods usually estimate a coefficient tensor acting on the Fourier coefficients to obtain a clear image in the frequency domain, thereby ignoring the real/imaginary parts of the Fourier coefficients in their respective roles.
DeepRFT~\citep{Residual_f} improves upon traditional methods by incorporating a frequency selection mechanism within a transformer-based framework. In contrast, our Multi-scale Cubic-Mixer directly processes the real and imaginary components of Fourier-transformed images using an MLP-based architecture. Furthermore, our slicing scheme optimizes the handling of ultra-high-definition (UHD) images, enabling real-time processing on a single GPU—an aspect not explicitly addressed by DeepRFT.
%To address this problem,  we propose a MLP-based network named as Siamese-Mixer.
% 描述 Simaese-Mixer 的优势 长依赖和消耗较少的计算资源。

%Reaching a trade-off between accuracy and efficiency of a network is a non-trivial task. 
%
We analyze the difference of the real and imaginary components obtained by FFT between the sharp and blurred images, and find that blurs of an image (i.e., an offset of the image pixels) is closely related to the phase change on the frequency domain. 
We show that combining both the real and imaginary components is necessary for effective image deblurring, especially the change in the real part is more significant.
In addition, similar to our view, Wave-MLP~\citep{tang2021image} also provides theoretical support for our research that the real and imaginary parts of the Fourier coefficients undergo mixed splashing to obtain accurate results in the frequency domain space of the image.
%
%And the arithmetic of the phase is closely related to the real and imaginary parts of the complex matrix obtained by the FFT, where the change in the imaginary part is more significant through our observations.
%
%This explains the reason why our model performs the real and imaginary parts separately, in order to achieve image deblurring more efficiently.
%
\begin{figure*}[t]
	\centering
	\includegraphics[width=0.945\textwidth]{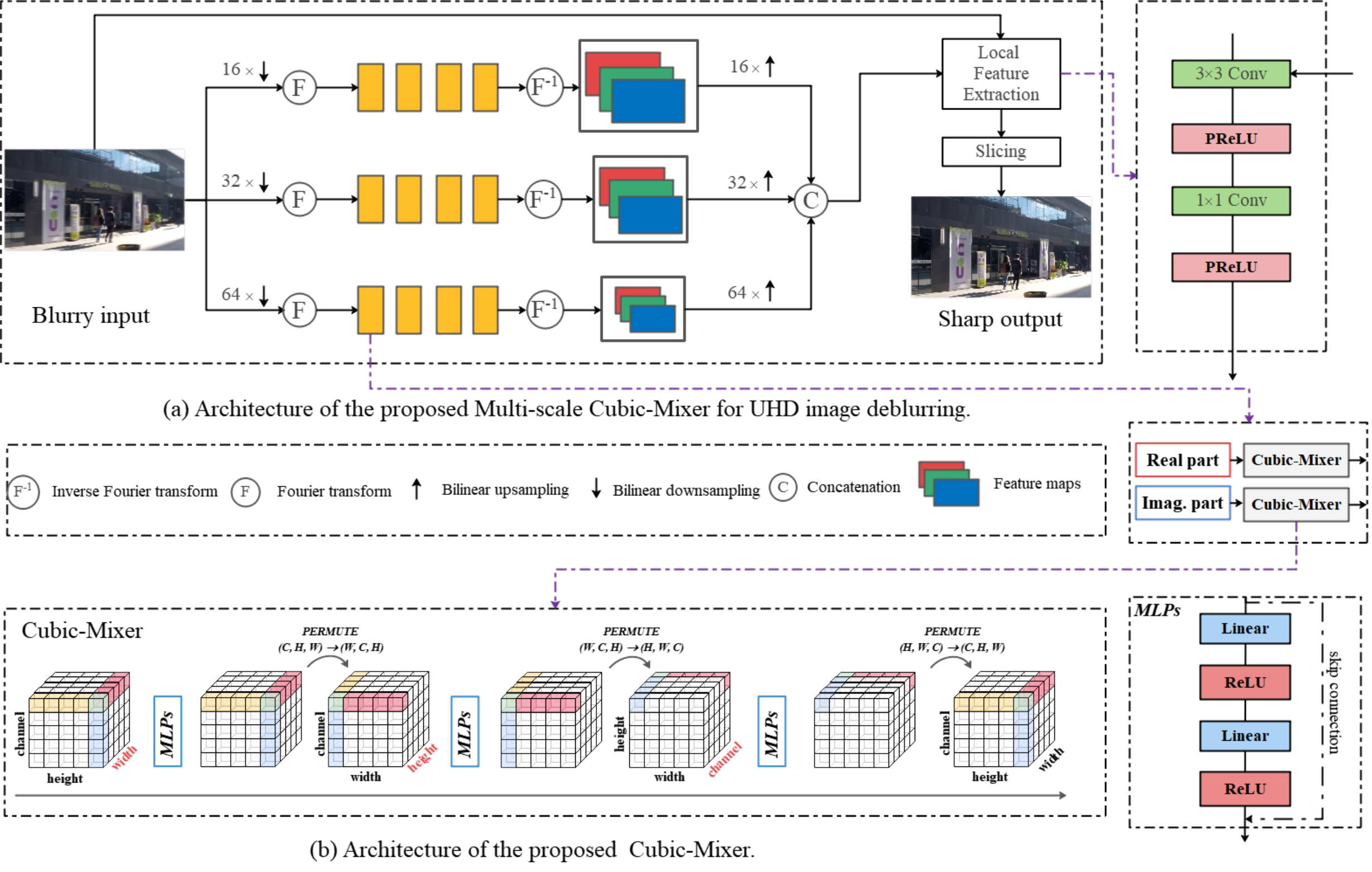}
	\caption{
		(a) shows the architecture of the proposed single image deblurring network, which consists of three parts. 
		The first part starts with three-path low resolution (LR) feature maps prediction stream (multi-scale cubic-mixer) that learns the frequency domain information to predict a basket of full-resolution feature maps. 
		The second part learns a high quality local feature (attention tensor with 6 channels) by using CNNs with PReLU. 
		The last part generates a full-resolution clear image via slicing scheme.
		Our proposed algorithm supports UHD image deblurring at 25 ms on a single Titan RTX GPU shader. 
		(b) shows the architecture of the Cubic-Mixer. 
		The basic framework of the model is proposed in modified MLP-Mixer, and we expand one dimension.
	}
	%\vspace{-1mm}
	\label{frameworks}
\end{figure*}

To mine frequency domain information with high accuracy, we propose a novel network, named multi-scale cubic-mixer, to capture the long-range dependencies and local characteristics in the frequency domain while keeping all the frequency domain information.
%we extracts the Fourier coefficients of the blurred image by FFT, 
%
%and recover high frequency details on the frequency domain.
%
%that acts on the frequency domain of images to by the fully connected network (MLP-Mixer~\citep  {TolstikhinCoRR}) and Fourier convolution which can acquire the long-range dependencies and local relations.
%
%Specifically, our algorithm extracts the Fourier coefficients of the blurred image by FFT.
%
Specifically, the proposed multi-scale cubic-mixer performs on the real and imaginary components of the frequency domain in multiple scales to generate several feature maps (three pairs of real-imaginary feature maps). 
%
%Then, these several feature maps are fused into three new complex tensor by inverse Fourier transform to get a sharp image.
Then, these feature maps are transformed into three new complex tensors by Fourier inverse transform, which are then fused to yield a clear image.
%and the output of the network is upsampled to obtain a ultra-high-resolution image.
%\textcolor{red}{Siamese-Mixer acts as an executor, providing the exact learning object for each pixel and thus obtaining a long range of dependencies. 
	% 这句话没有理论性，前后没有关联，是个executor，怎么就thus 长程依赖了。
	%}
%
%Since a single fully-connected layer in the Siamese-Mixer has fewer floating-point operations than a single convolutional layer (e.g., a $3 \times 3$ kernel) when processing the same image, the proposed model is more computationally efficient than convolutional blocks. 
%
%This is due to a single fully connected layer has less floating point operations than a single convolutional layer (number of convolutional kernels is greater than one) when processing the same resolution image.
%
Note that the basic cell of our algorithm engulfs the complete input information without a CNN stem, and then acts the input information on \emph{MLPs} at three dimensions ($C$, $H$, and $W$).
%
%Siamese-Mixer focuses on reconstructing the over-smoothed areas of the image into high-frequency information because it is able to capture the local relationship between long range dependencies and neighboring pixels for variations in imaginary space.
% 介绍第二个模块。

%However, our model can only handle blurred images below 2K resolution.
%
%Although our Multi-scale Cubic-Mixer is efficient in common image resolution (e.g., 720p and 1080p), it cannot directly handle 4K UHD ($3840 \times 2160$) images. 
%Since UHD \textcolor{red}{text}. % 说明为什么非要处理UHD， 现在的手机等摄影设备都是4K，4K图像视频处理在现阶段显得尤为重要等。
Since many mobile device manufacturers have released new ultra-high-definition (UHD) enabled devices (such as smartphones and tablets), the corresponding methods of enhancing UHD have become extremely important.
% which is a problem of trade-off between accuracy and efficiency in recent years. 
In response, slicing scheme combined with the multi-scale cubic-mixer learning is proposed to deblur arbitrary resolution images. 
%enhance the output sharp image of Siamese-Mixer. 
%
%This enhancement strategy is more efficient than directly filtering the high-resolution images with the help of convolutional blocks. 这是属于上一部分的优点了，在这又重复干啥？
%The high-resolution (HR) images are fused with high quality features and then performed with performs linear slicing to obtain UHD sharp images. 
%
The scheme works on the three color channels of the arbitrary resolution image and enhances the chromaticity and texture details of the arbitrary resolution image by learning a basket of affine transform coefficients.
%
%However, we note that there is a small amount of noise in the transformation coefficients of the bilateral grid. Therefore, we propose a 3D filter to obtain high-quality bilateral grids by eliminating false edges and noises. 
%
Our method takes less than 25 ms to process a 4K resolution ($3840 \times 2160$) image on a single Titan RTX GPU shader with 24G RAM, which is highly efficient for deployment in practical applications.
At the end of this paper, we discuss the limitations, future works, applications and social implications of the model.

The contributions of this paper are summarized as following:
\begin{itemize}
	\item We propose a multi-scale cubic-mixer with wave-frequency processing framework to estimate the Fourier coefficients of the sharp image from the blurry input by learning MLP in three dimensions. 
	Cubic-mixer can focus on long-range information of the image to provide a wide range of receptive fields, while the fusion module can capture the local structure.
	\item We propose a slicing scheme to use the spatial variation and color information between the clear and blurred images to enhance a UHD sharp image since MLP-based approaches cannot directly deal with 4K resolution images.
	\item Experimental results on several synthetic datasets and real-world images demonstrate the proposed algorithm performs favorably against the state-of-the-art image deblurring methods. Our method can run a  4K resolution image at 40 fps.
\end{itemize}

\begin{figure*}[t]\scriptsize
	\begin{center}
		\tabcolsep 1pt
		\begin{tabular}{@{}cccccc@{}}
			\includegraphics[width = 0.16\textwidth]{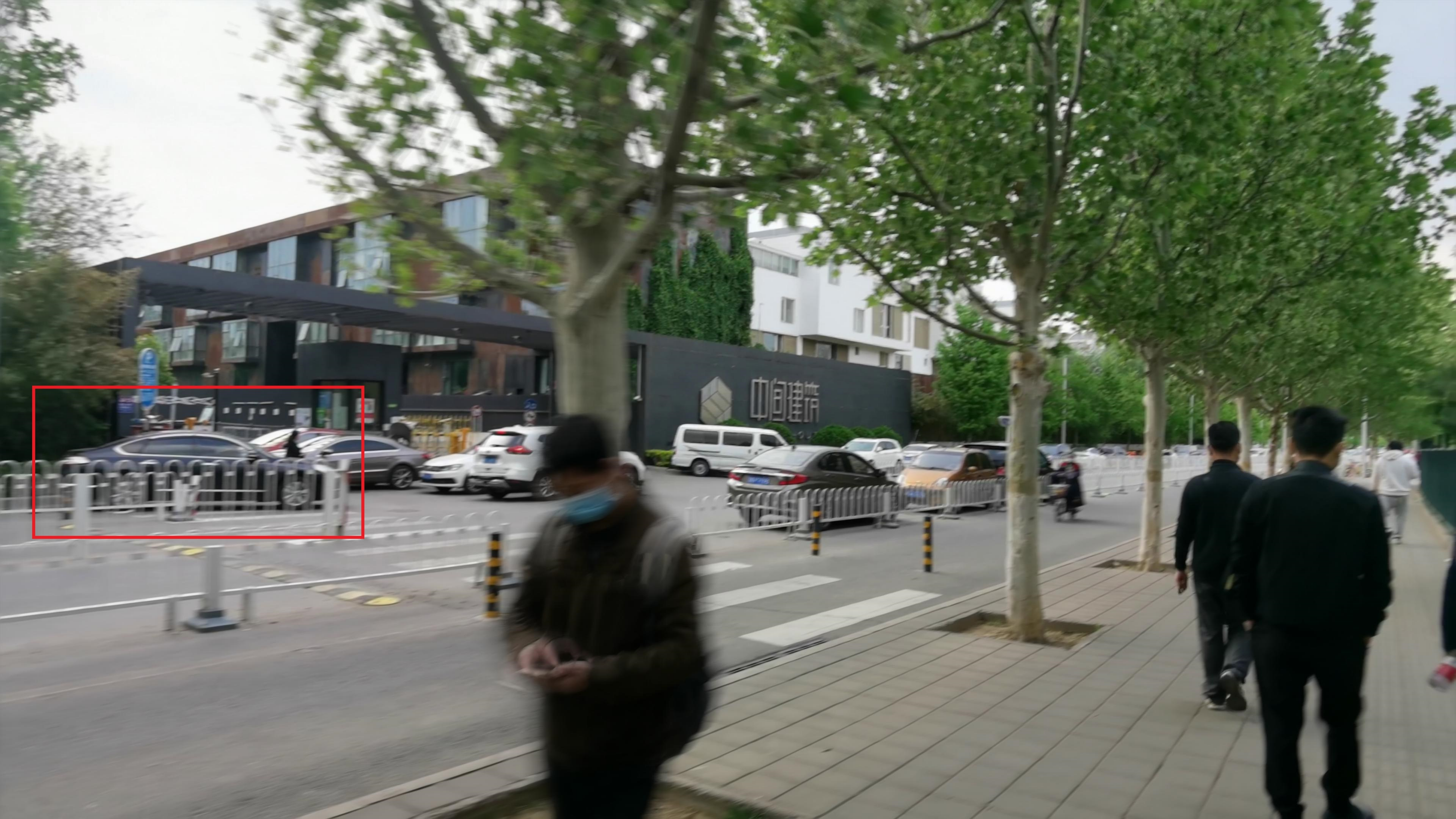}                 &
			\includegraphics[width = 0.16\textwidth]{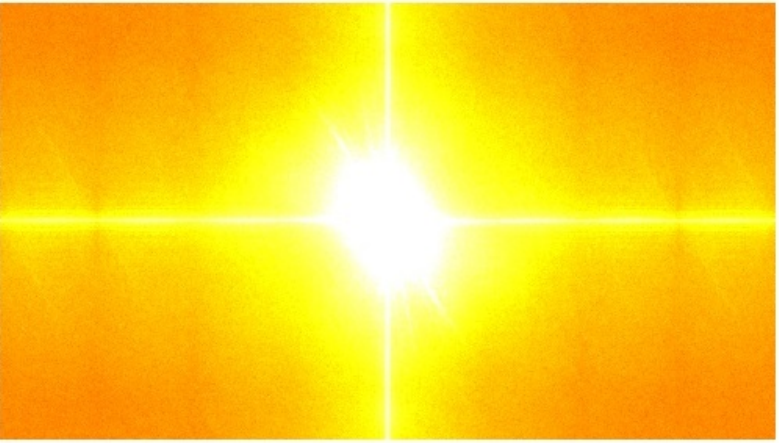}                 &
			\includegraphics[width = 0.16\textwidth]{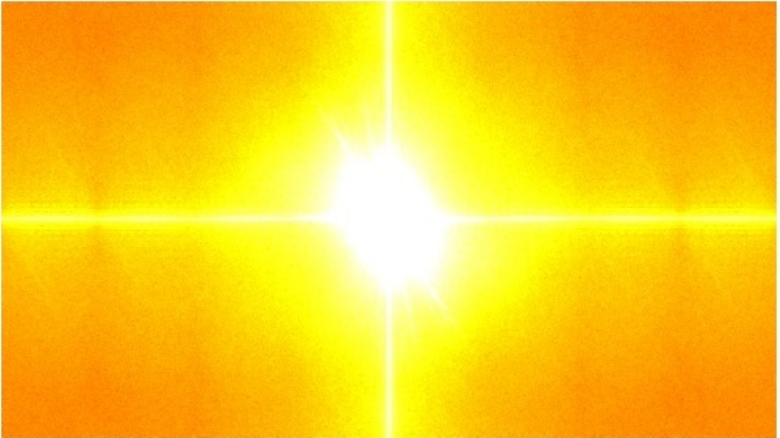}                 &
			\includegraphics[width = 0.16\textwidth]{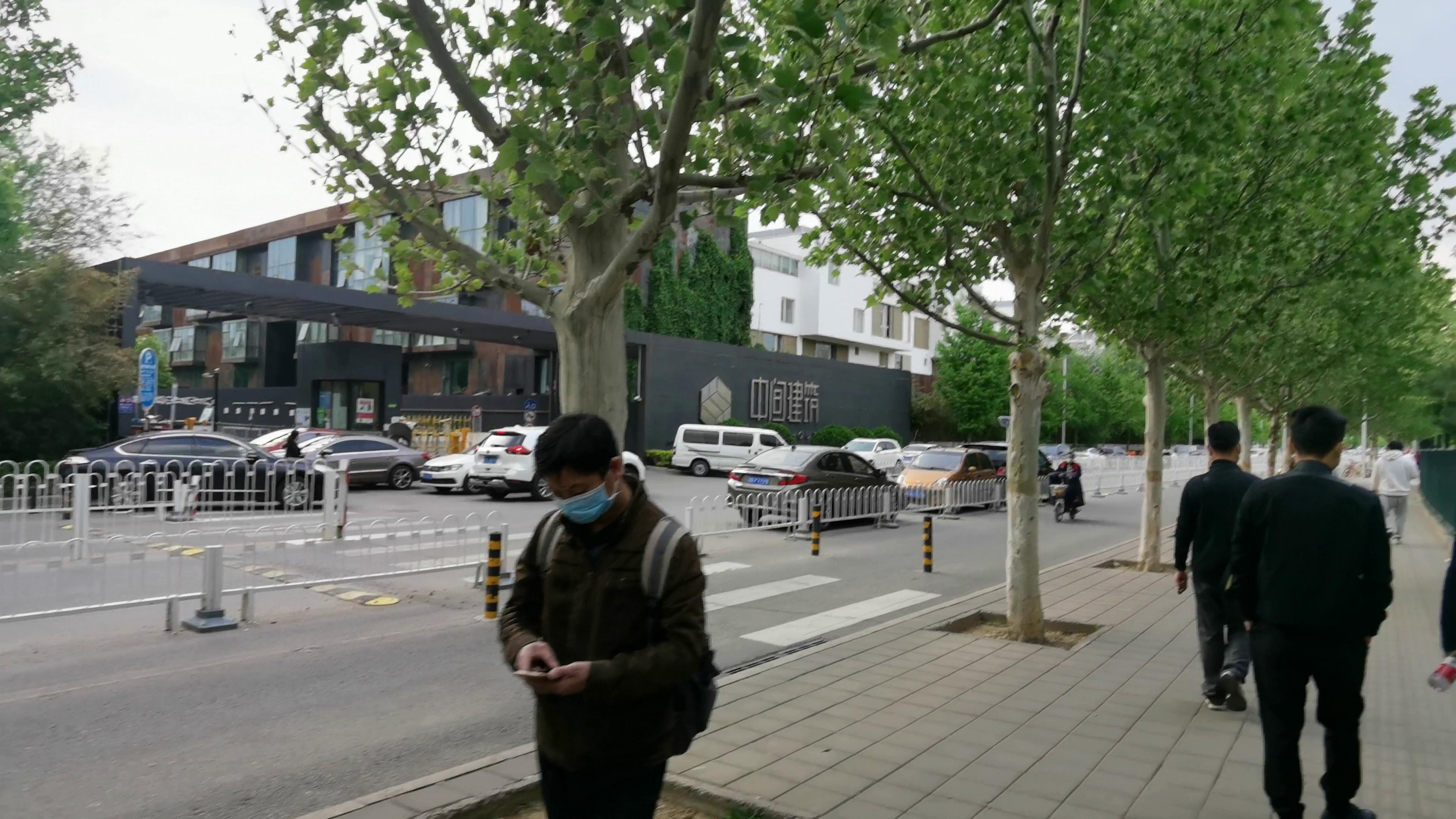}                 &
			\includegraphics[width = 0.16\textwidth]{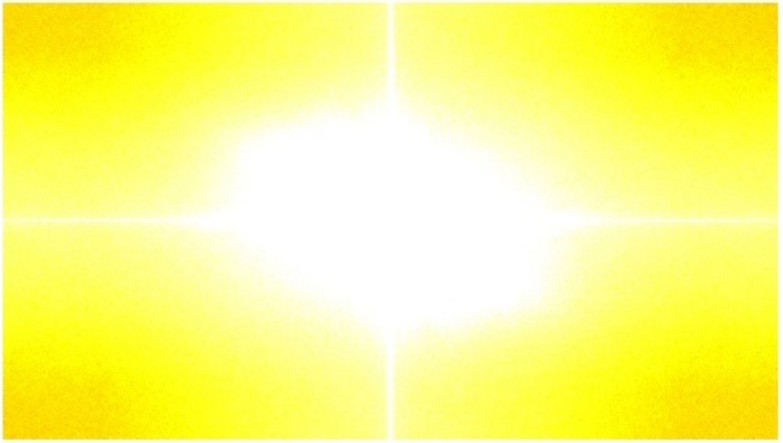}                &
			\includegraphics[width = 0.16\textwidth]{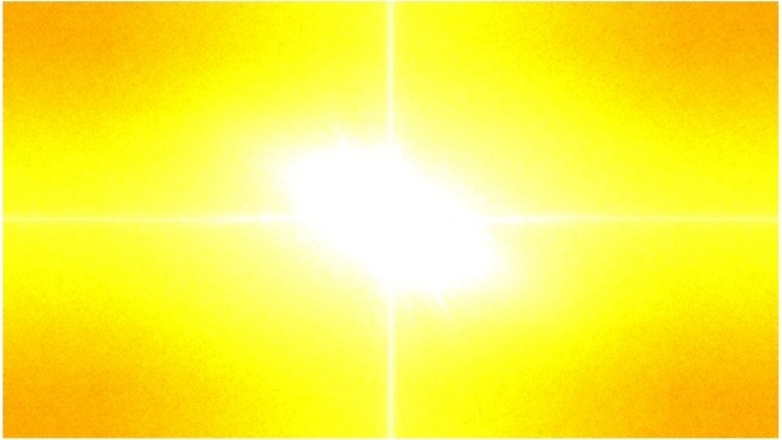}                 \\
			(a)  Blurry image  &
			(b)  Real part of (a)    &				
			(c)  Imag. part of (a)    & 
			(d)  Sharp image &  
			(e)  Real part of (d)    & 
			(f)  Imag. part of (d)    \\
			%(e)  Real part of (d)  &
			%(f)  Imag. part of (d) &
			%(g)  range of colors  \\
		\end{tabular}
	\end{center}
	%\vspace{-2mm}
	\caption{This figure shows that the change in the real/imaginary part of the Fourier coefficients on a pair of clear/blurred images. Note that the Fourier coefficients are normalized (the pixel range is 0$\sim$10) and executed on images sampled at 64 $\times$. }
	%\vspace{-2mm}
	\label{fig-ttf}
\end{figure*}

\section{Proposed Method}
\subsection{Wave-Frequency Processing}
\label{wfp}
Currently, Wave-MLP~\citep  {tang2021image} has made a big splash for high-level image processing task.
This is thanks to the fact that the real and imaginary parts of the Fourier coefficients as orthogonal signals are fused and splashed into the MLPs.
In other words, the real and imaginary parts of the Fourier coefficients are handled independently in the deep MLP network.
So far, we have to ask the question: \textit{Is it possible to split the real and imaginary parts of the Fourier coefficients as wave and particle signals to be processed independently for the deblurring task?} The answer is yes.
%Conventional methods usually involve a large number or large-scale convolution kernels in the spatial domain to accomplish the deblurring task.

In this study, we design a framework of \textbf{W}ave-\textbf{F}requency \textbf{P}rocessing  (WFP) to filter the real and imaginary components of the blurry input.
%Meanwhile, two limitations of image deblurring based on frequency domain information are avoided.
%
%we propose a new framework, distributed frequency domain processing framework (DFDP).
We begin our analysis using the example shown in Figure~\ref{fig-ttf}.
As shown, a blurry image and its real and imaginary parts after FFT are shown in Figure \ref{fig-ttf}(a)-(c), respectively.
%We can observe how the Fourier coefficients profiles of an image becomes after the blurs. 
With the image clears, both the real and imaginary components are changed from the high-frequency to a low-frequency domain as shown in Figure \ref{fig-ttf} (e) and (f). 
The main reason is that the real and imaginary components of the Fourier coefficients correspond to the phase change $\varphi$ of the image in the frequency domain (i.e., the displacement of the pixels in the spatial domain),
\begin{equation}
	\varphi	=~\text{arctan}(\mathcal{F}_{i}(B) \bigggl / \mathcal{F}_{r}(B)),
%	\varphi	=\mathrm{arctan}(\mathcal{F}_{i}(B) \bigggl / \mathcal{F}_{r}(B)),
\end{equation}	
%where $\varphi$ denotes the phase.
where $\mathcal{F}_{r}(B)$ and $\mathcal{F}_{i}(B)$ denote the real component and imaginary component extracted by FFT of the blurry input $B$, respectively, and arctangent ($\text{arctan}$) refers to the inverse tangent function.
The phase spectrum of the image retains information about the edges and the overall structure of the image~\citep  {1998Two}.
Based on the observation from Figure~\ref{fig-ttf}, it is confirmed that our motivation is supported by physical meaning.
Furthermore, it is known from quantum theory that the entropy increases during the transmission of orthogonal signals in the form of probability clouds, which requires a deep network to observe it. 
Therefore, we set up two parallel depth subnetworks to run the real and imaginary parts of the Fourier coefficients.

We propose the WFP framework for image deblurring, which can be formalized as
%
%	\begin{equation}
	%		\mathcal{Z} = \mathcal{F}^ {-1}(\mathcal{C}\{(\phi_{1}(\mathcal{F}(x_{\downarrow})_{r}), \phi_{2}(\mathcal{F}(x_{\downarrow})_{i})\}))\uparrow,	
	%	\end{equation}
%	%
%
\begin{equation}
	\label{SFDP}
	I = \mathcal{F}^ {-1}(\mathcal{C}\{\phi_{1}(\mathcal{F}_{r}(B)), \phi_{2}(\mathcal{F}_{i}(B))\}),	
\end{equation}
where $\phi_{1}$ and $\phi_{2}$ denote the mirror networks that act on the real and imaginary component, respectively; $\mathcal{C}$ indicates the Fourier coefficients combined from the features of the real part $\phi_{1}$ and the imaginary part $\phi_{2}$. 
Finally, the mapped Fourier coefficients is transformed to the sharp image $I$ by performing inverse Fourier transform ($\mathcal{F}^ {-1}$).
However, we note that if we choose the lightweight deblurring networks~\citep  {KupynMWW19,NahKL17,SuinPR20} as $\phi_{1}$ and $\phi_{2}$, we cannot directly handle 4K resolution images in a single GPU shader with 24G RAM by using the WFP framework.
Therefore, we propose to reformulate Eq.\eqref{SFDP} as:
\begin{equation}
	\label{SFDP_a}
	I = \{\mathcal{F}^ {-1}(\mathcal{C}\{\phi_{1}(\mathcal{F}_{r}(B_\downarrow)), \phi_{2}(\mathcal{F}_{i}(B_\downarrow))\})\} \uparrow,
\end{equation}
where $I$ is obtained by deblurring $B$ in a down-sampled $\downarrow$ resolution first, and then reconstructed by bicubic interpolation $\uparrow$. %(see Section \ref{label}).

	\subsection{Pipeline}
	Given a UHD blurry image $B\in \mathbb{R} ^{\left(W\times H\times C \right)}$, we use the multi-scale cubic-mixer to recover Fourier coefficients for a down-sampled sharp image reconstruction in the frequency domain.
	%Give an UHD blurry image, our model first reconstructs a simple sharp image via the proposed a Siamese-Mixer in the frequency domain. 
	Then, we propose a slicing scheme to generate a transform coefficient with a compact structure, which is used to reconstruct the high-definition sharp image from the output of the multi-scale cubic-mixer.
	% to obtain a high-definition sharp image. 
	Figure~\ref{frameworks} illustrates the architecture of the proposed UHD image deblurring network.

	\begin{figure*}[t]\scriptsize
		\begin{center}
			\tabcolsep 1pt
			\begin{tabular}{@{}cccccc@{}}
				\includegraphics[width = 0.16\textwidth]{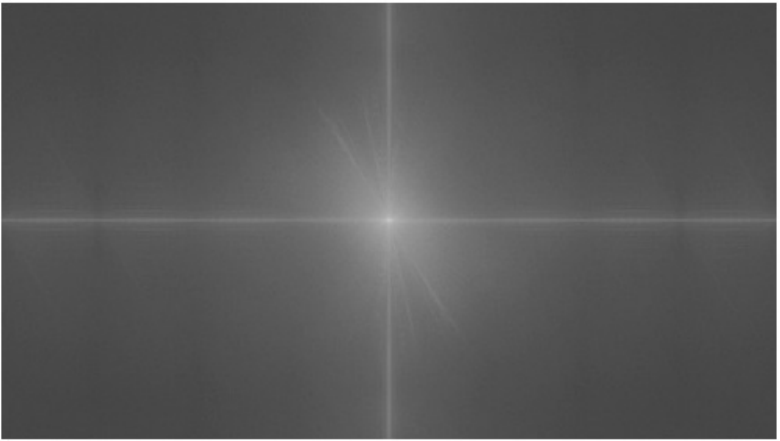} &
				\includegraphics[width = 0.16\textwidth]{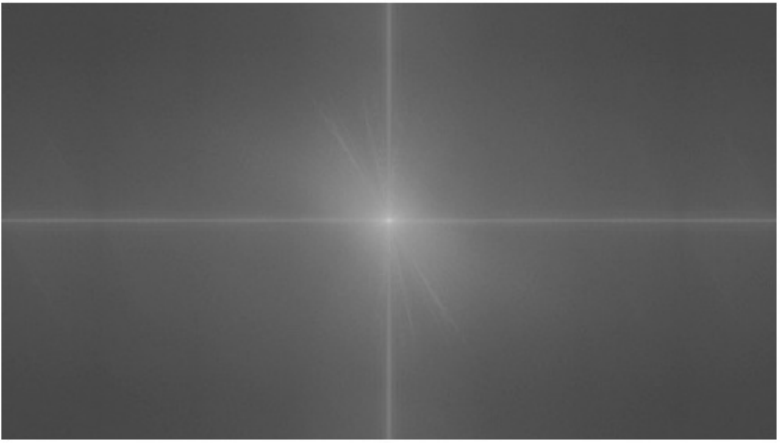}  &
				\includegraphics[width = 0.16\textwidth]{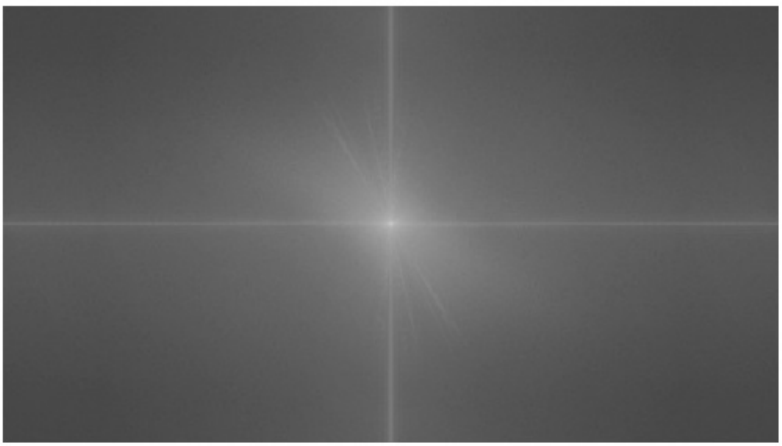}                 &
				\includegraphics[width = 0.16\textwidth]{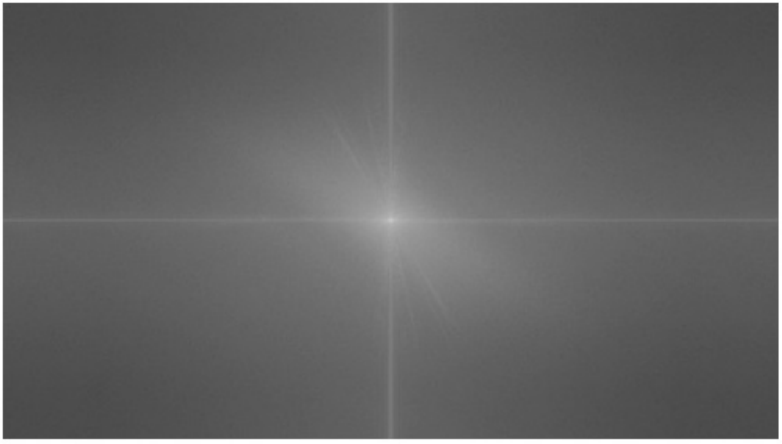}                 &
				\includegraphics[width = 0.16\textwidth]{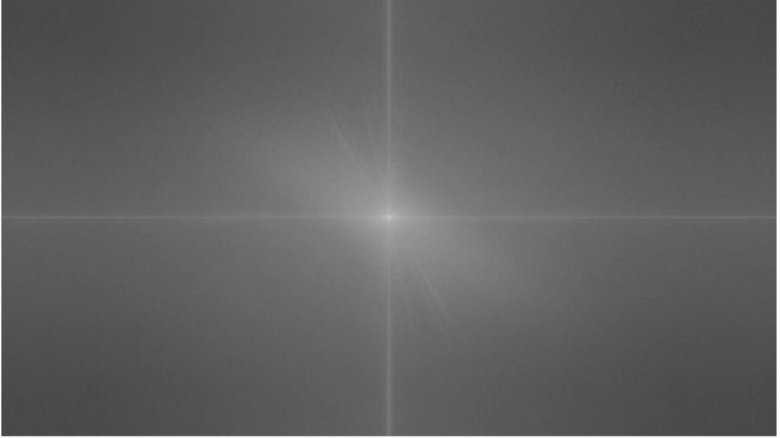}                 &
				\includegraphics[width = 0.16\textwidth]{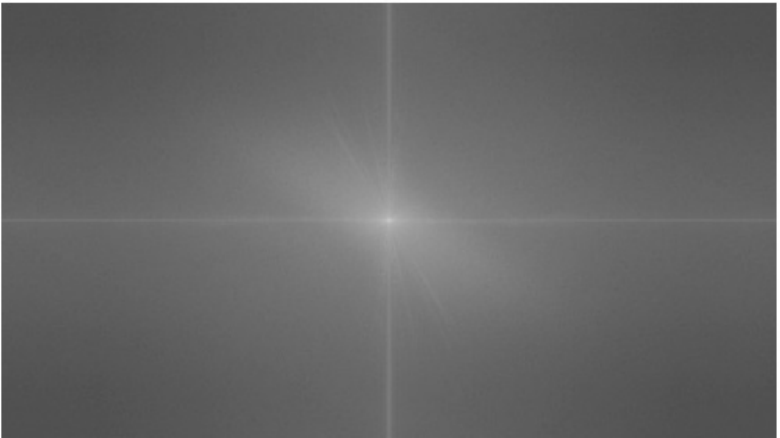}                 \\
				
				(a)  Blurry image  &
				(b)  1th    &				
				(c)  2th    & 
				(d)  3th    & 
				(e)  4th  &
				(f)  Sharp image \\
			\end{tabular}
		\end{center}
		%\vspace{-1mm}
		\caption{This figure shows the spectrum of the output of each yellow block in the top path of the network. 
			With the propagation of data through the network, the output spectrum of the trailing yellow block is closer to the spectrum of the clear image.
			Note that the output of each yellow block is transformed into a complex tensor before being rendered.}
		%\vspace{-2mm}
		\label{fig-fftt}
	\end{figure*}

	\textbf{Multi-scale cubic-mixer.}
	%Based on the observation in Section 3.1, it seems easier to restore a clear image in the frequency domain space. 
	The mainstream of deblurring networks usually under-performing on resource-constrained devices since they over-rely on self-attention scheme to model relationship of the token.
	To address this problem, we apply a multi-scale cubic-mixer, as shown in Figure~\ref{frameworks} (a), to recover sharp images by establishing long-range dependencies on the frequency domain.
	%
	%The main component of Siamese-Mixer are the fully-connected layer based on MLP-Mixer~\citep  {TolstikhinCoRR} and the 1D Fourier convolution.
	%
	The cell of multi-scale cubic-mixer is constructed by paralleling two identical cubic-mixer with the WFP framework (Eq.\eqref{SFDP_a}) in the yellow block of Figure~\ref{frameworks} (a).

	The main components of the cubic-mixer are the fully-connected layers based on modified MLP-Mixer.
	%
	%In order to be able to quickly establish long-range dependencies on the Fourier coefficients,  MLP-Mxier is referenced.
	% 这边要改
	%The MLP-Mixer consists purely of a fully connected neural network that discards the self-attention mechanism and accomplishes the task through residuals with channel attention.
	The cubic-mixer achieves the `mixing' enhancement of element spatial position and channel information in the frequency domain, which can capture long-range of perceptual fields.
	Compared to MLP-Mixer~\citep  {TolstikhinCoRR}, there are two improvements, i) our model does not conduct the input information in patches, and the native input information is swallowed into the deep network, and; ii) what comes naturally is that we have one more dimensional signal to process than MLP-Mixer, here we establish the long-range dependencies by rolling on three dimensions $C$, $H$, and $W$. 

			\begin{figure*}[t]\scriptsize
				\begin{center}
					\tabcolsep 1pt
					\begin{tabular}{@{}cccccc@{}}
						\includegraphics[width = 0.16\textwidth]{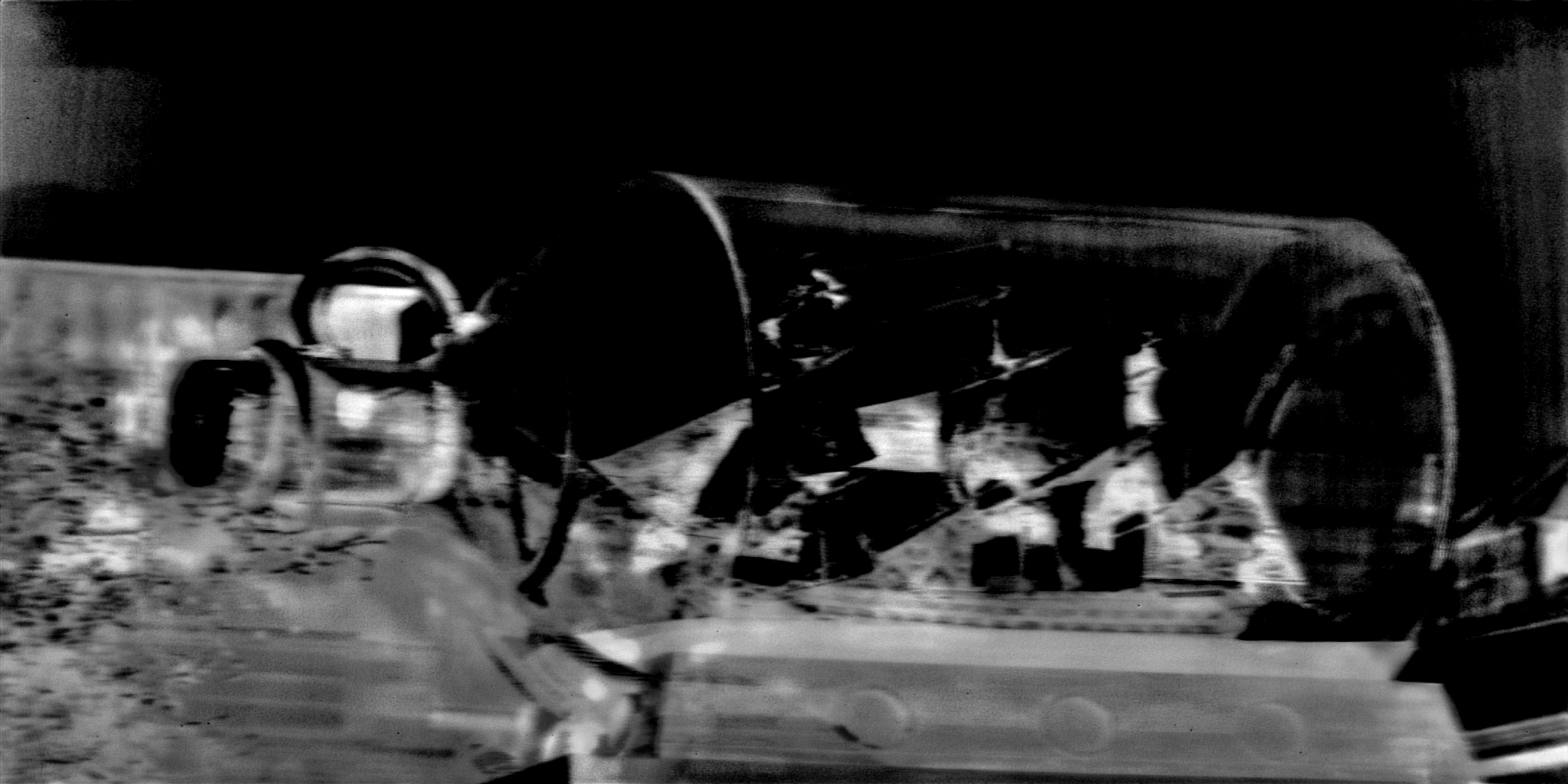}  &
						\includegraphics[width = 0.16\textwidth]{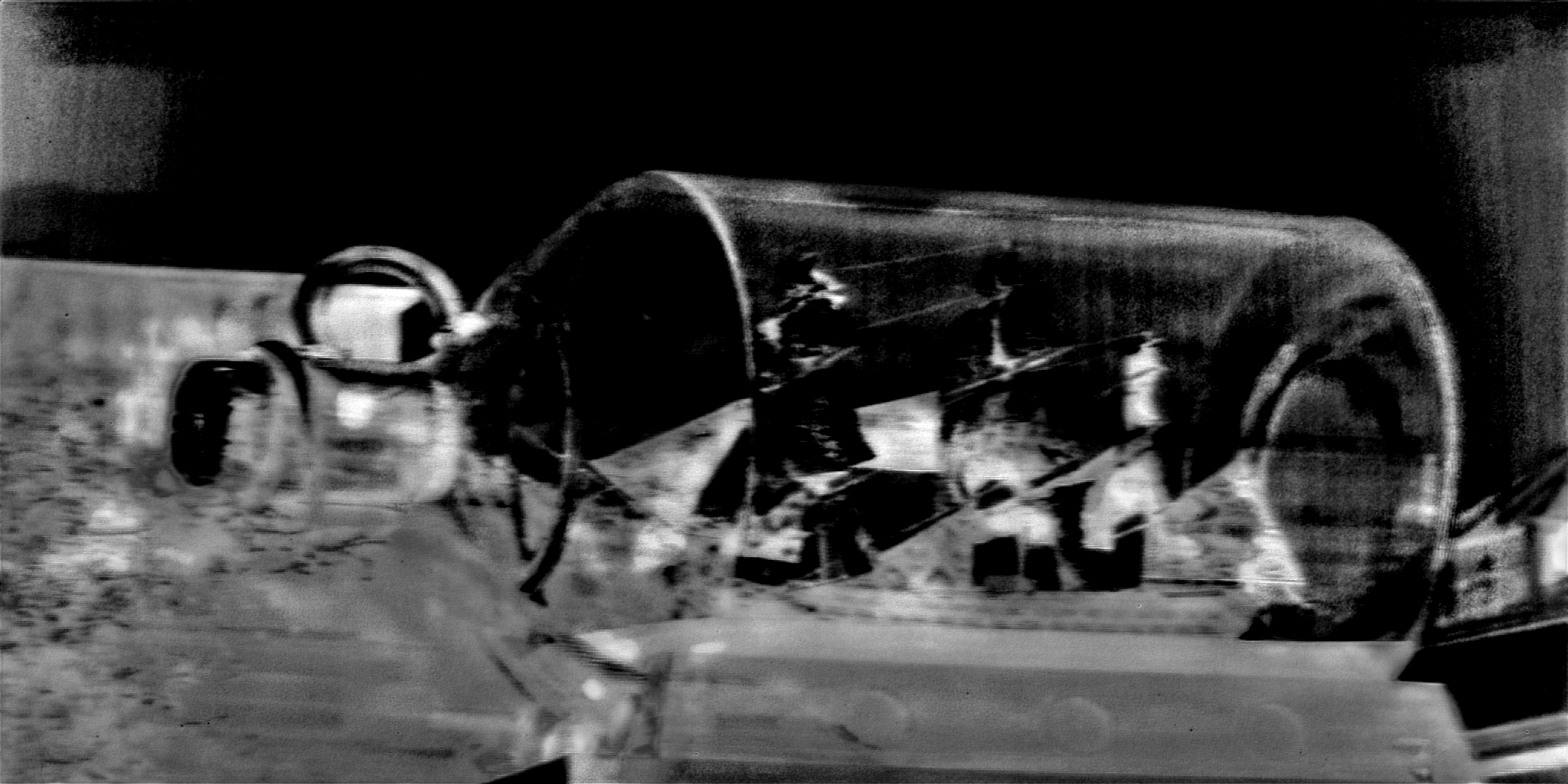}                 &
						\includegraphics[width = 0.16\textwidth]{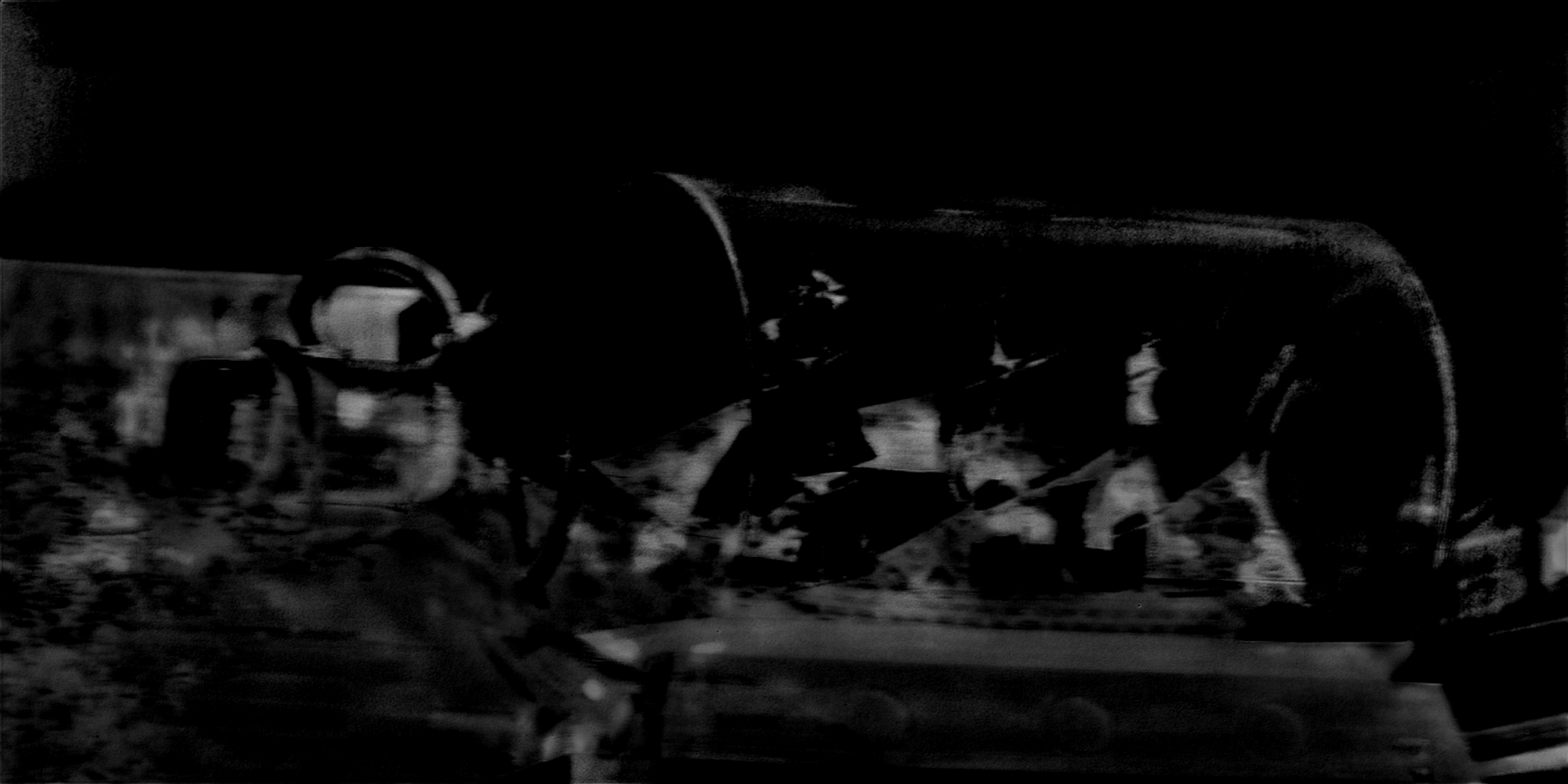}                 &
						\includegraphics[width = 0.16\textwidth]{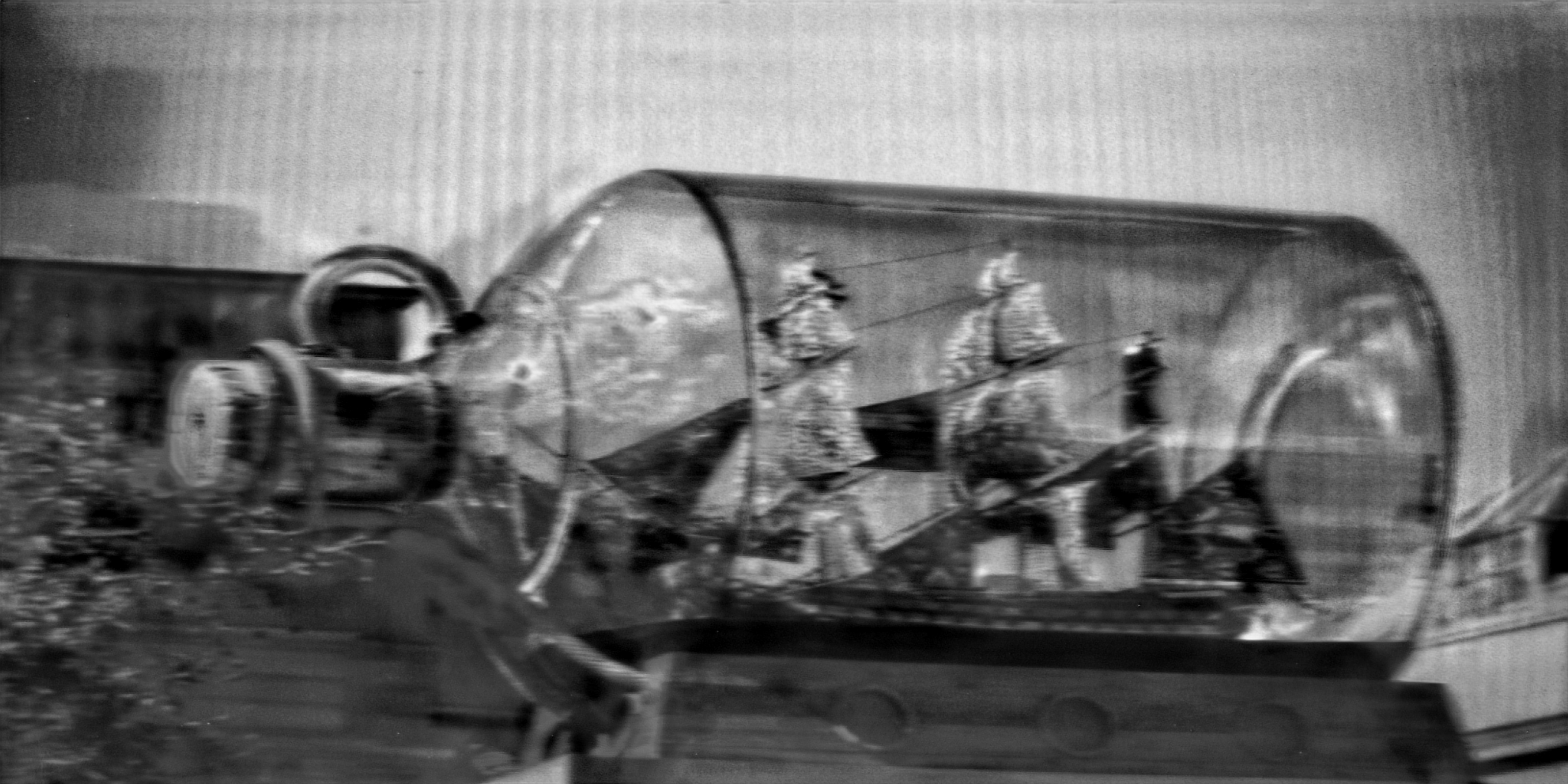}                 &
						\includegraphics[width = 0.16\textwidth]{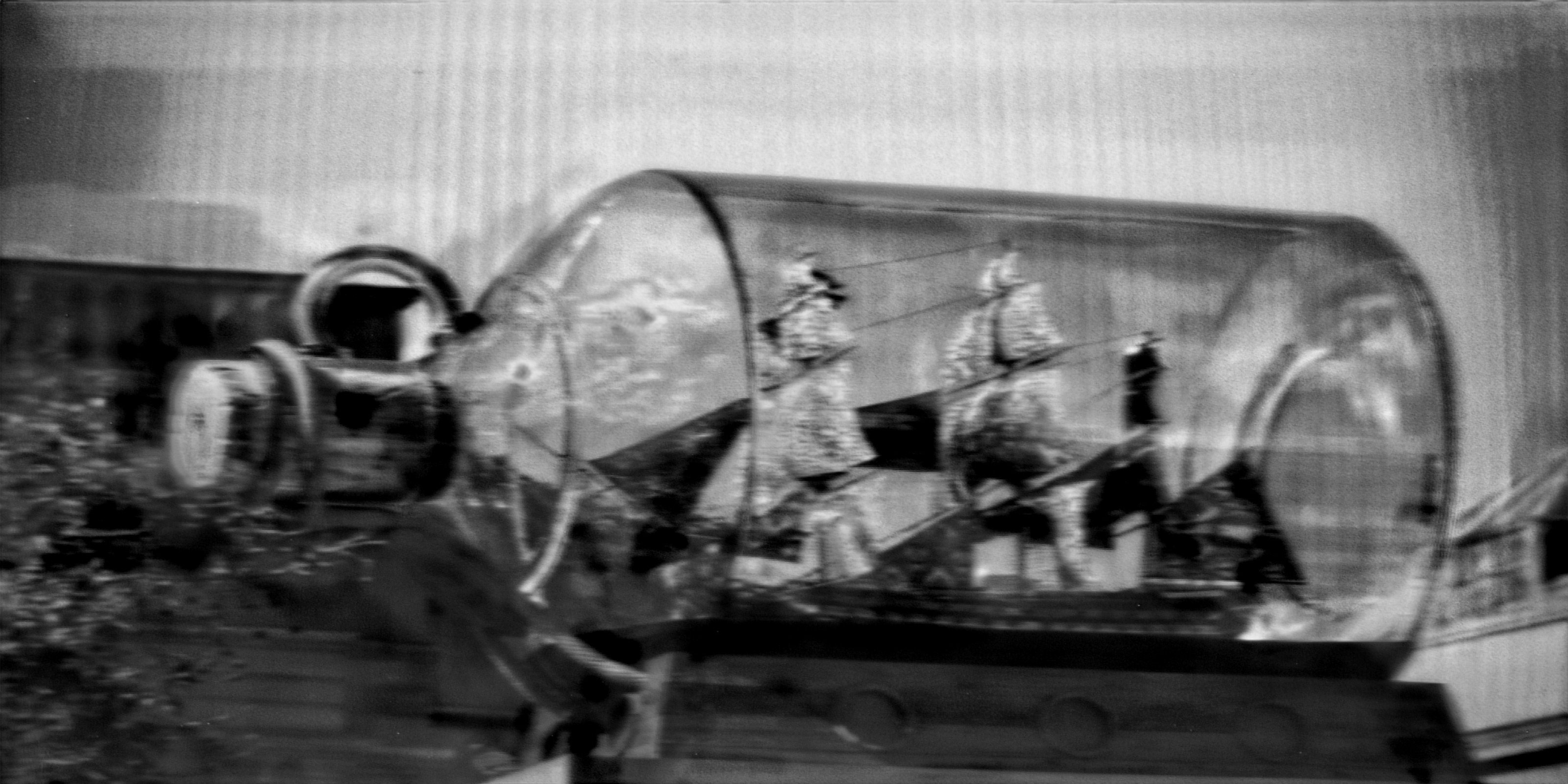}                 &
						\includegraphics[width = 0.16\textwidth]{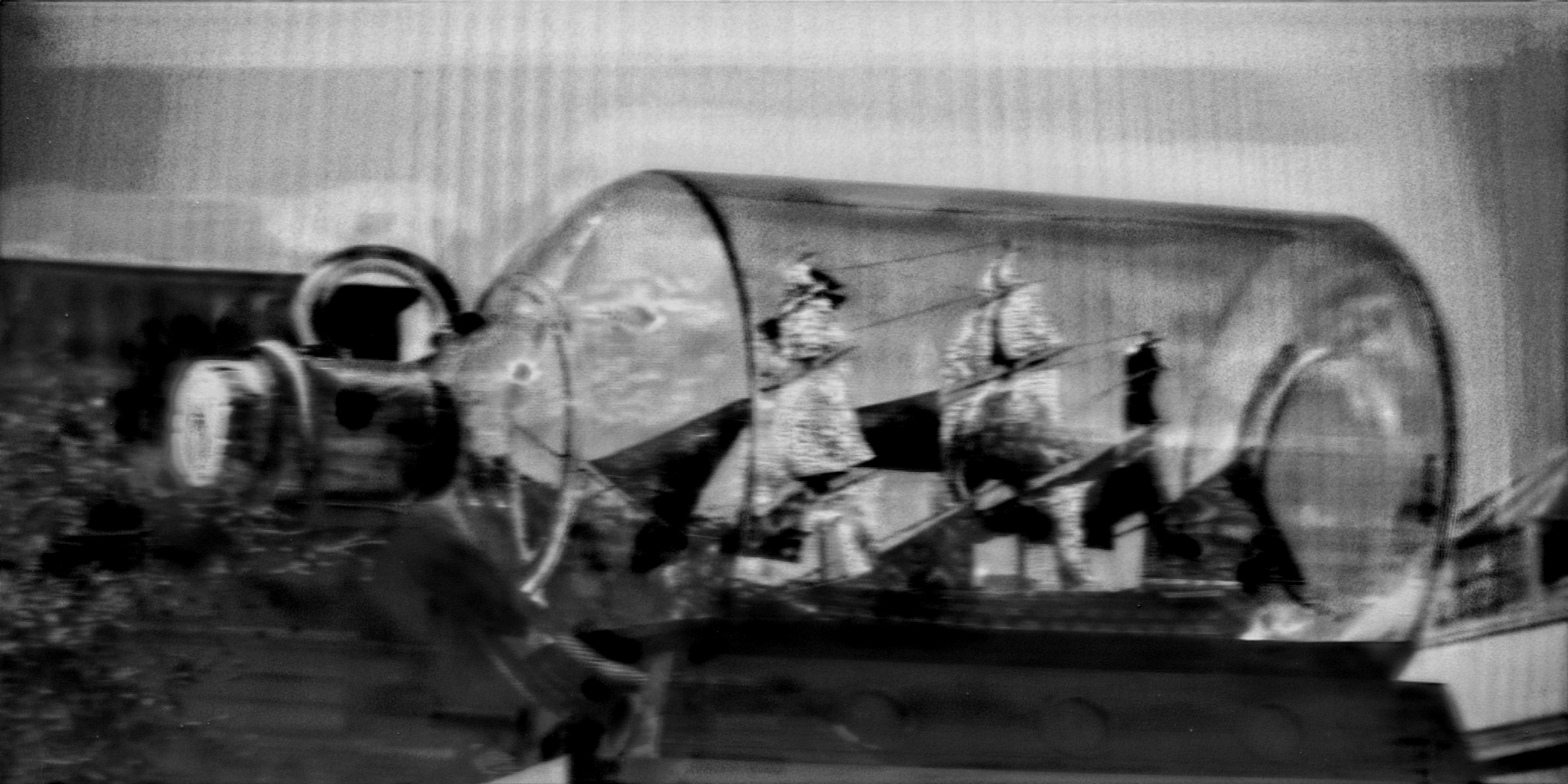}                 \\
						%				\includegraphics[width = 0.12\textwidth]{slicing_gts.jpg}                 \\
						
						%	(a)  Blurry image  &
						(a)  $W_{red}$    &				
						(b)  $W_{green}$   & 
						(c)  $W_{blue}$    & 
						(d)  $b_{red}$  &
						(e)  $b_{green}$  &
						(f)  $b_{blue}$  \\
						%	(h)  Sharp image \\
					\end{tabular}
				\end{center}
				%\vspace{-2mm}
				\caption{This figure shows the characteristics of each feature map in the slicing operation, and it is clear that the details acting on the three groups of feature maps in the color channel are complementary to each other.}
				%\vspace{-2mm}
				\label{fig-scling}
			\end{figure*}
			
			For instance, one path of the proposed network consists of 4$\times$ cubic-mixer with WPF framework (the yellow blocks).
			Every cubic-mixer includes three \textit{MLPs}, and each \textit{MLPs} involves two linear layers as well as the corresponding activation layer.
			The first \textit{MLPs} mixtures the width-dimension with shared parameters, maps $ {\mathbb{R}}^W\rightarrow {\mathbb{R}}^W $. The second \textit{MLPs} mixtures the height-dimension with shared parameters, maps $ {\mathbb{R}}^H\rightarrow {\mathbb{R}}^H $. The last \textit{MLPs} mixtures the channel-dimension with shared parameters, maps $ {\mathbb{R}}^C\rightarrow {\mathbb{R}}^C $.
			Specifically, for the input tensor  $X\in {\mathbb{R}} ^{\left(W\times H\times C \right)}$, the  transformation in each mixer layer can be written as follows (the mixer layer index is omitted):
			%\begin{small}
			\begin{equation}
				F_{i,*,*} = X_{i,*,*}+w_{2}\sigma{\left(w_{1}{\left(X\right)}_{i,*,*} \right)},  ~~~{\rm for}~~i=1...W
			\end{equation}
			%\end{small}
			%\begin{small}
			\begin{equation}
				F_{*,j,*} = F_{*,j,*}+w_{4}\sigma{\left(w_{3}{\left(F\right)}_{*,j,*} \right)},  ~~~{\rm for}~~j=1...H
			\end{equation}
			%\end{small}
			%\begin{small}
			\begin{equation}
				F_{*,*,k} = F_{*,*,k} +w_{6}\sigma{\left(w_{5}{\left(F\right)}_{*,*,k}  \right)},  ~~~{\rm for}~~k=1...C
			\end{equation}
			
			\noindent where $w_{i}$ is the parameter of the $i$th fully connected layer, $ \sigma $ is the ReLU function.
			
			%the proposed Cubic-Mixer takes a sequence of equally divided images patches $\{p_i \}^s_{i=1}$ as the input. Each patch (with resolution of $32 \times 32$) is then projected to the desired hidden layer dimension of $1 \times l$, where $l=4096$, which results in a two-dimensional matrix $X \in \mathbb{R} ^{s \times l}$, where the number of patches is $s=mn \bigggl / 32^{2}$.
			%
			
			%

			%
			Specifically, we downsample the blurry input $B$ to a low-resolution of $B_\downarrow \in \mathbb{R} ^{\left(960\times 540\times C \right)}$ in the top path, and then feed the Fourier coefficients $\mathcal{F}_{r}(B_\downarrow)$ into cubic-mixer.
			%
			%On the $C$ dimension is then projected to the desired hidden layer dimension of $1 \times l$, where $l=3$, $l=960$ on the $W$ dimension, and $l=540$ on the $H$ dimension.
			%
			The proposed cubic-mixer has three rolls, and each roll consists of \textit{MLPs} as shown in Figure \ref{frameworks}(b).
			%
			%\textcolor{red}{The MLP block has the same structure as the MLP block in MLP-Mixer.}
			%
			%We set three different scale convolution kernels of $7 \times 7$, $5 \times 5$, and $3 \times 3$ to capture the structural information of inner the patch from coarse to fine for the 1D convolution block.
			%
			%\textcolor{red}{The size of the output image of MC-Mixer is the same as the size of the input image (the dimensionality of the last hidden layer in MC-Mixer is $m \times n$). }
			%
			
			The real part and the imaginary part feed into two cubic-mixer respectively to obtain the real part features and the imaginary part features.
			Finally, they are combined into a complex number tensor to obtain the feature maps in spatial domain $I$ through $\mathcal{F}^ {-1}$.
			The middle and bottom paths in the network are mirror versions of the top path, where the hidden layer is a fixed value for the input dimension.
			In addition, we show the Fourier coefficient change in the top path of the network as shown in Figure~\ref{fig-fftt}.
			{\flushleft \textbf{Local feature exaction and slicing enhancement.}} 
			Although the multi-scale cubic-mixer has the capability to construct a low-cost global perspective on a blurred image, local rendering can still be bad. 
			To this end, we use several convolutional layers to learn the local features of a blurred image in complementary manner.
			Specifically, as shown in Figure~\ref{frameworks}(a), our network splashes out of the three groups of feature maps are upsampled to three high-quality feature maps ($F_{t}$, $F_{m}$, and $F_{b}$), respectively.
			The raw image $B$, $F_{t}$, $F_{m}$, and $F_{b}$ are concatenated together to obtain a full-resolution tensor $T$ on the channel domain.
			Next, $T$ passes through the $3 \times 3$ and the $1 \times 1$ convolutional layer and the corresponding activation layer (PReLU), where two items need to be noted.
			$3 \times 3$ convolutional layer does not change the number of channels, while $1 \times 1$ convolutional layer squeezes the number of channels of $T$ to 6, which prepares the appetizer for the later slicing operation.

			Inspired by bilateral learning~\citep  {zheng2021ultra}, for UHD images, instead of directly regressing the enhanced image, we can fit a tensor $T$ with the attention characteristics to act on the original image.
			Specifically, $T$ is regarded as three groups of affine transform operations ($W_{red}$, $b_{red}$; $W_{green}$, $b_{green}$; $W_{blue}$, $b_{blue}$) on the three channels of the original image $B$.
			We can formalize it as:
			\begin{align}	
				&I_{sharp}(r) = W_{red} \odot B_{red} + b_{red},\\
				&I_{sharp}(g) = W_{green} \odot B_{green} + b_{green},\\
				&I_{sharp}(b) = W_{blue} \odot B_{blue} + b_{blue}. 
			\end{align}
			As shown in the Figure~\ref{fig-scling}, the three channels of the original image are affine transformed into a sharper spatial domain.
			These feature maps complement each other's texture details on the color channels.
			\textbf{The polynomial slicing method} is also considered, i.e., given a quadratic ($(W \odot B + b)^{2}$) to obtain a sharp image. 
			The details are described in the supplementary material.
				\subsection{Loss Function}
				%
				%\vspace{-2mm}
				We optimize the weights and biases of the proposed network by minimizing the $L_{1}$ and perceptual loss~\citep  {JohnsonAF16} on the training set,
				\begin{equation}
					\mathcal{L}=\frac{1}{N}\sum_{i=1}^{N} \left \| \hat{I}_{i}-J_{i} \right \| + \lambda\mathcal{L}_{p},
				\end{equation}
				where $N$ is the number of training images, $\hat{I}$ is the deblurred result by our model, and $J$ is the corresponding ground truth. The weight $\lambda$ of the loss term $\mathcal{L}_{p}$ is set to 0.03 in our experiments.  
				%$\mathcal{L}_{p}$ is used to improve visual quality.
				We use VGG19~\citep  {SimonyanZ14a} as the pre-trained model for the perceptual loss function.
				We also try to use the total variation and adversarial losses, but we notice that
				the $L_{1}$ and $\mathcal{L}_{p}$ can generate vivid colors and clear texture in the deblurred results.
				%\vspace{-2mm}
				\section{Experiments}
				%\vspace{-2mm}
				In the section, we evaluate the proposed algorithm on both synthetic datasets and real-world 4K images against the state-of-the-art image deblurring methods in terms of accuracy and visual effect. 
				In addition, we conduct three ablation studies to demonstrate the effectiveness of each module of our method.
				The implementation code and datasets will be available to the public for further discussion and research.
				More results can be found in the supplementary material.

				\begin{figure*}[h]\scriptsize
					\begin{center}
						\tabcolsep 1pt
						\begin{tabular}{@{}ccccccccc@{}}
							\includegraphics[width = 0.11\textwidth]{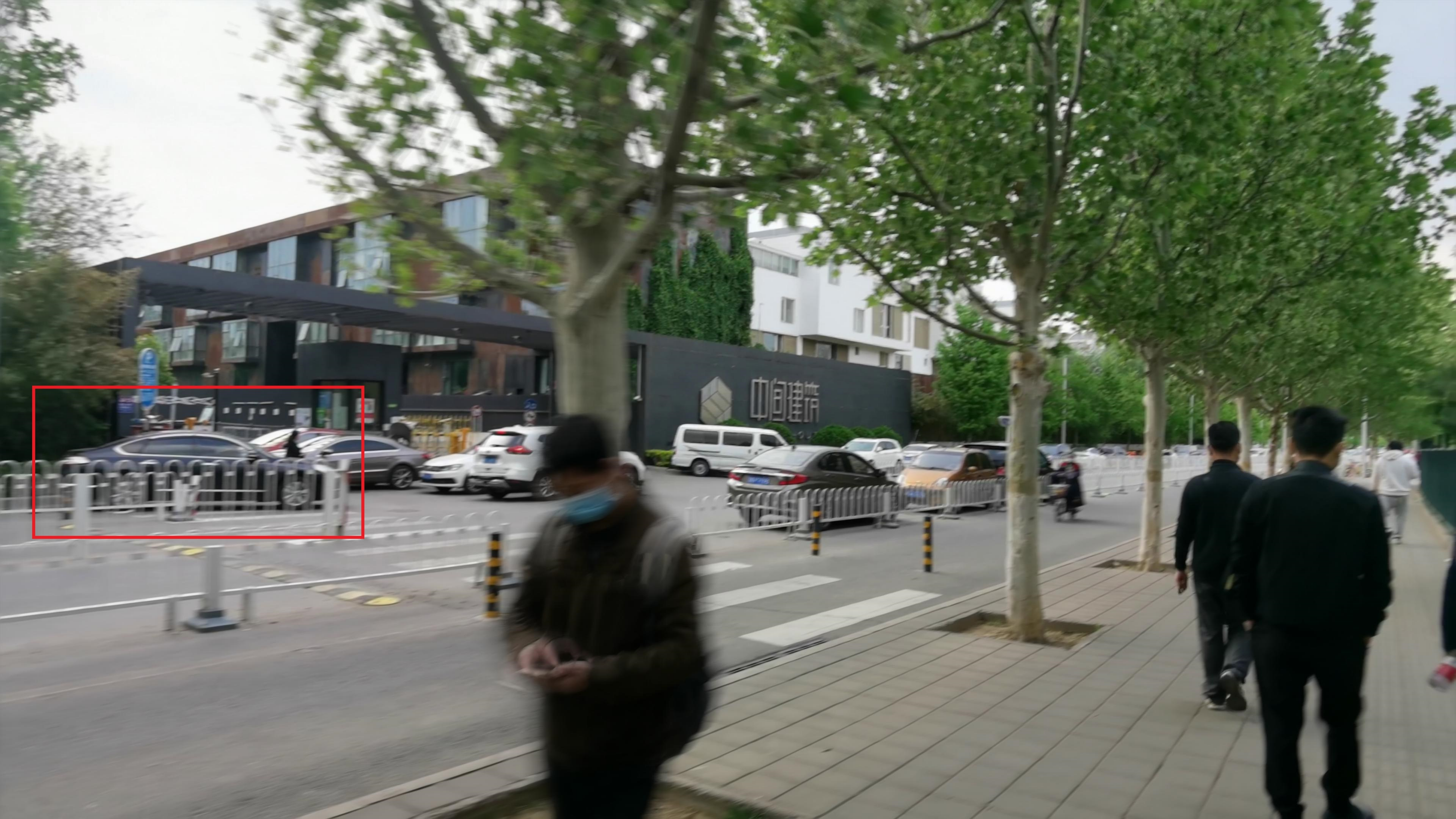}                &
							\includegraphics[width = 0.11\textwidth]{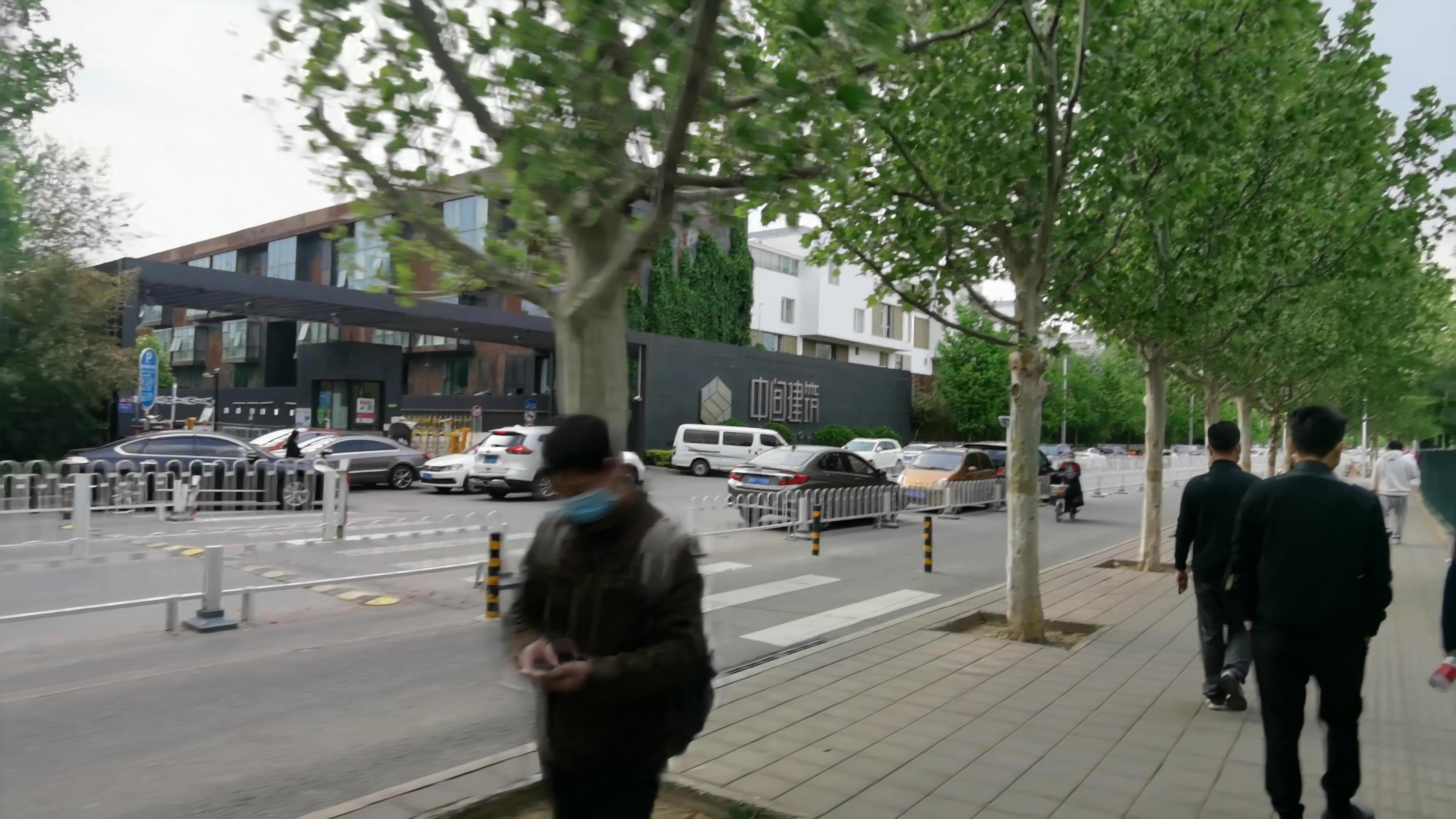}            &
							\includegraphics[width = 0.11\textwidth]{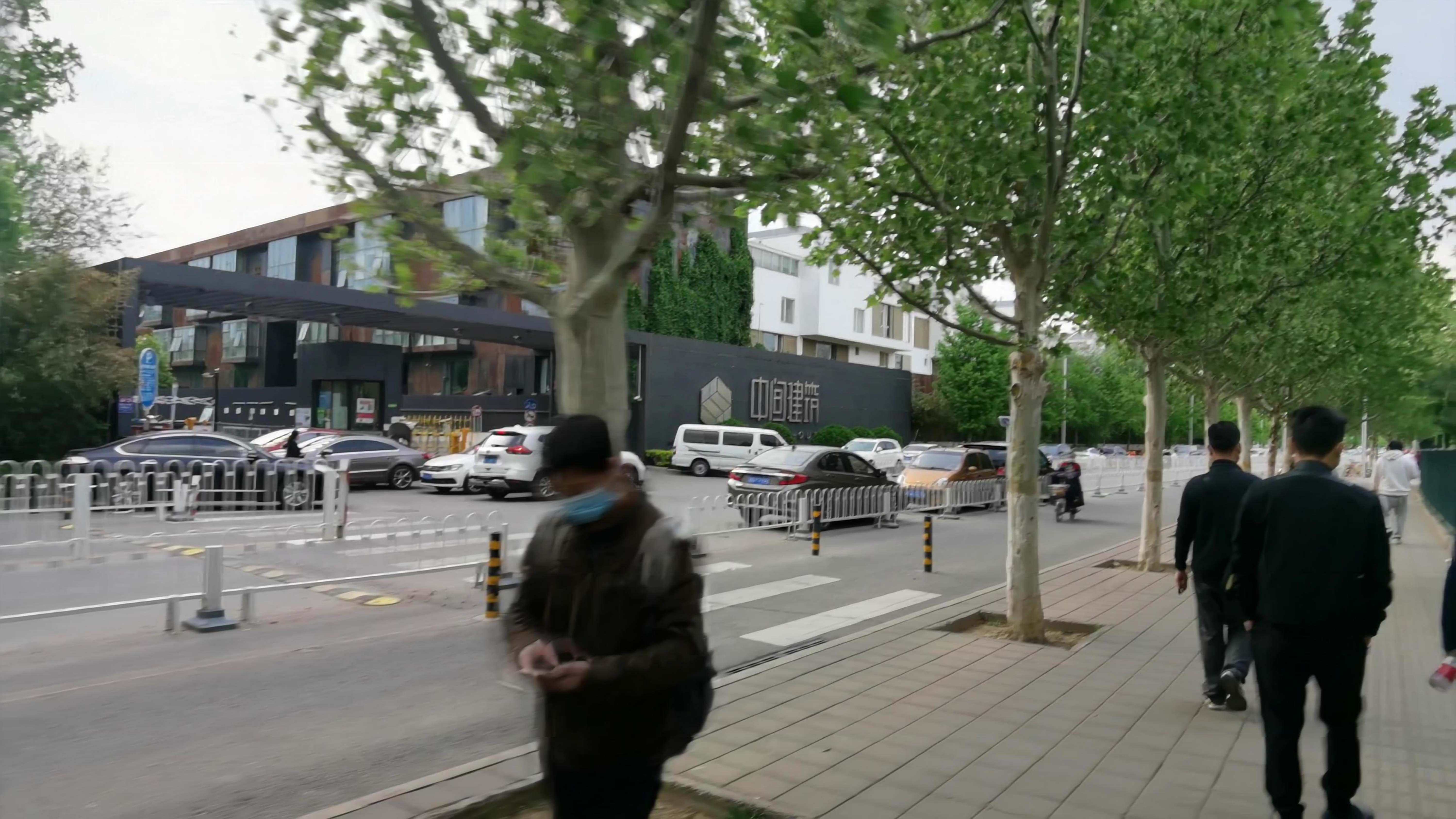}                         &
							\includegraphics[width = 0.11\textwidth]{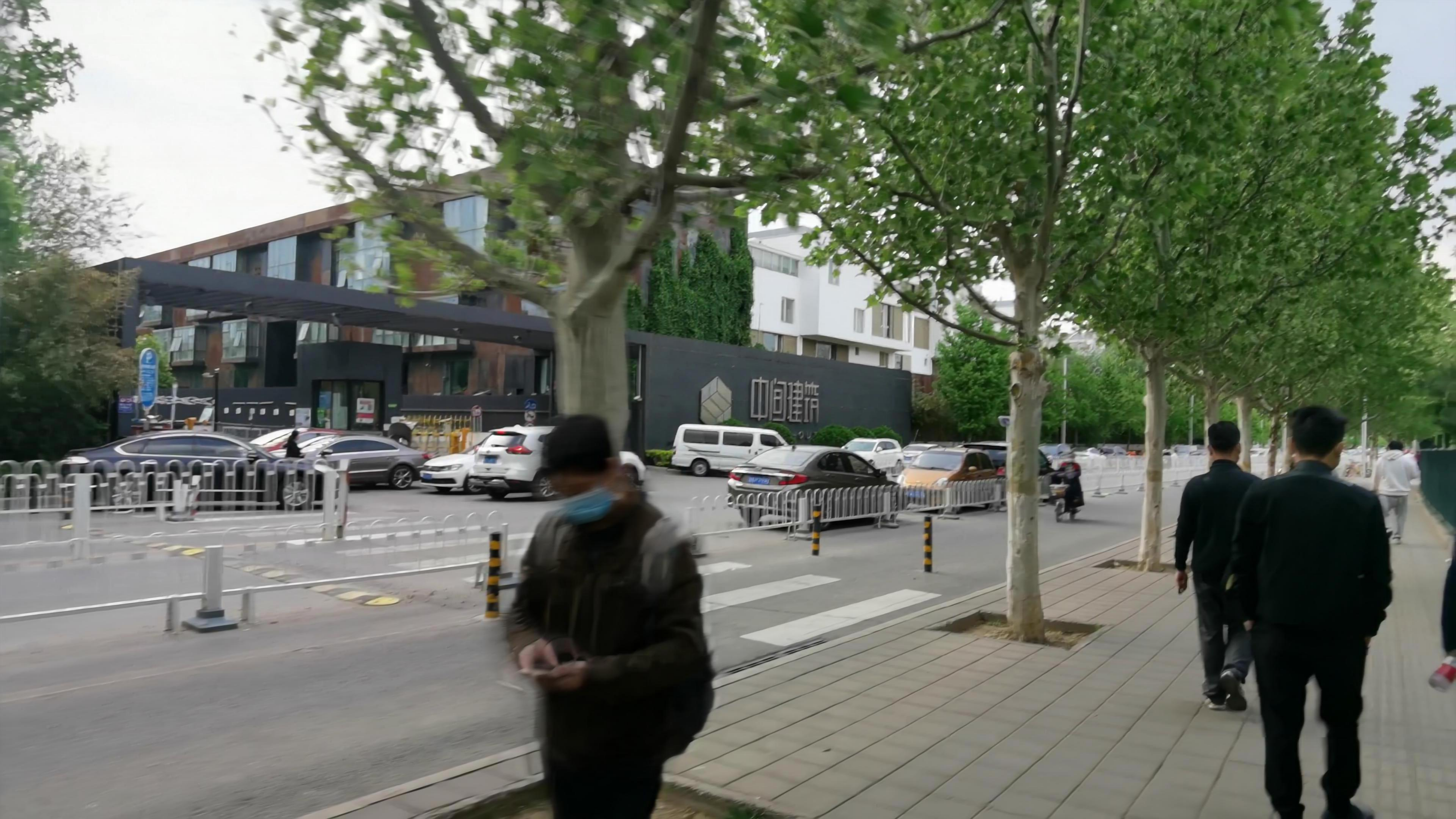}                         &   
							\includegraphics[width = 0.11\textwidth]{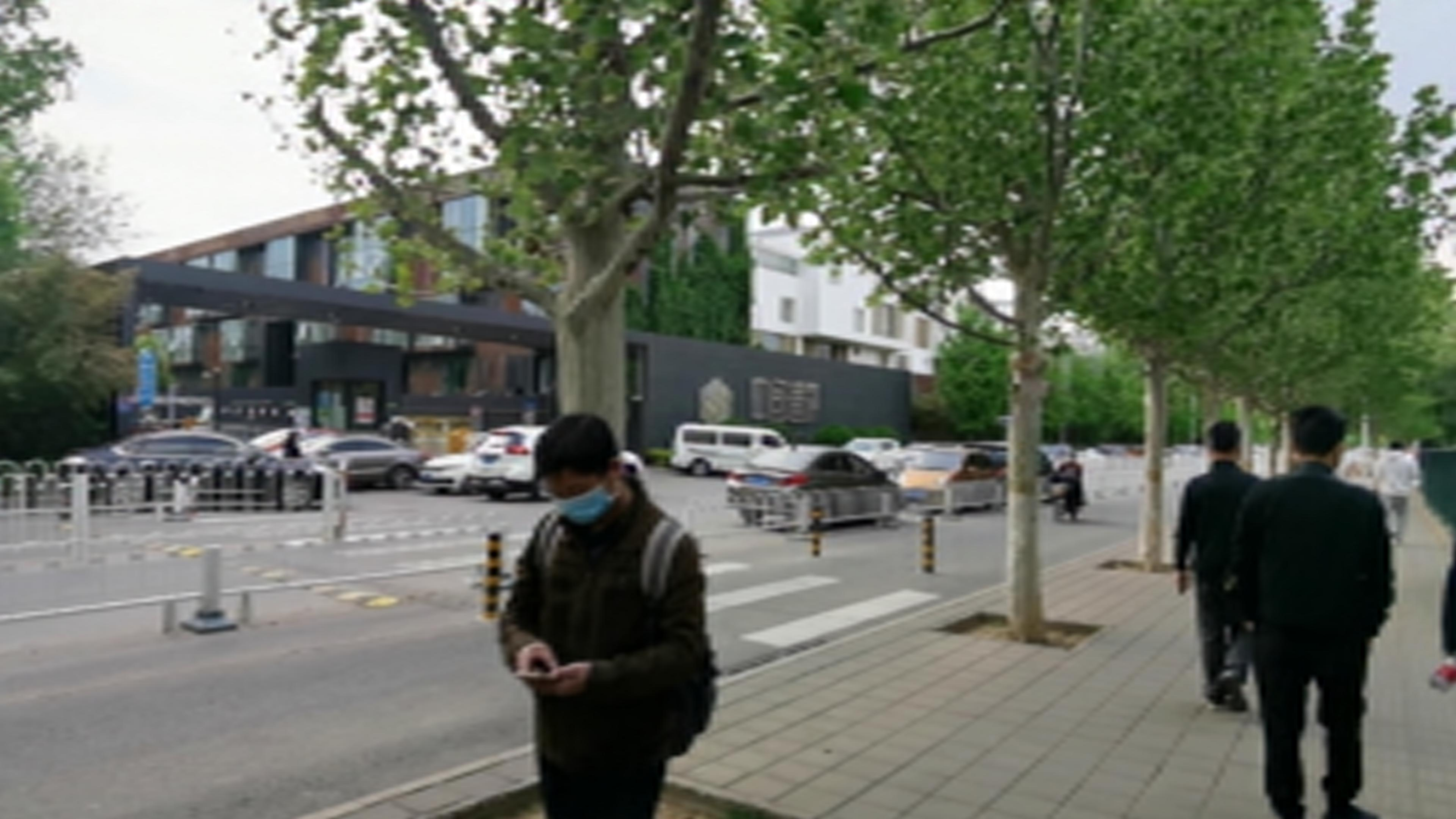}                         & 
							\includegraphics[width = 0.11\textwidth]{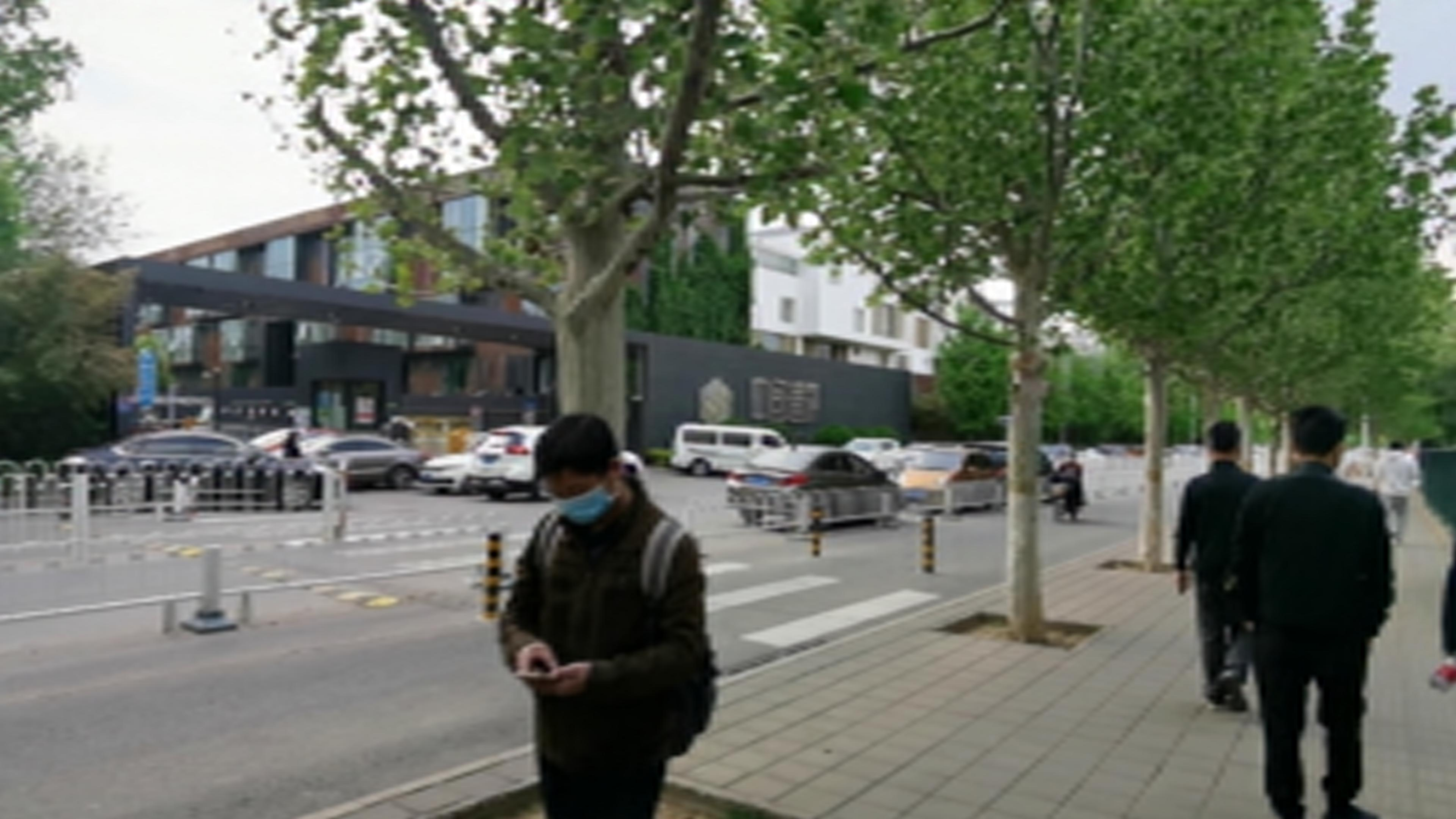}                         &
							\includegraphics[width = 0.11\textwidth]{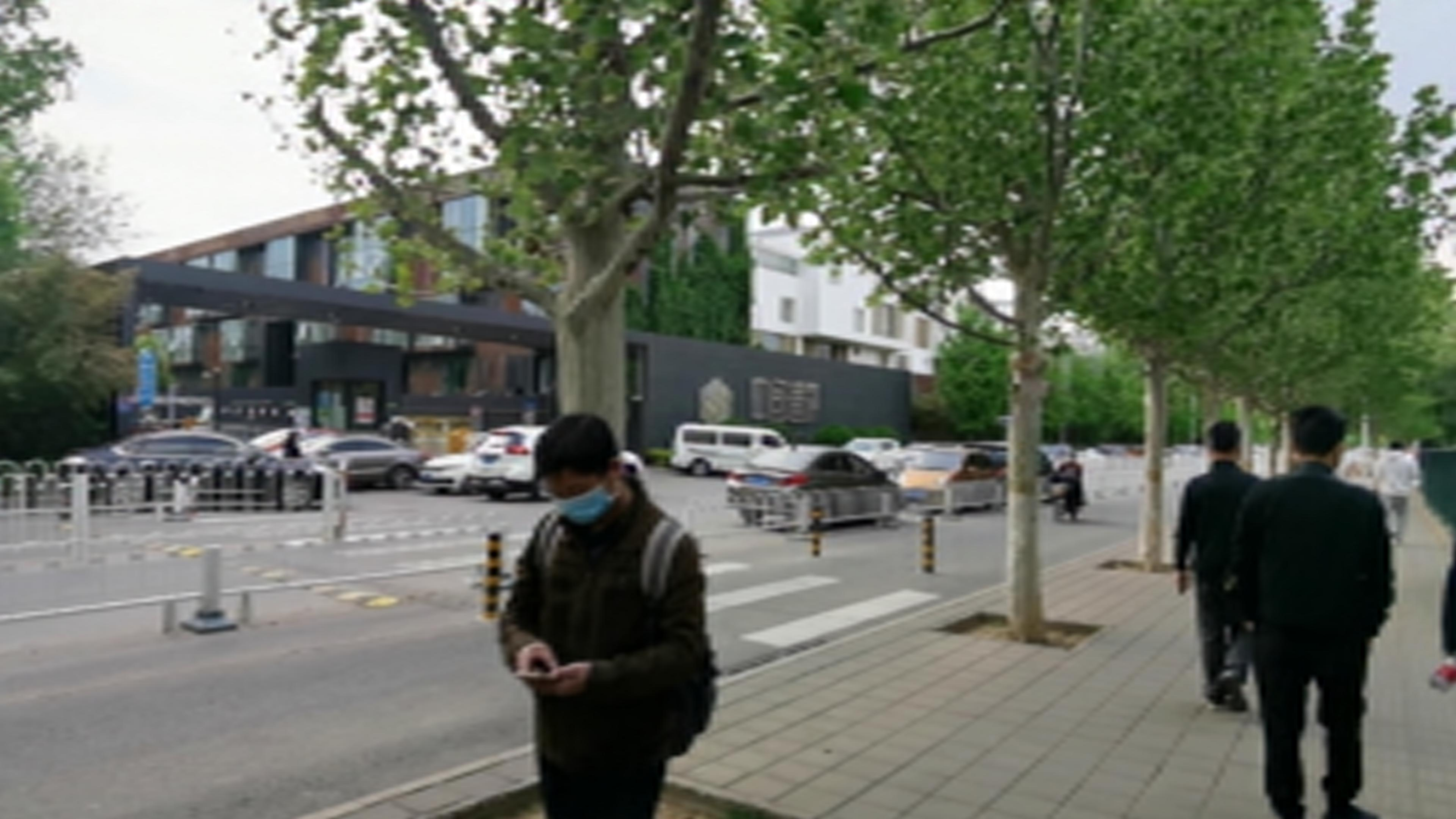}                         &
							\includegraphics[width = 0.11\textwidth]{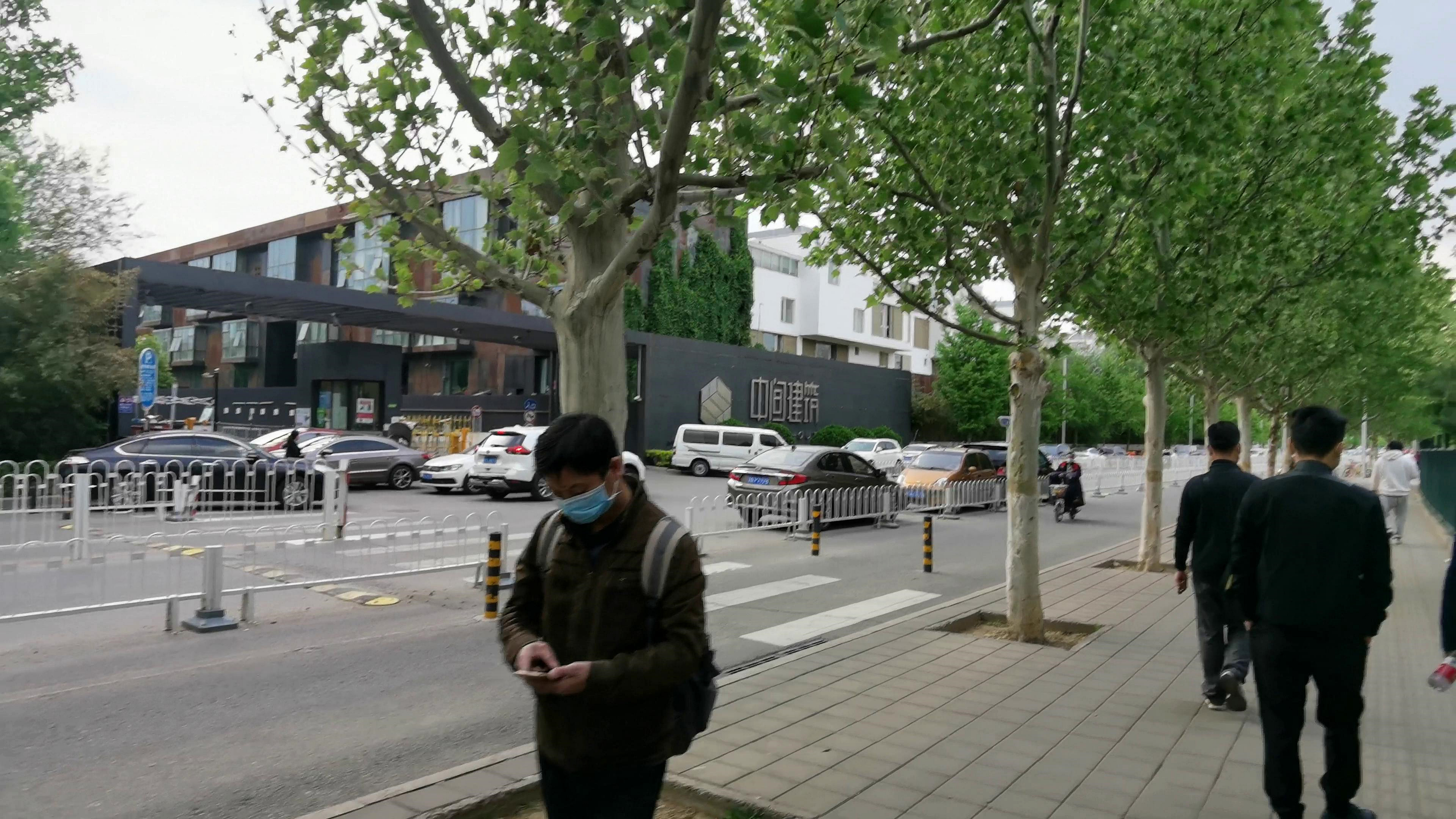}  &
							\includegraphics[width = 0.11\textwidth]{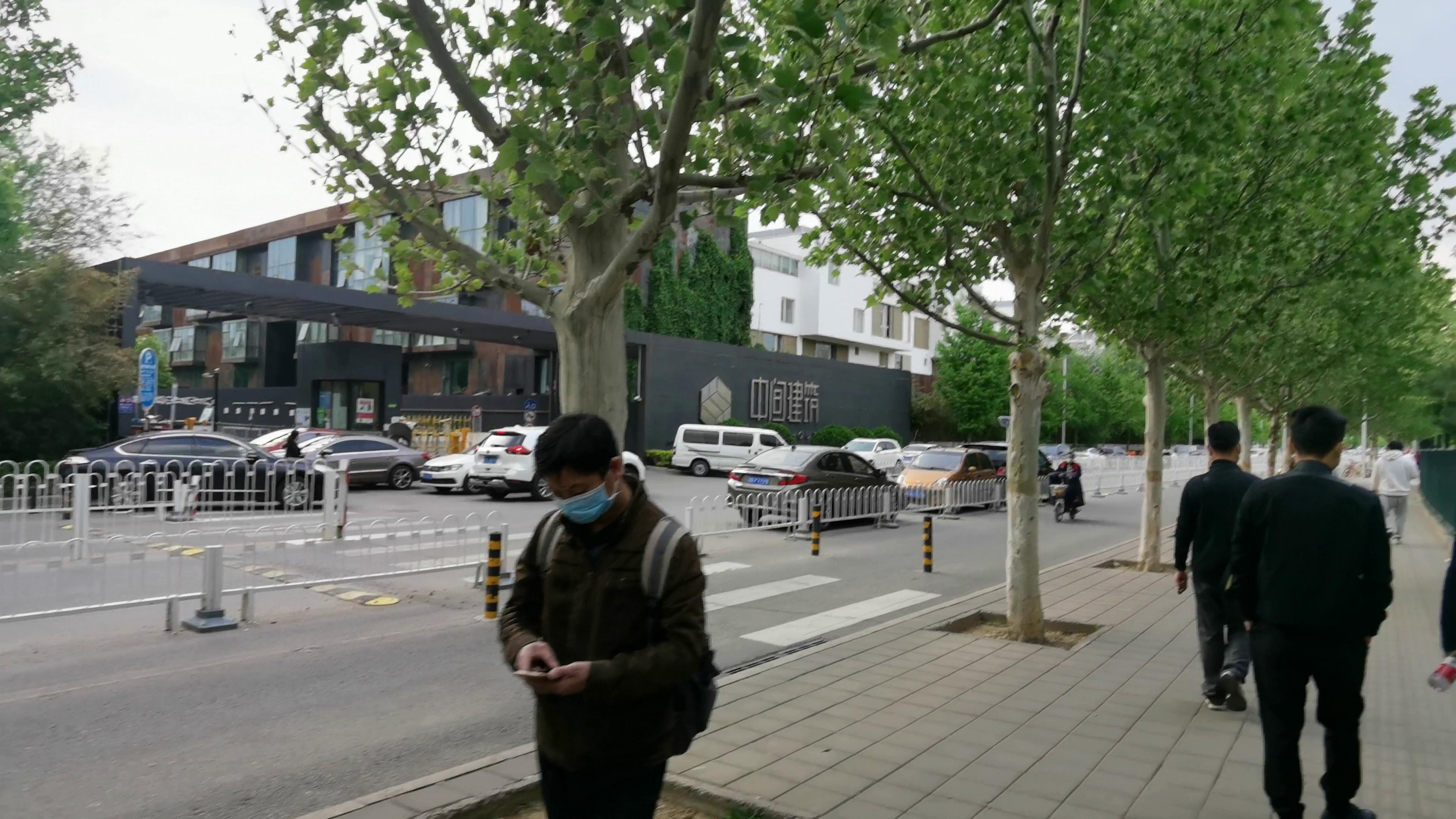}      
							\\
							
							\includegraphics[width = 0.11\textwidth]{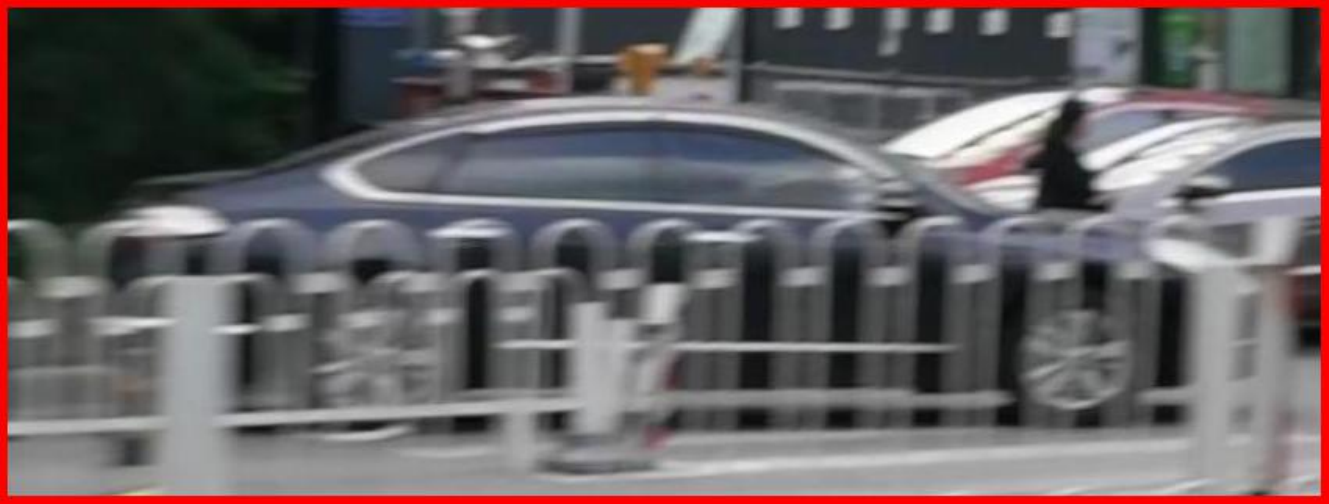}                &
							\includegraphics[width = 0.11\textwidth]{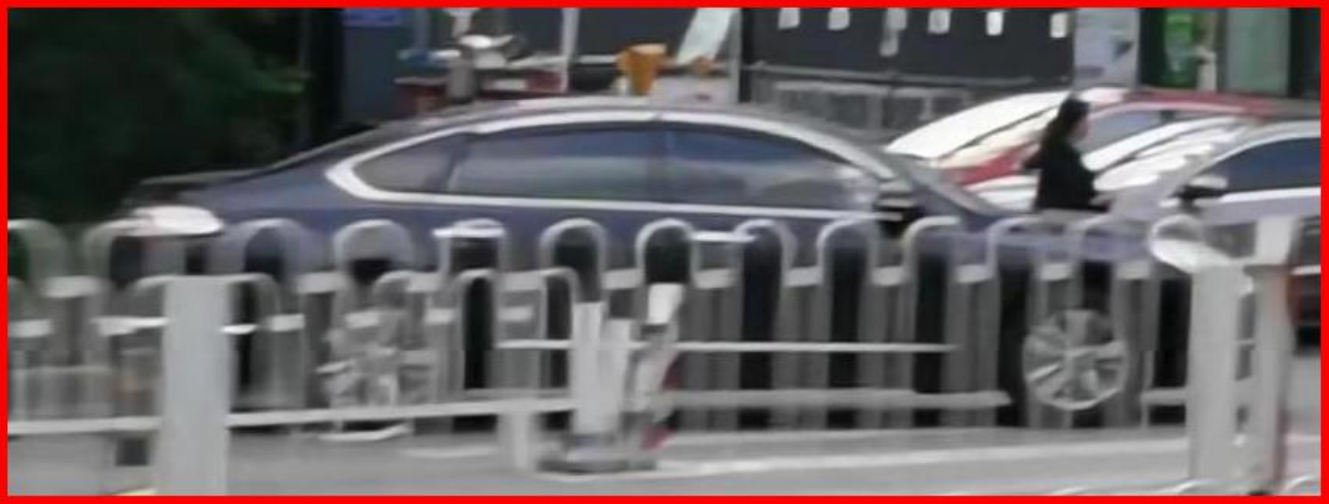}            &
							\includegraphics[width = 0.11\textwidth]{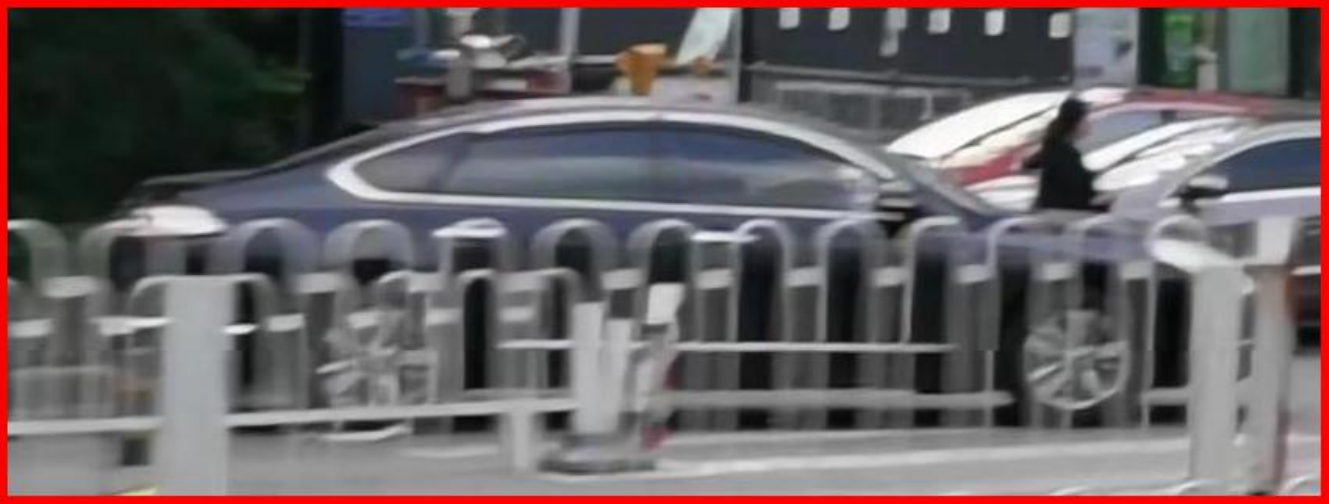}                         &
							\includegraphics[width = 0.11\textwidth]{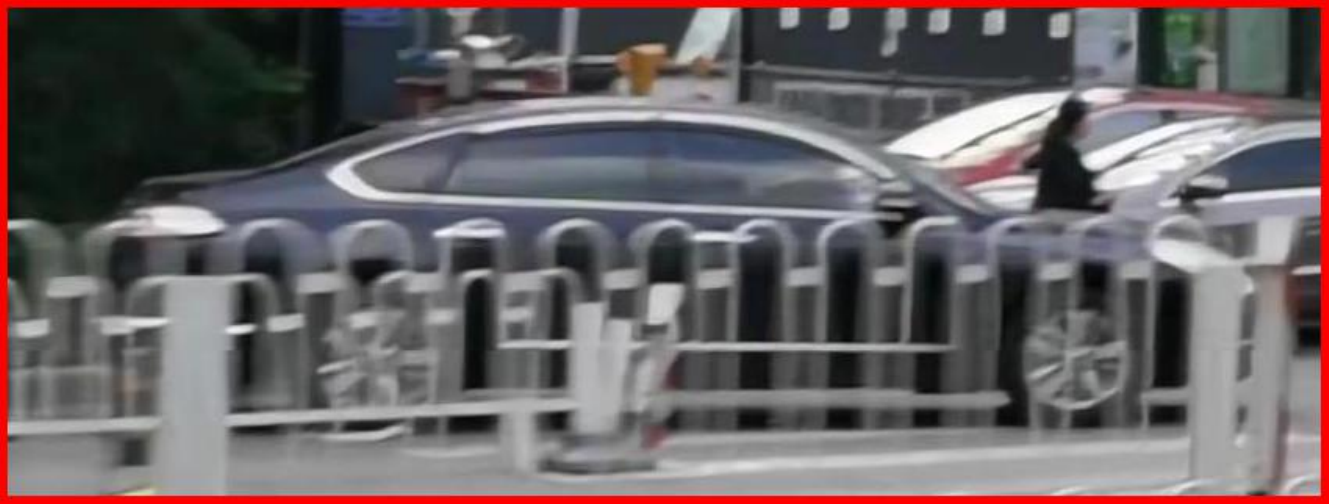}                         &   
							\includegraphics[width = 0.11\textwidth]{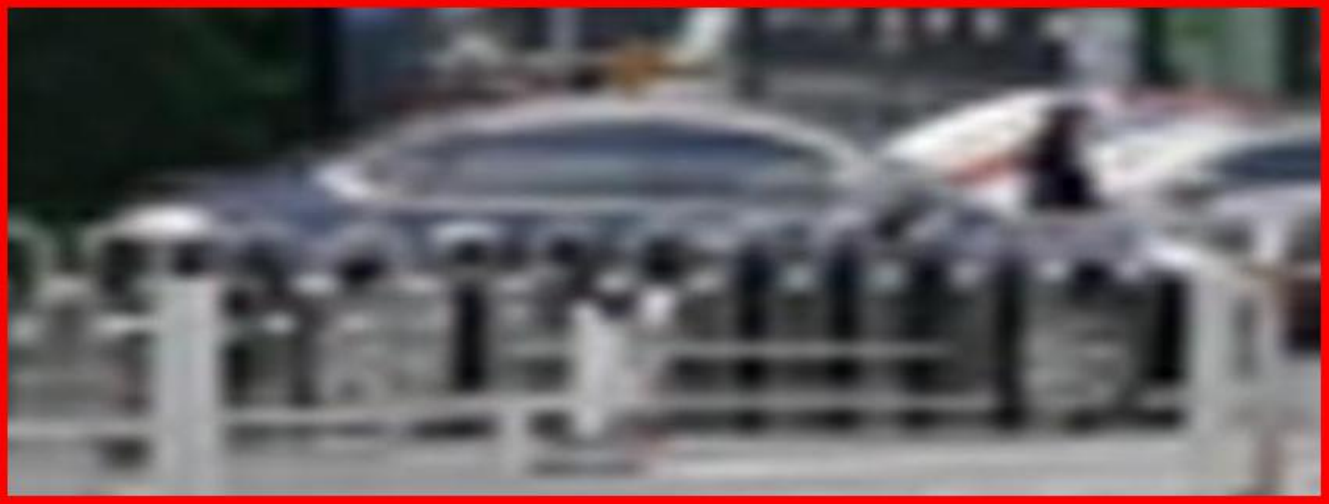}                         &
							\includegraphics[width = 0.11\textwidth]{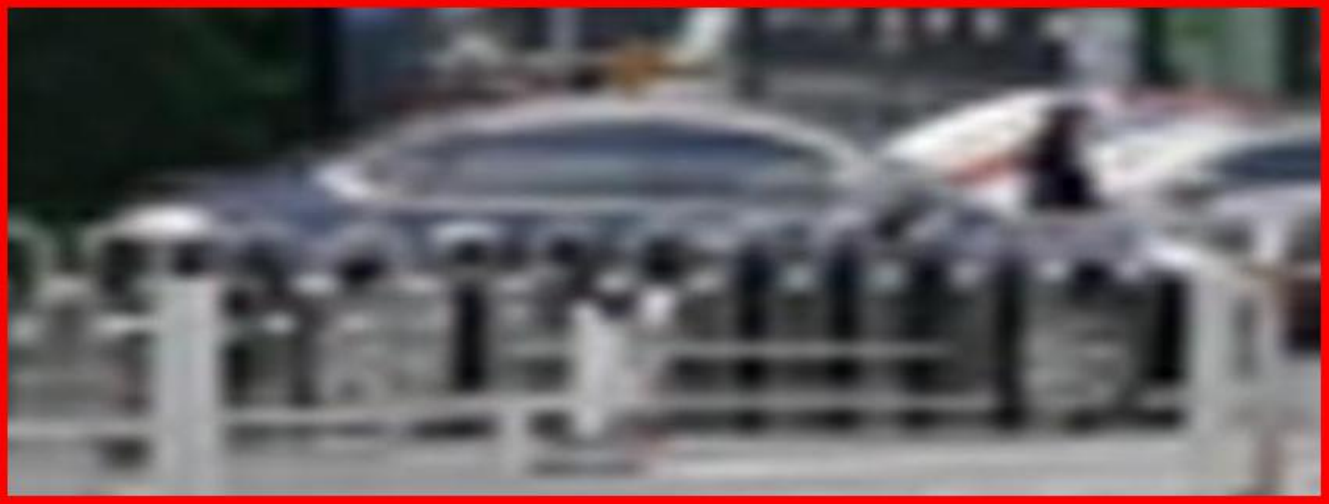}                         &
							\includegraphics[width = 0.11\textwidth]{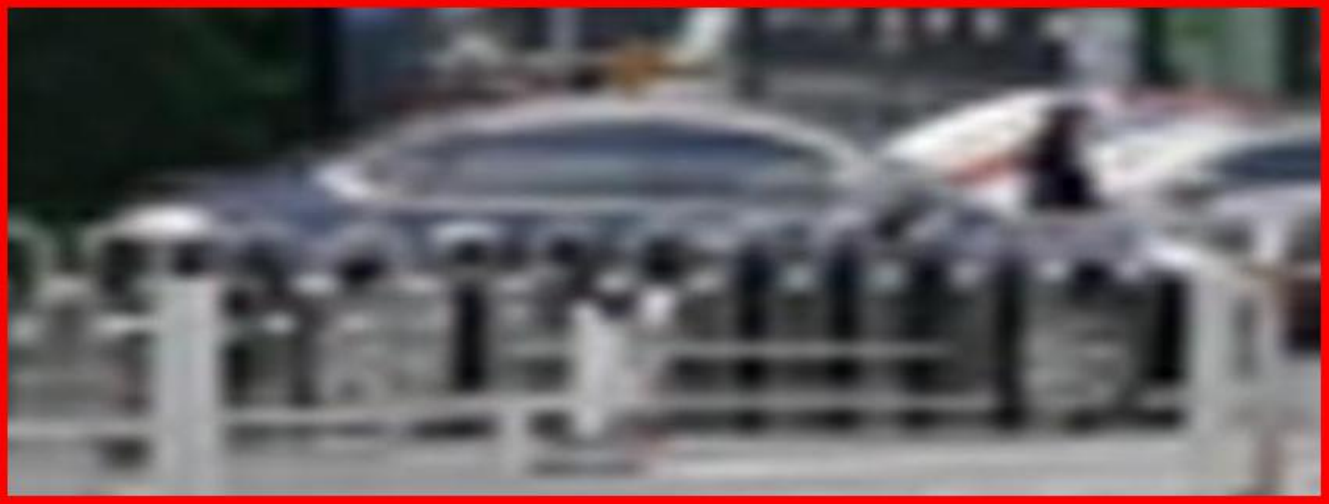}                         & 
							\includegraphics[width = 0.11\textwidth]{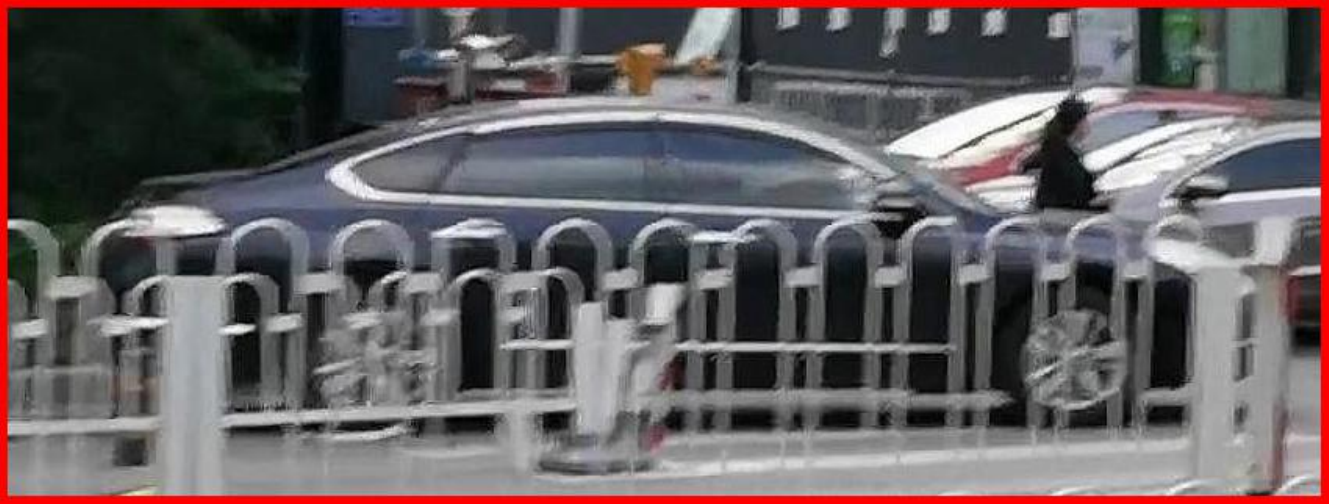}  &
							\includegraphics[width = 0.11\textwidth]{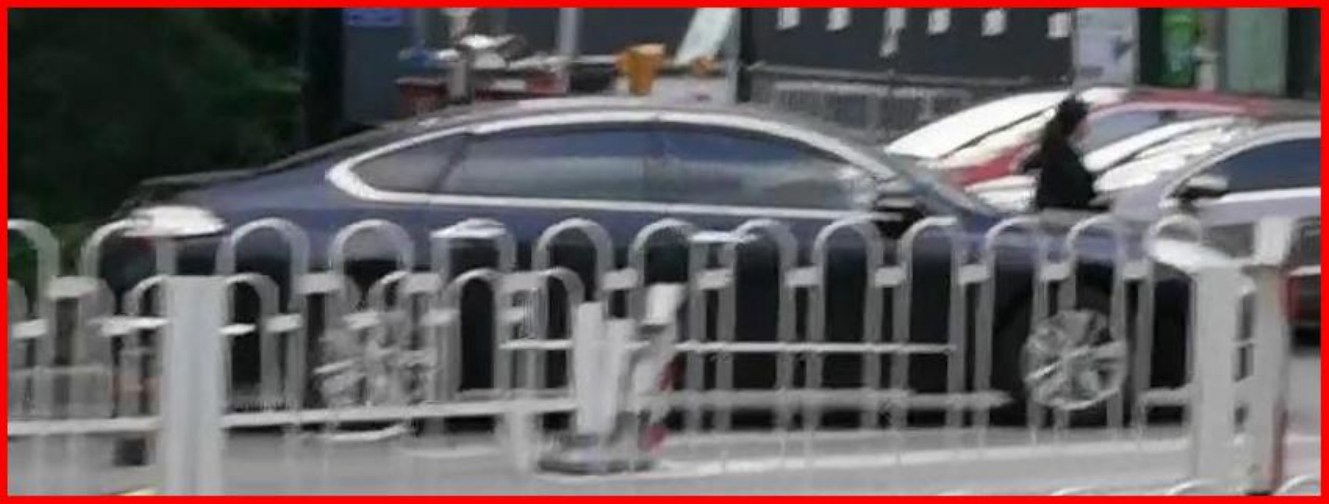}      
							\\
							
							(a)  PSNR/SSIM (4K)   &
							(b)  29.31/0.835     &
							(c)  30.22/0.845    &  
							(d)  29.67/0.847    & 
							(e)  29.35/0.850   & 
							(f)  28.12/0.844   &
							(g)  29.11/0.839   &
							(h)  \textbf{29.69}/\textbf{0.852}    &
							(i)  $+\infty$/1     \\
							
							%	\includegraphics[width = 0.14\textwidth]{plot/ex2/input_74.jpg}                &
							%	\includegraphics[width = 0.14\textwidth]{plot/ex2/output_patch_n.jpg}            &
							%	\includegraphics[width = 0.14\textwidth]{plot/ex2/output_patch_t.jpg}                         &
							%	\includegraphics[width = 0.14\textwidth]{plot/ex2/output_patch_z.jpg}                         &   
							%	\includegraphics[width = 0.14\textwidth]{plot/ex2/s.jpg}                         & 
							%	\includegraphics[width = 0.14\textwidth]{plot/ex2/our.jpg}  &
							%	\includegraphics[width = 0.14\textwidth]{plot/ex2/gt_74.jpg}      
							%	\\
							
							%	\includegraphics[width = 0.14\textwidth]{plot/ex2/crop/input_74_crop.jpg}                &
							%	\includegraphics[width = 0.14\textwidth]{plot/ex2/crop/output_patch_n_crop.jpg}            &
							%	\includegraphics[width = 0.14\textwidth]{plot/ex2/crop/output_patch_t_crop.jpg}                         &
							%	\includegraphics[width = 0.14\textwidth]{plot/ex2/crop/output_patch_z_crop.jpg}                         &   
							%	\includegraphics[width = 0.14\textwidth]{plot/ex2/crop/s_crop.jpg}                         & 
							%	\includegraphics[width = 0.14\textwidth]{plot/ex2/crop/our_crop.jpg}  &
							%	\includegraphics[width = 0.14\textwidth]{plot/ex2/crop/gt_74_crop.jpg}      
							%	\\
							
							%	(a)  PSNR/SSIM   &
							%	(b)  23.24/0.62     &
							%	(c)  24.80/0.66     &  
							%	(d)  23.37/0.61    & 
							%	(e)  22.54/0.59    & 
							%	(f)  \textbf{27.9/0.77}    &
							%	(g)  $+\infty$/1     \\

							\includegraphics[width=0.11\textwidth]{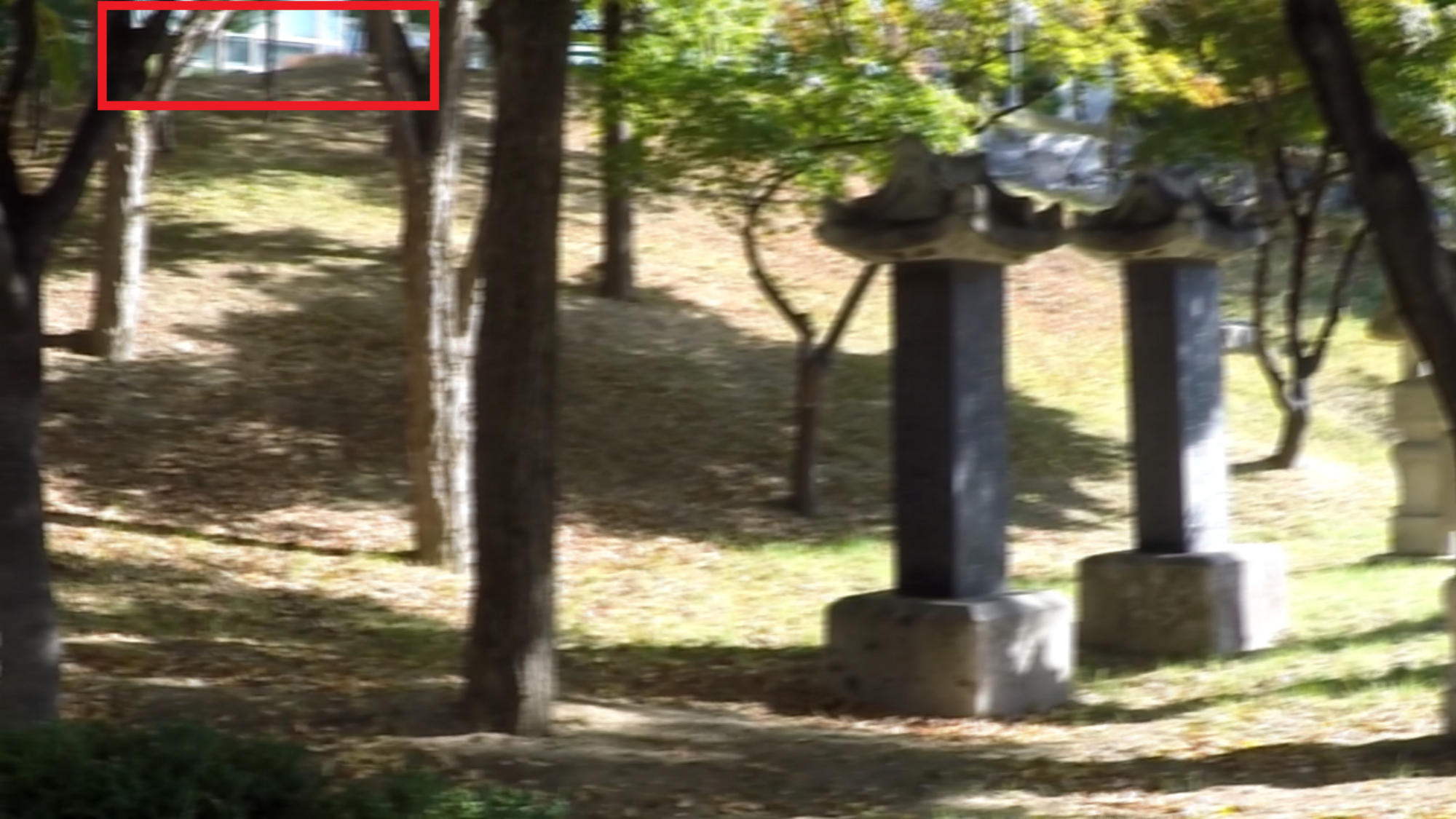}  &
							\includegraphics[width=0.11\textwidth]{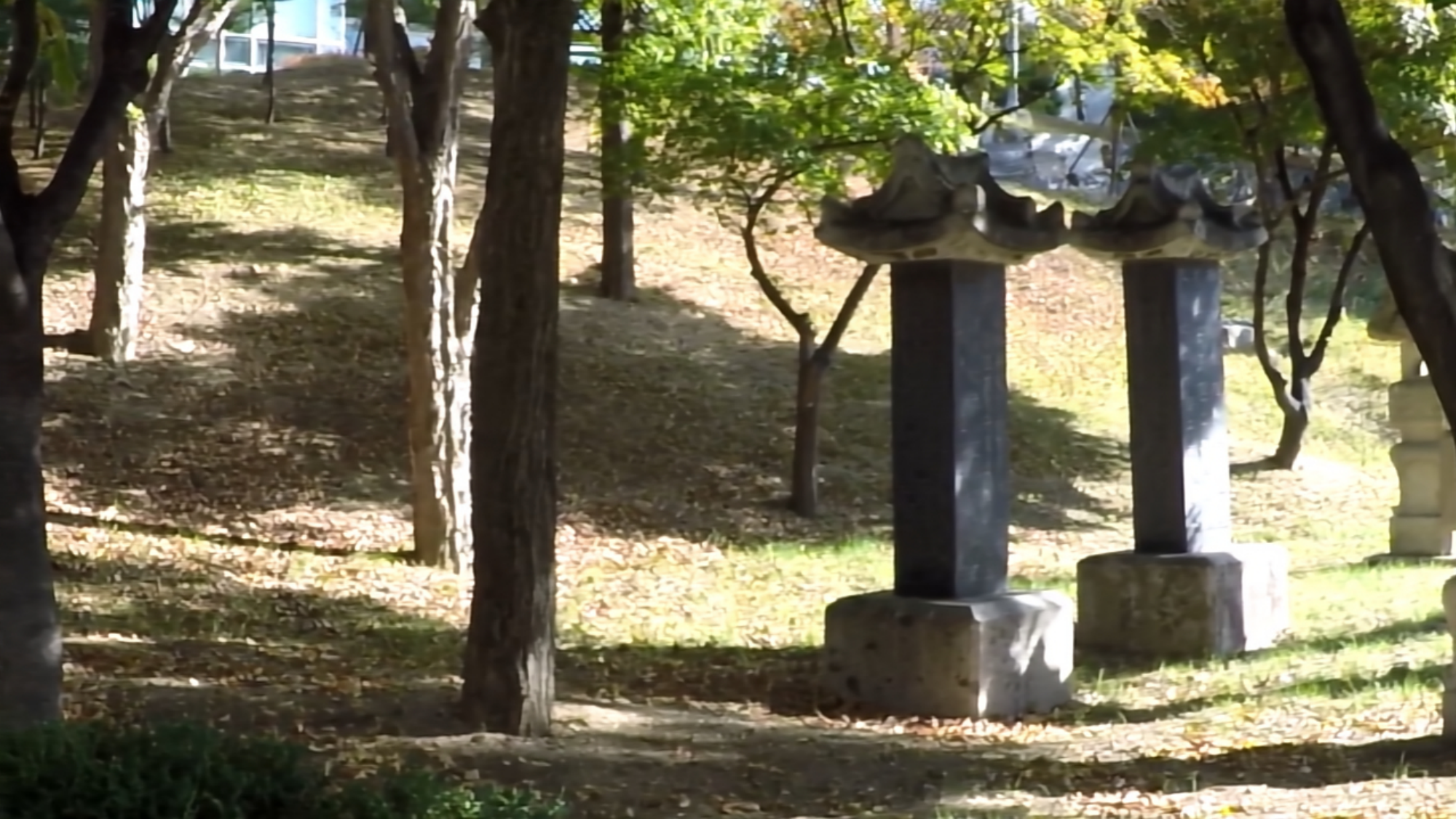}   &
							\includegraphics[width=0.11\textwidth]{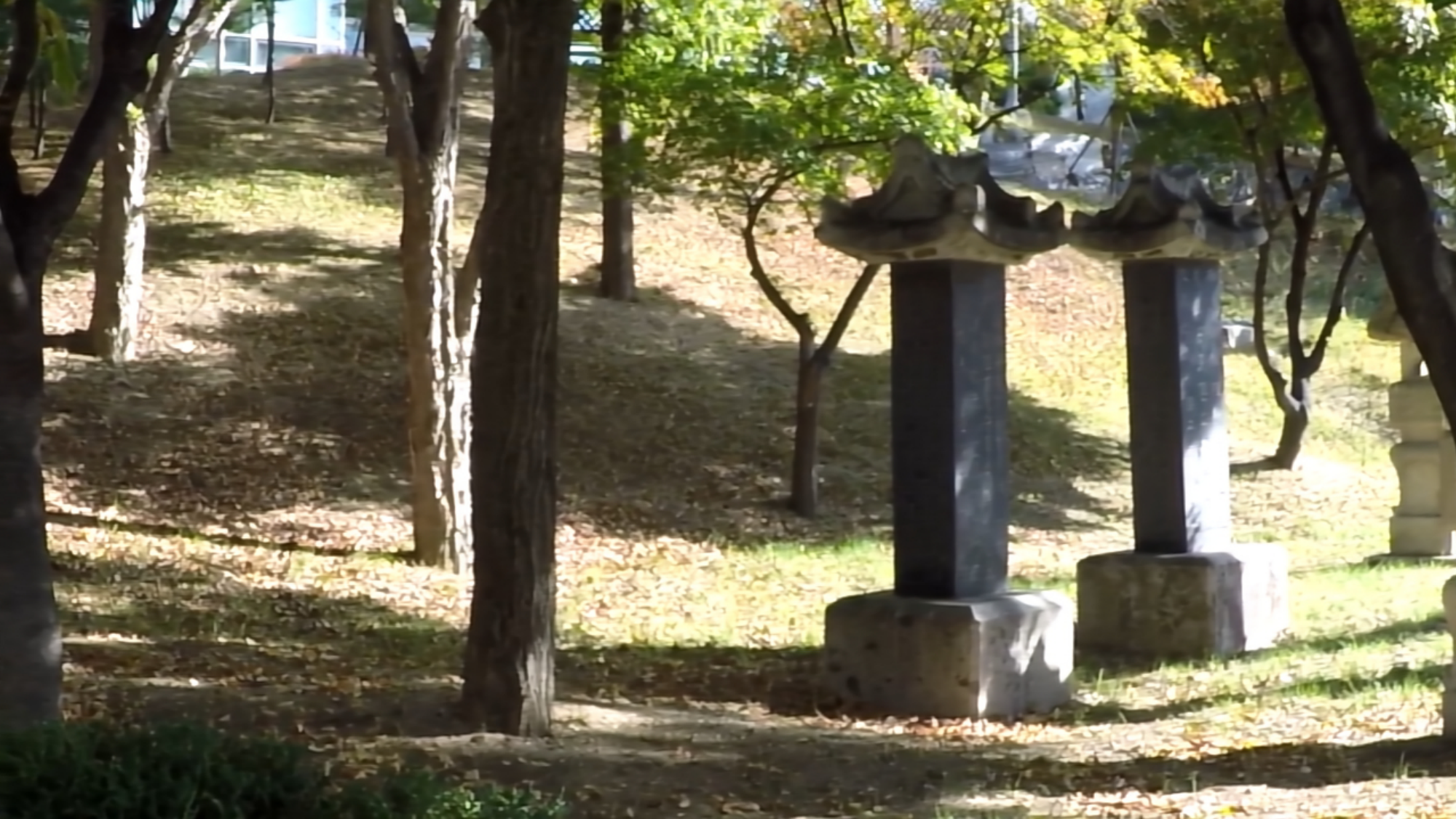}  &
							\includegraphics[width=0.11\textwidth]{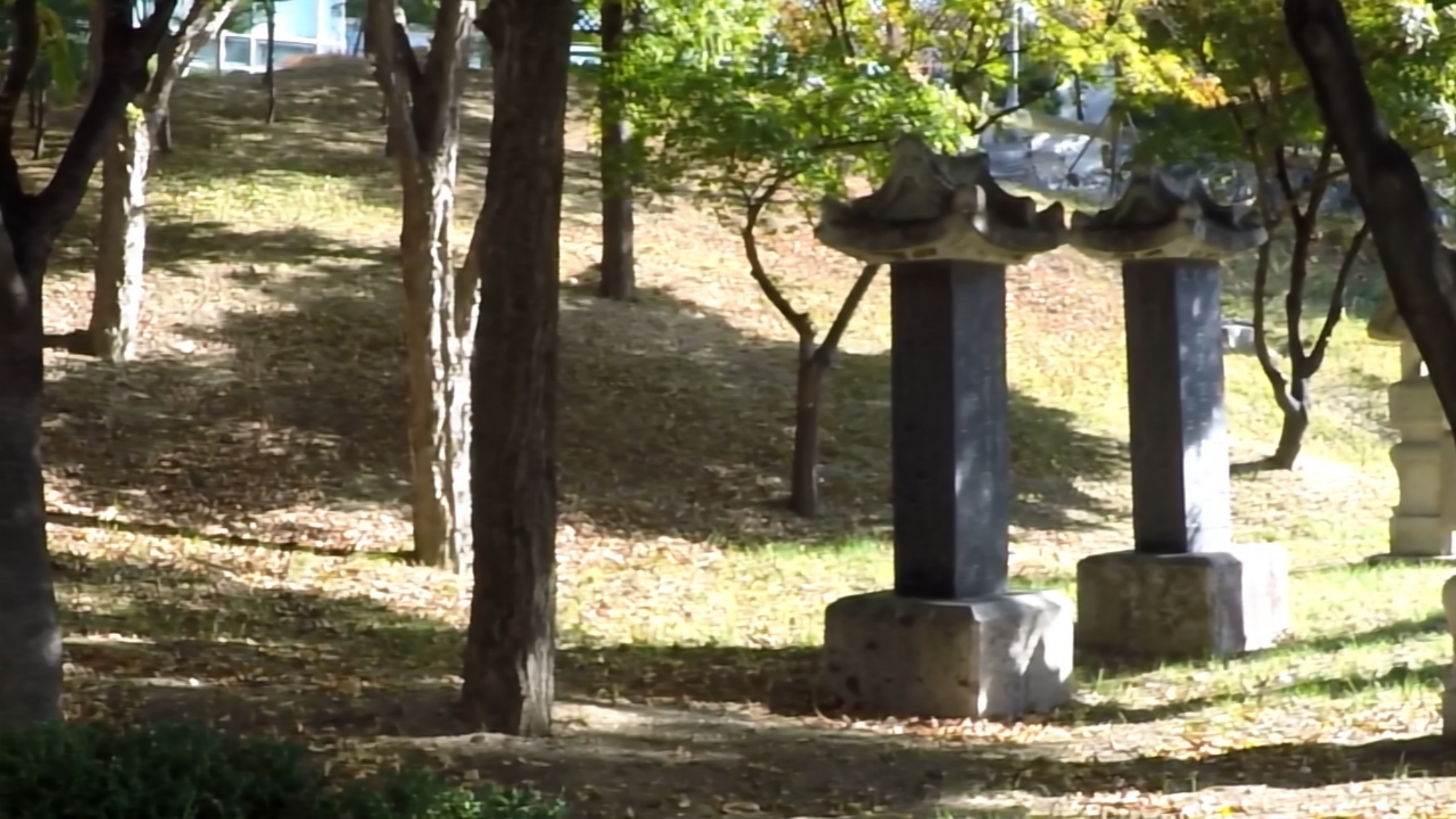}      &   
							\includegraphics[width=0.11\textwidth]{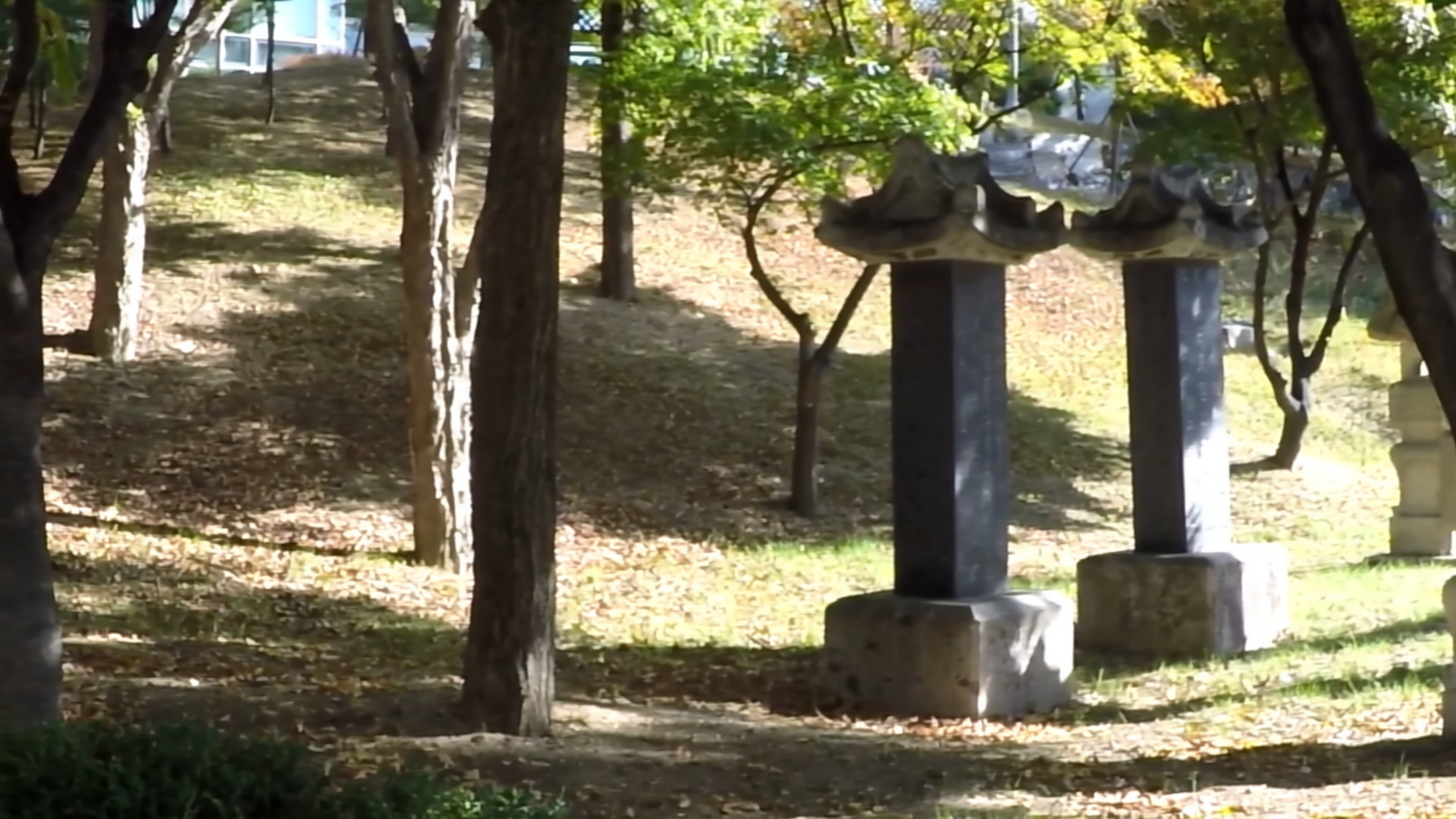}      & 
							\includegraphics[width=0.11\textwidth]{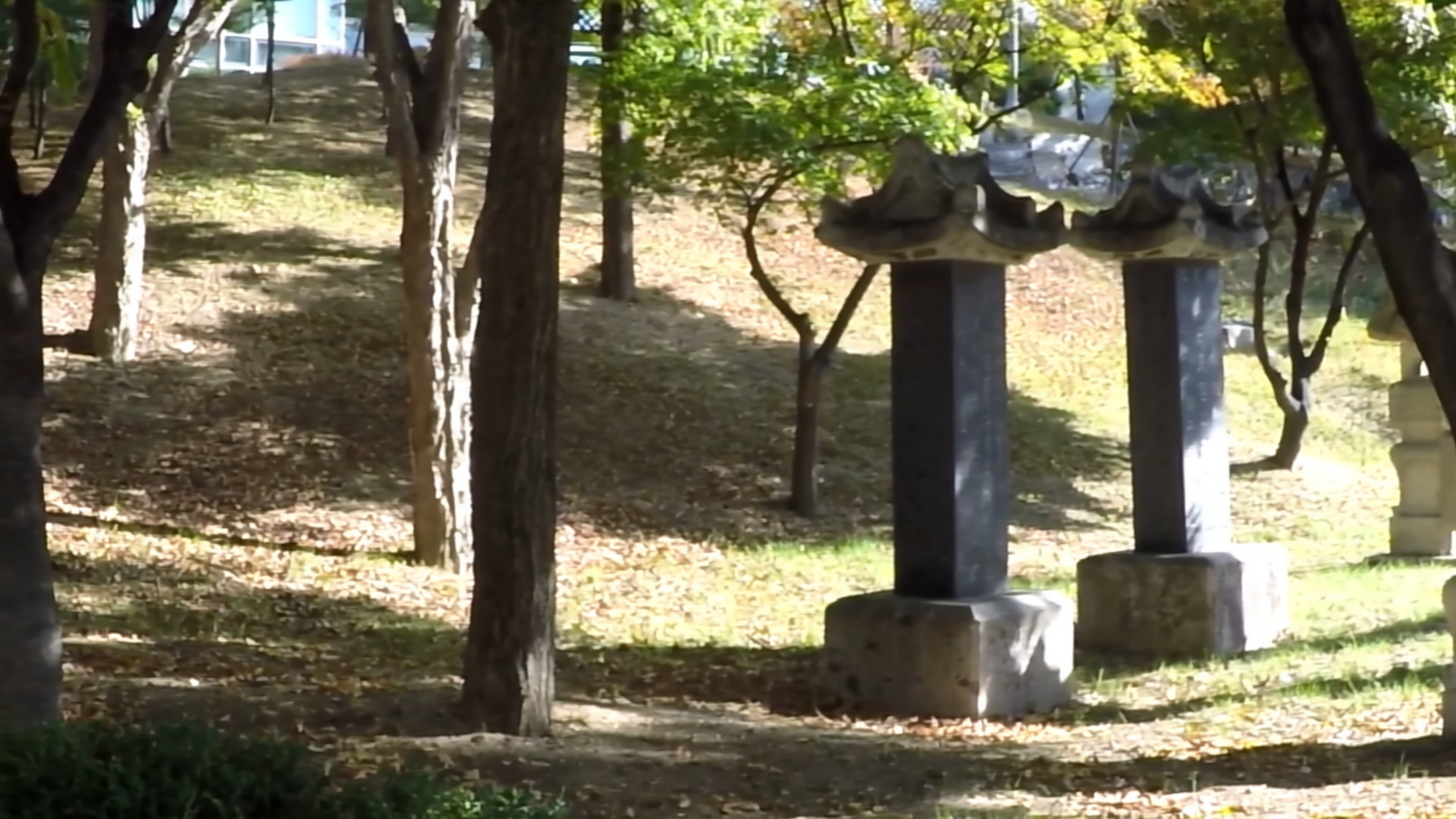}      & 
							\includegraphics[width=0.11\textwidth]{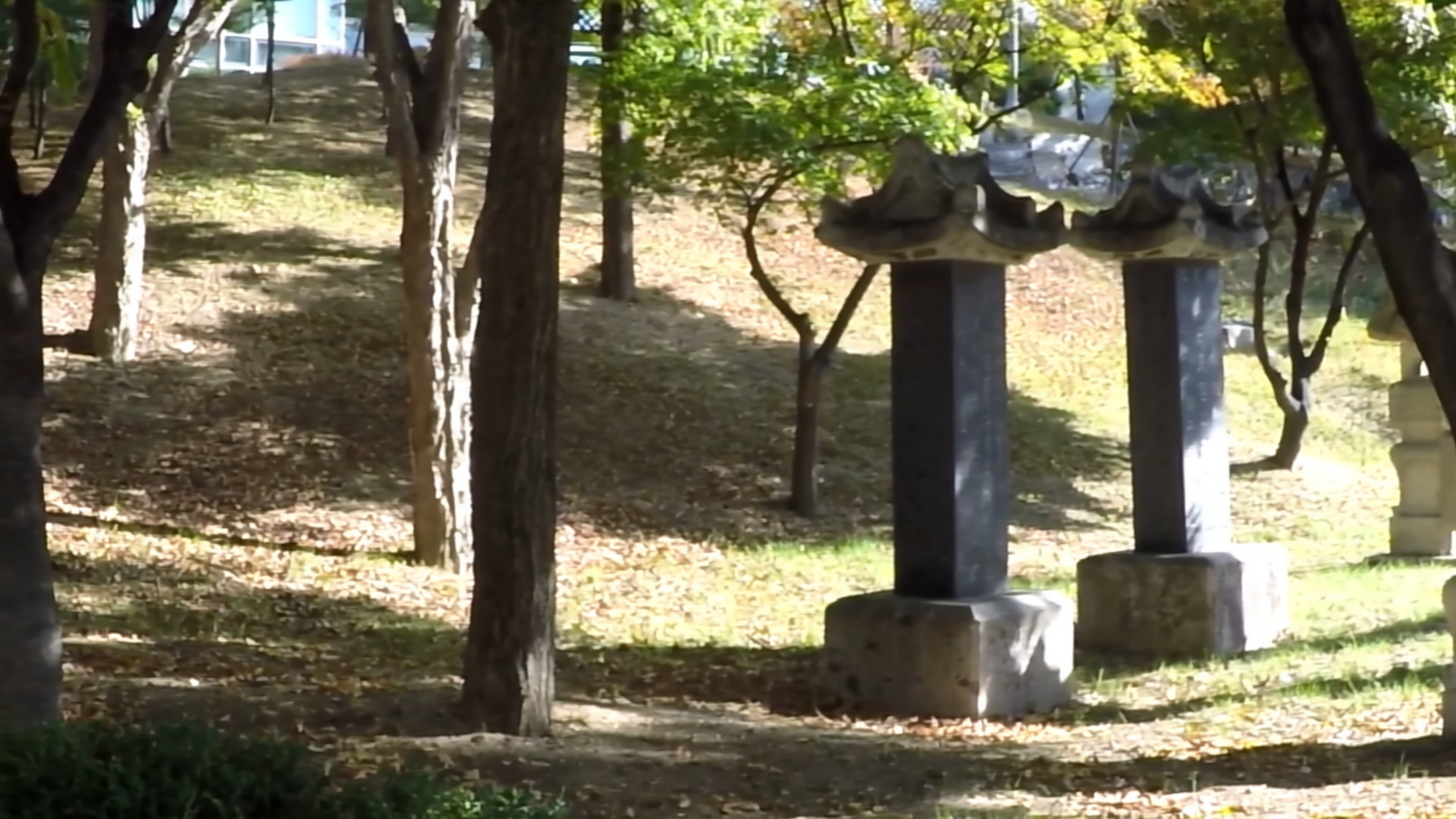}      & 
							\includegraphics[width = 0.11\textwidth]{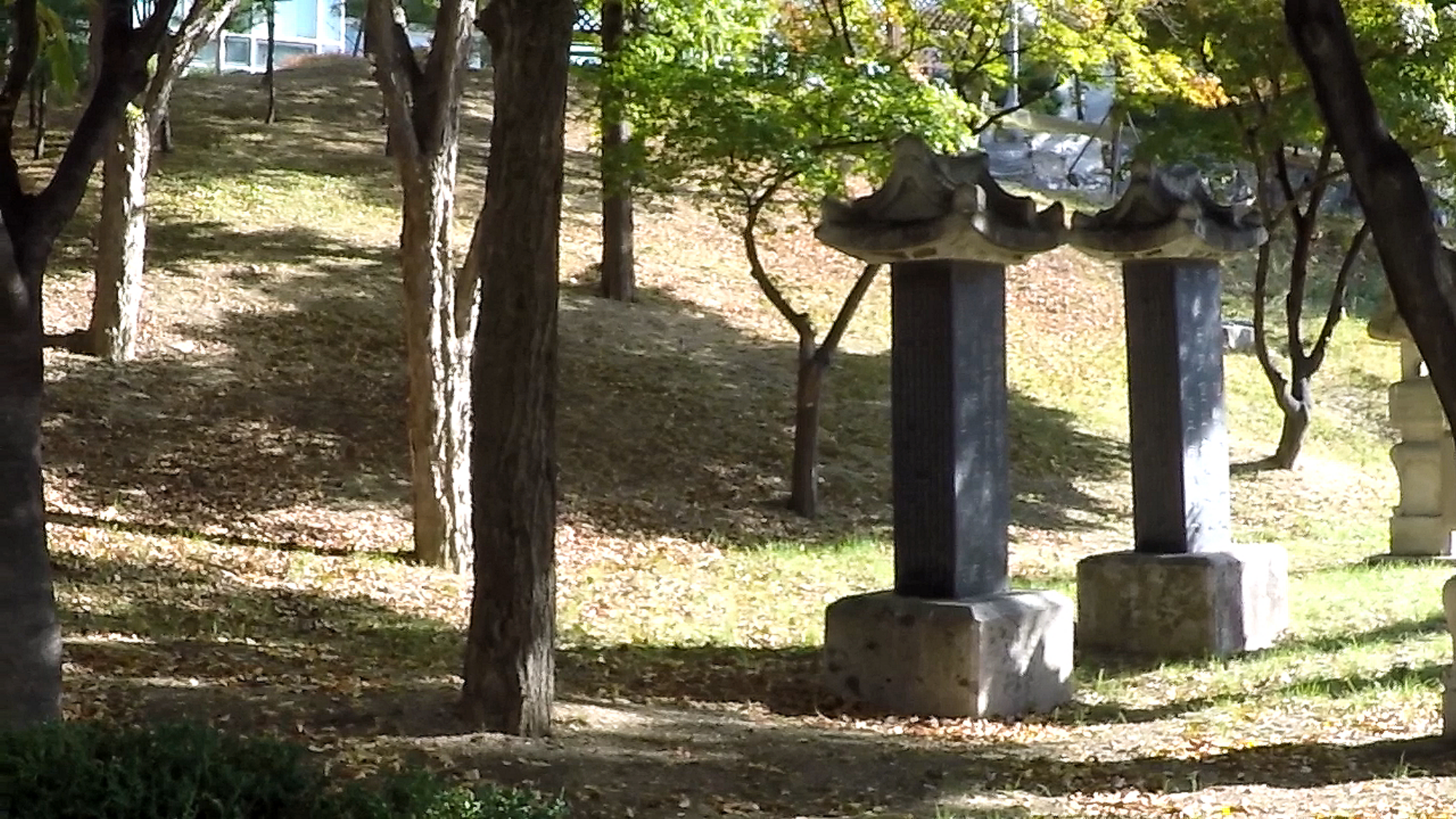}  &
							\includegraphics[width = 0.11\textwidth]{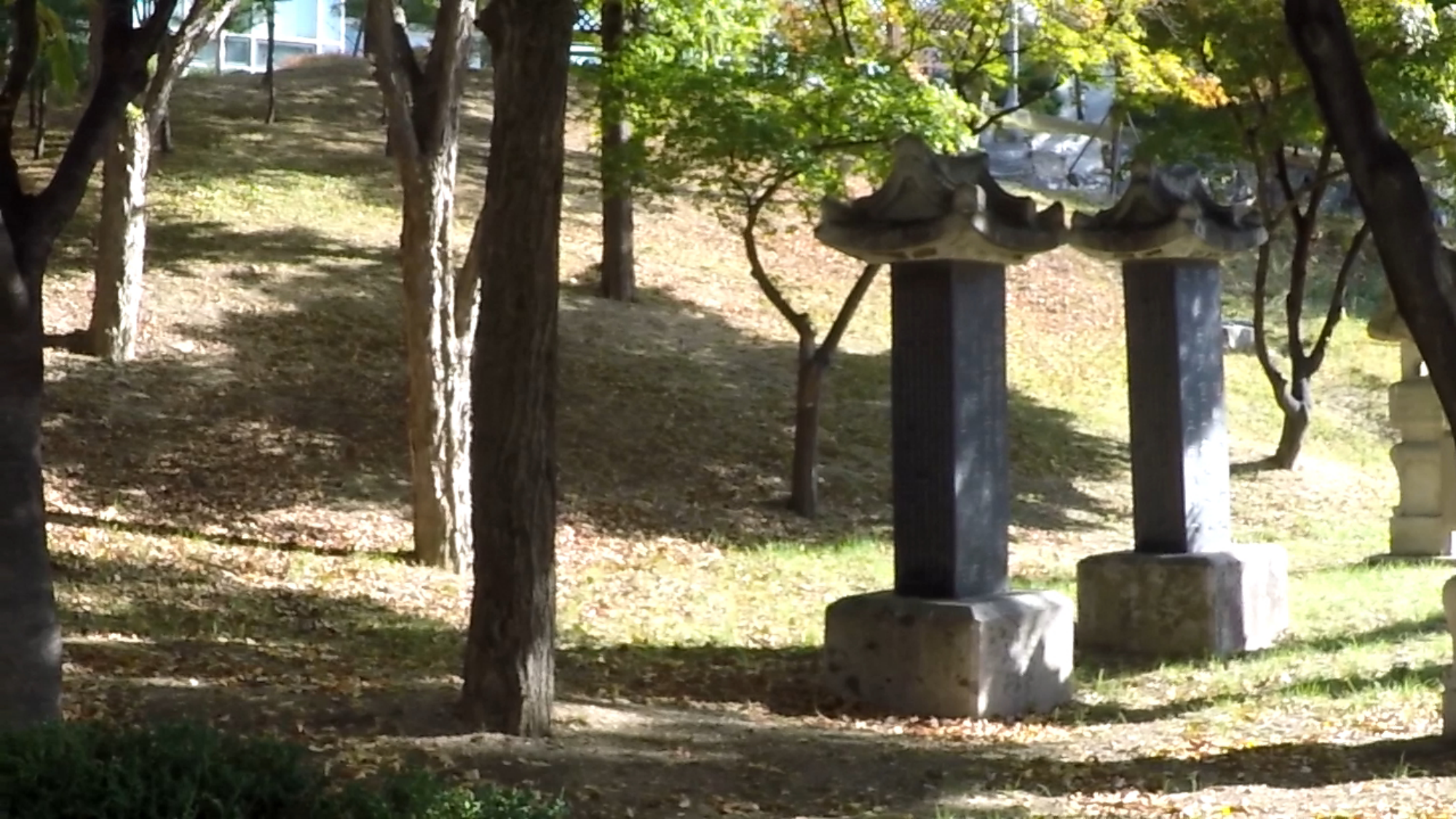}     \\
							
							\includegraphics[width=0.11\textwidth]{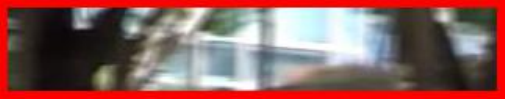}  &
							\includegraphics[width=0.11\textwidth]{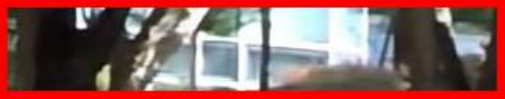}  &
							\includegraphics[width=0.11\textwidth]{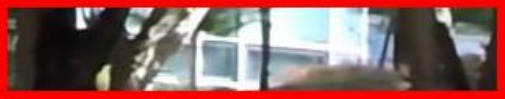}  &
							\includegraphics[width=0.11\textwidth]{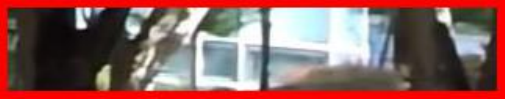}  &
							\includegraphics[width=0.11\textwidth]{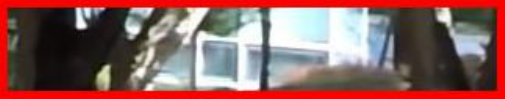}  &
							\includegraphics[width=0.11\textwidth]{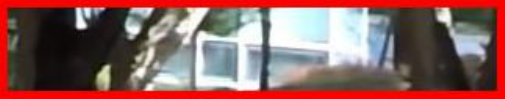}  &
							\includegraphics[width=0.11\textwidth]{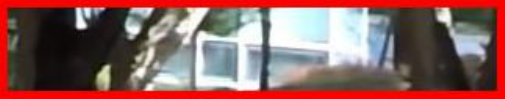}  &
							\includegraphics[width=0.11\textwidth]{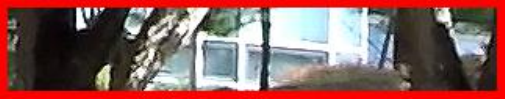}  &				
							\includegraphics[width=0.11\textwidth]{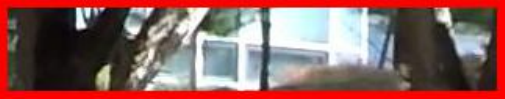}  \\
							
							(a)  PSNR/SSIM (G-P)      &
							(b)  \textbf{33.33}/0.965     &
							(c)  32.98/0.960     &  
							(d)  33.31/0.964     & 
							(e)  32.86/0.957     & 
							(f)  31.15/0.943     & 
							(g)  32.77/0.920     & 
							(h)  33.31/\textbf{0.968}    &
							(i)  $+\infty$/1     \\
							
							\includegraphics[width = 0.11\textwidth]{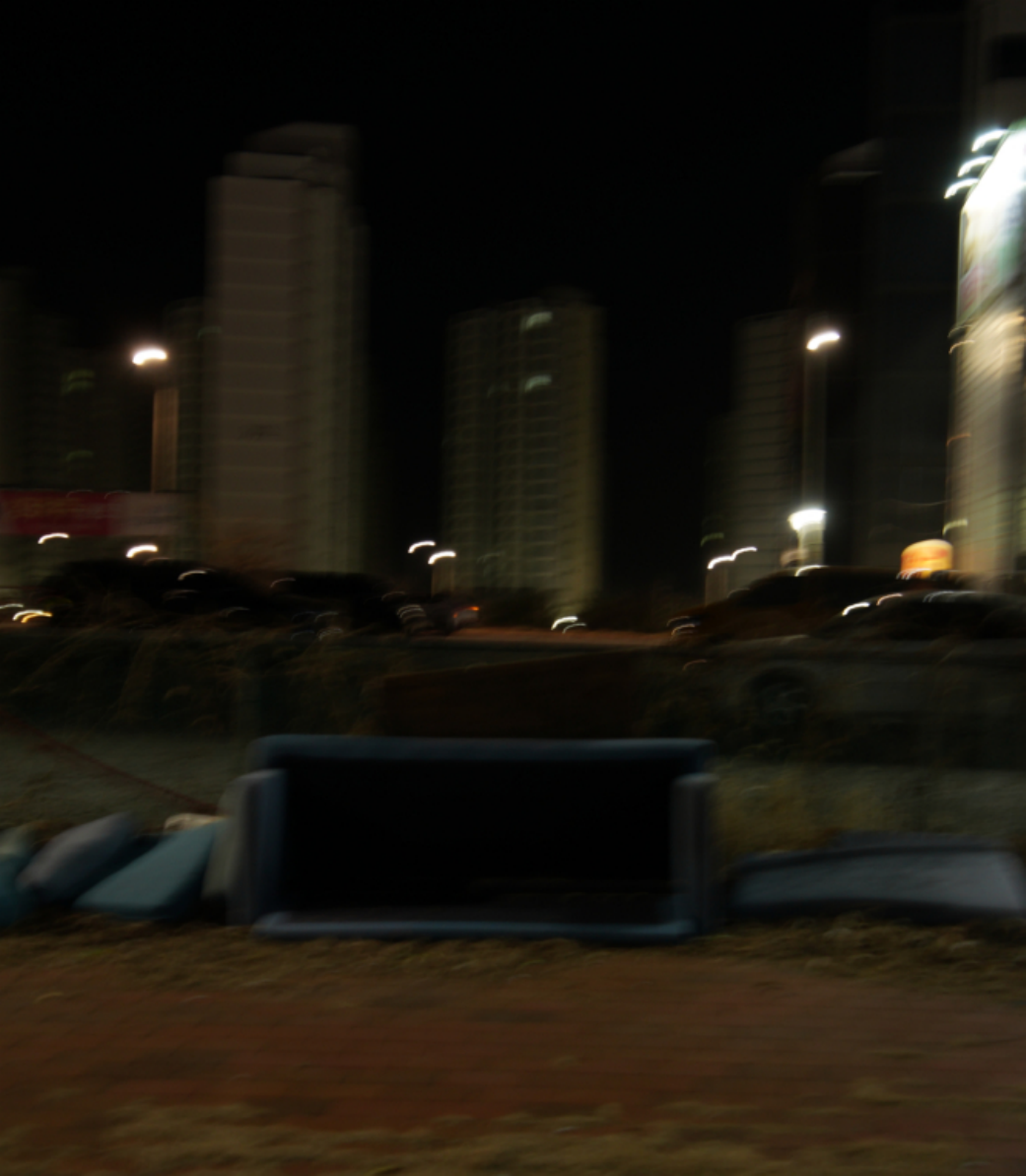}                &
							\includegraphics[width = 0.11\textwidth]{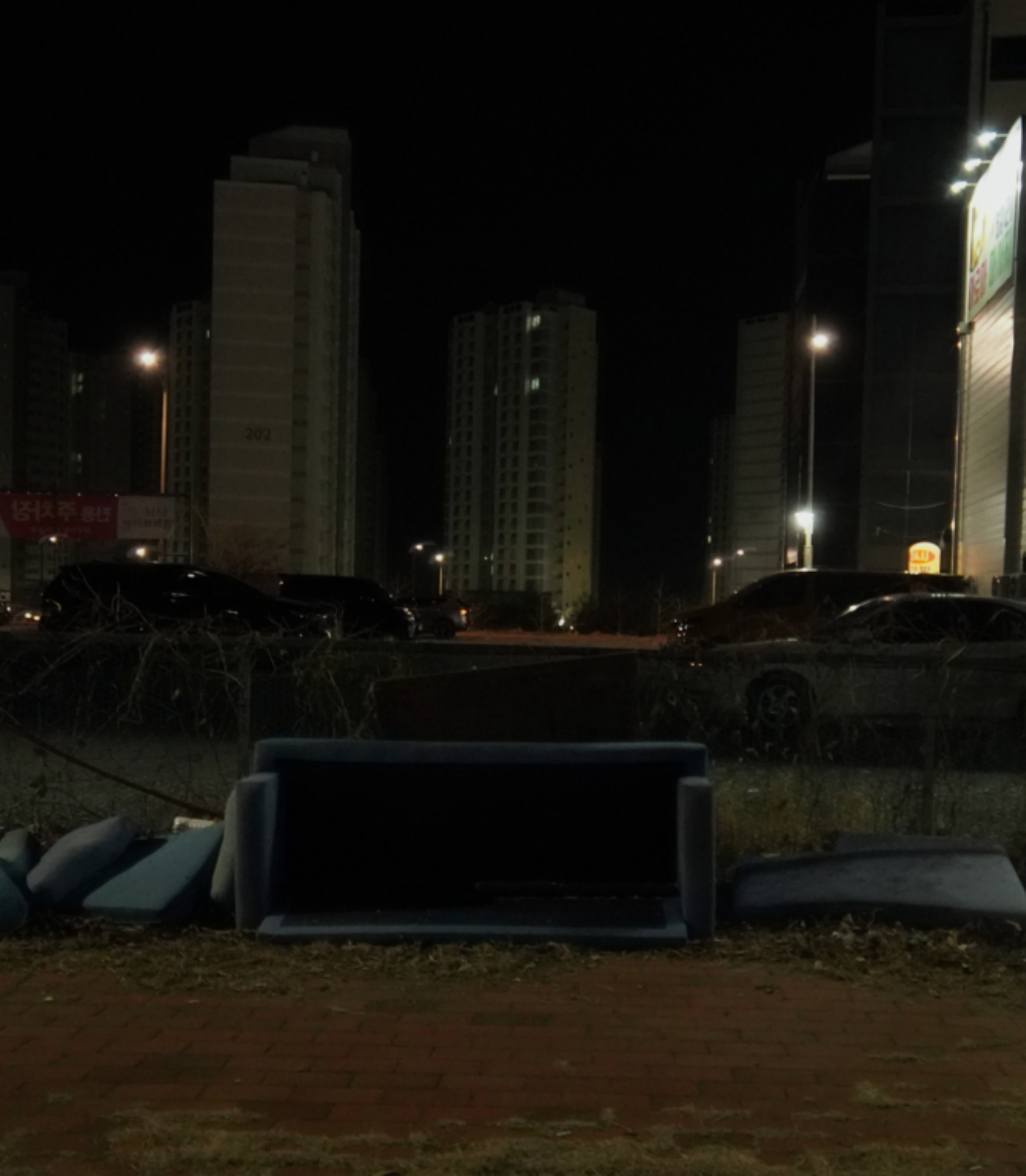}            &
							\includegraphics[width = 0.11\textwidth]{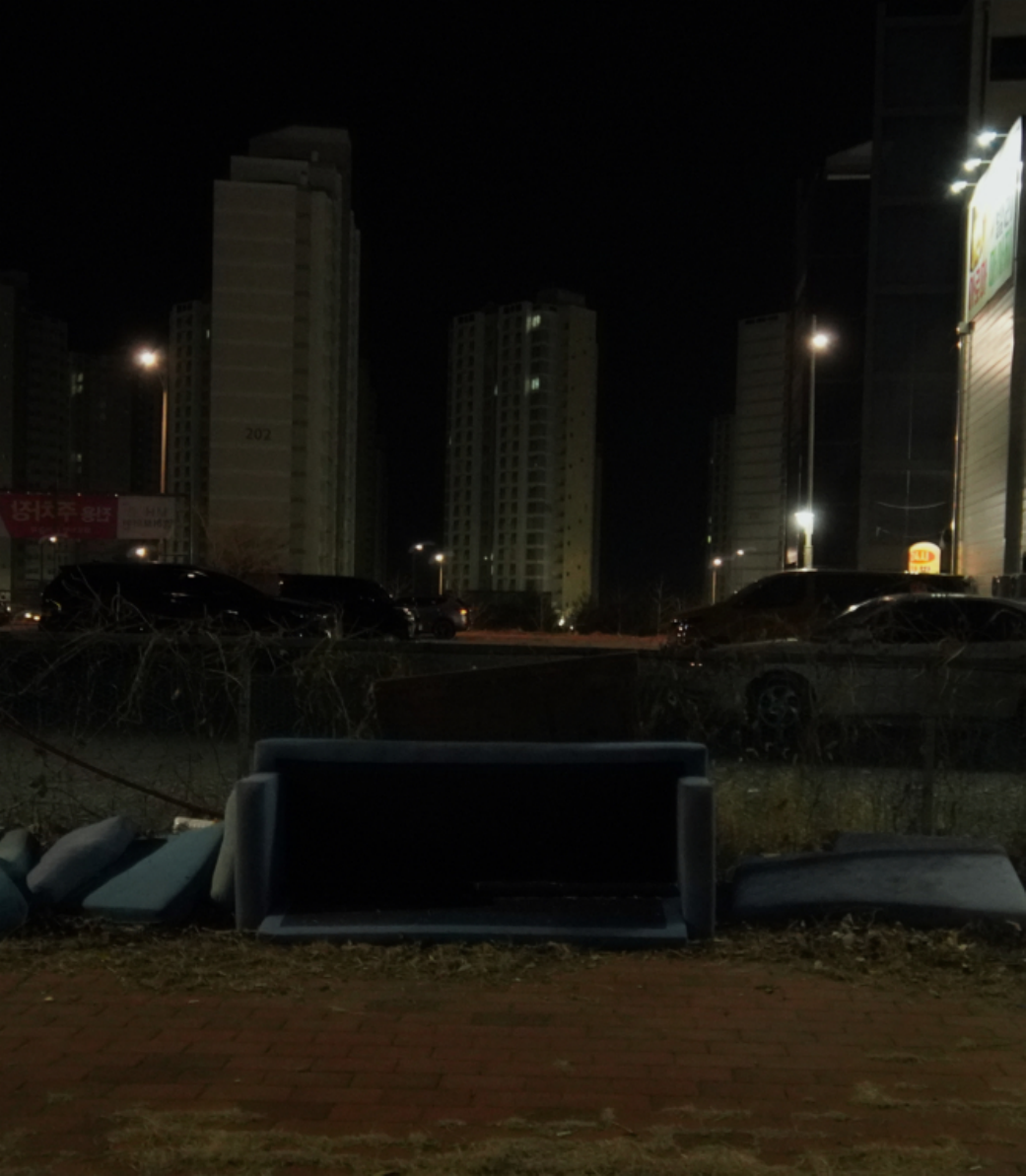}                         &
							\includegraphics[width = 0.11\textwidth]{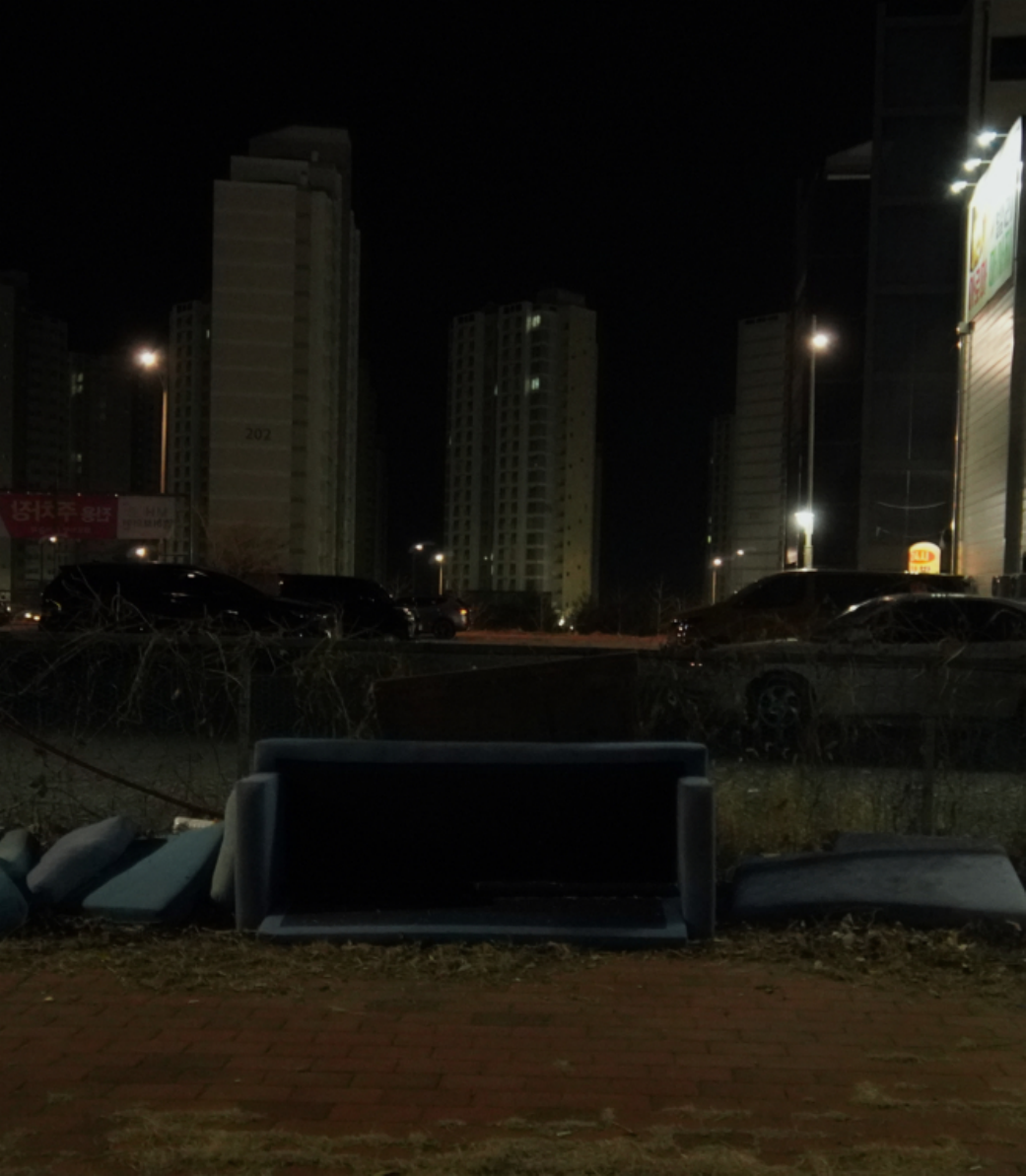}                         &   
							\includegraphics[width = 0.11\textwidth]{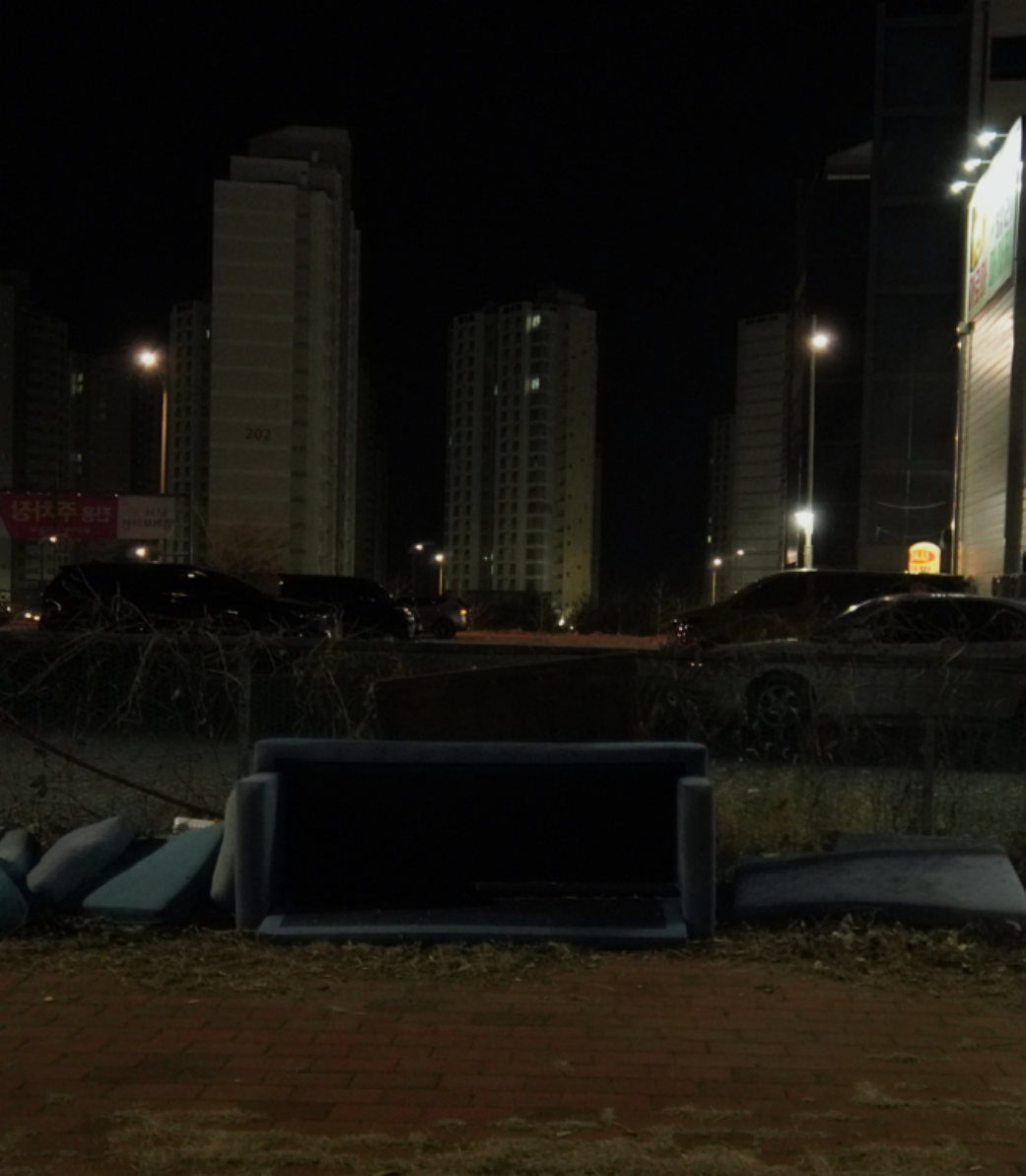}                         & 
							\includegraphics[width = 0.11\textwidth]{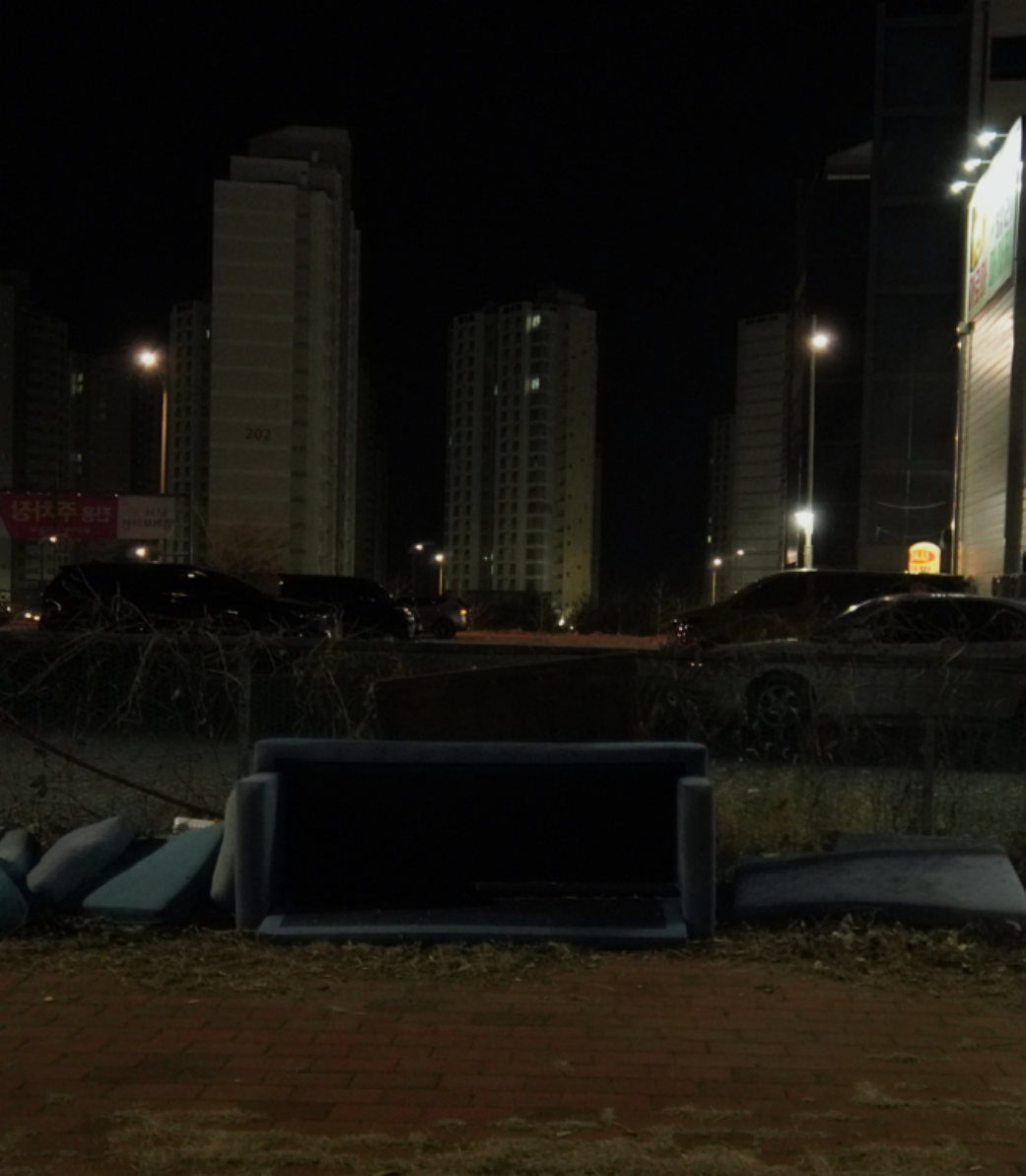}                         &
							\includegraphics[width = 0.11\textwidth]{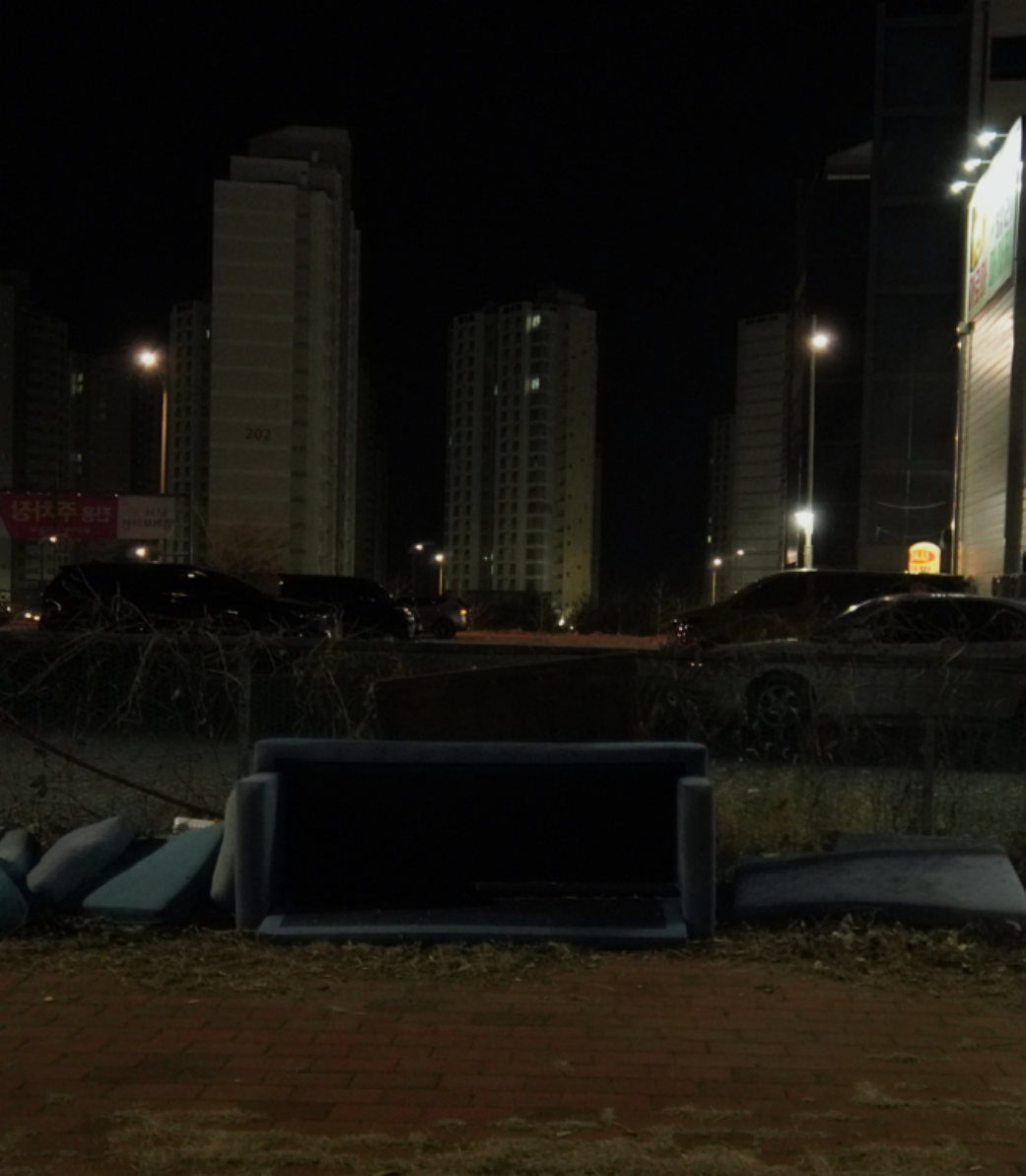}                         &
							\includegraphics[width = 0.11\textwidth]{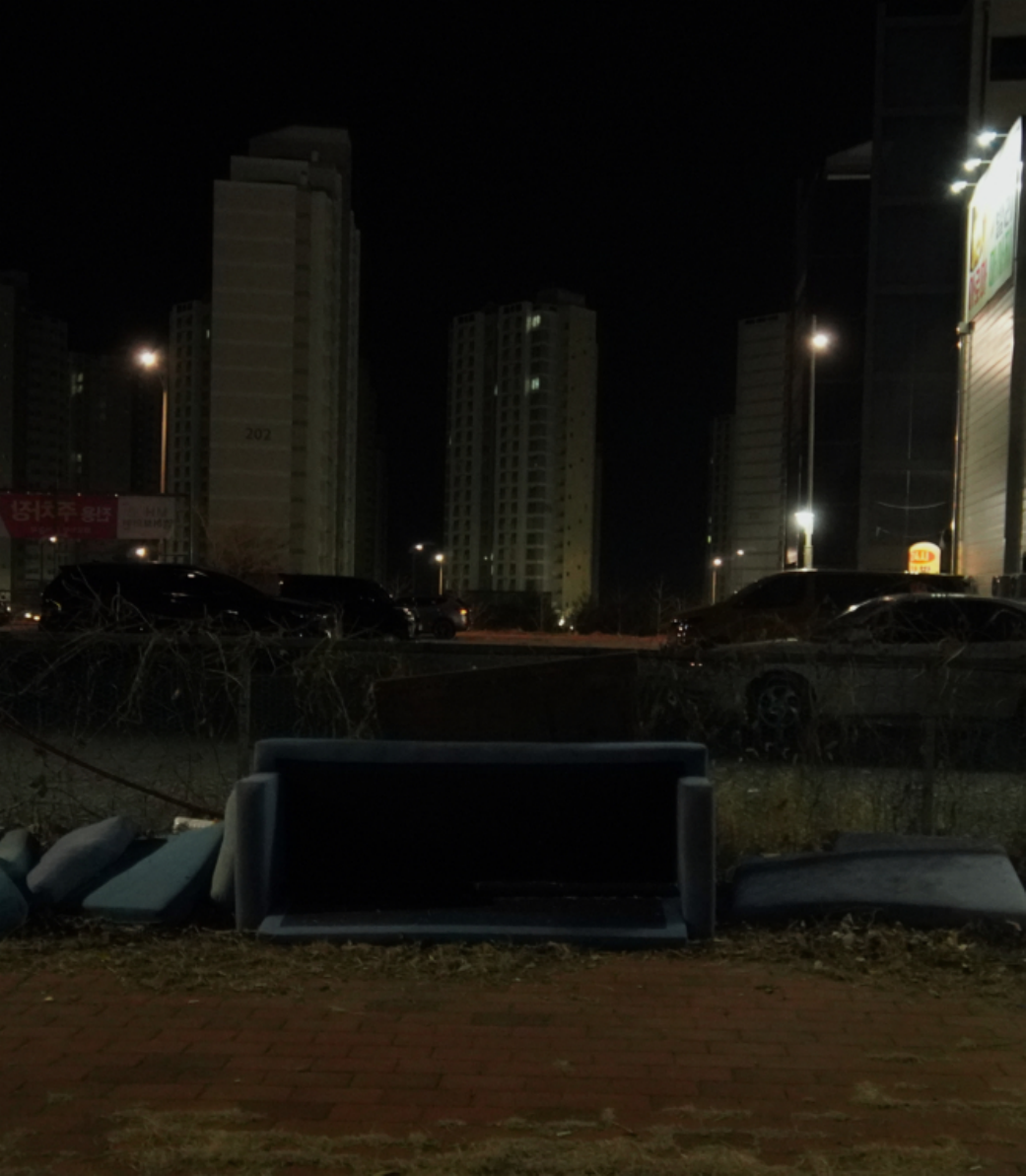}  &
							\includegraphics[width = 0.11\textwidth]{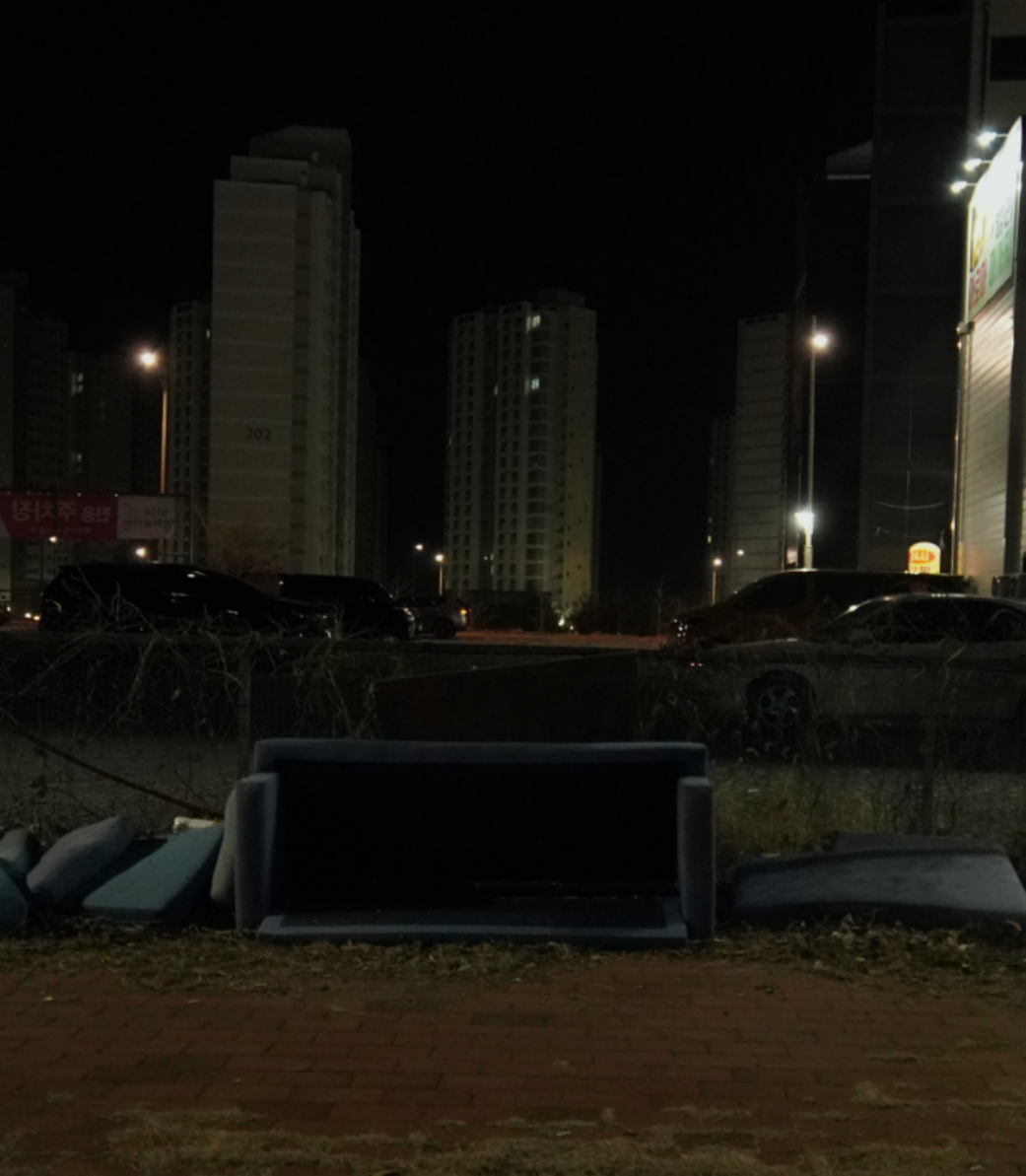}      
							\\
							
							\includegraphics[width = 0.11\textwidth]{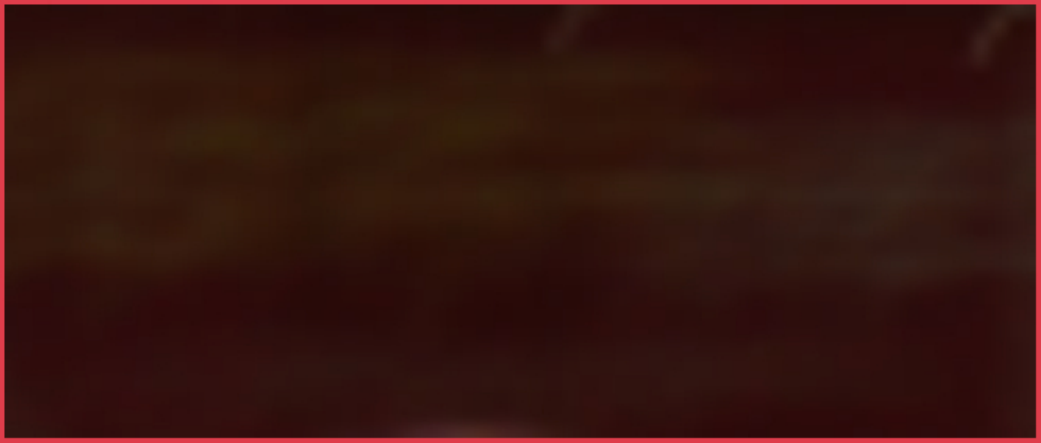}                &
							\includegraphics[width = 0.11\textwidth]{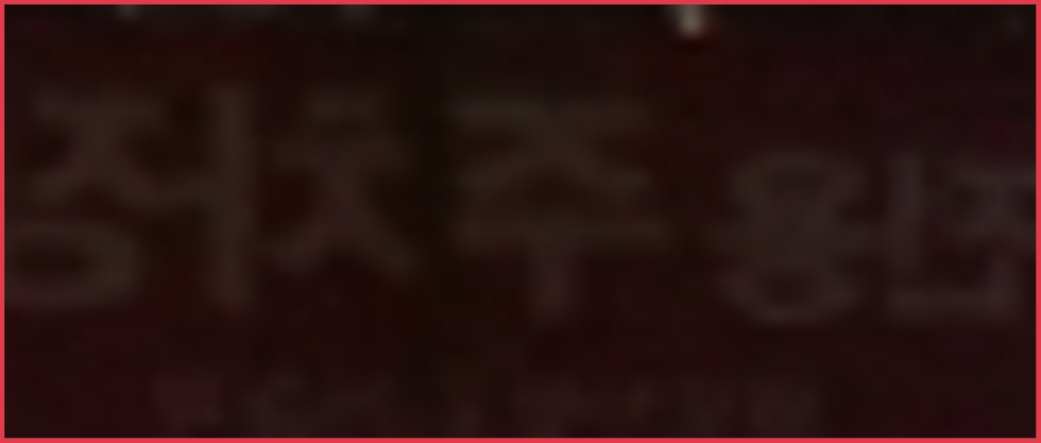}            &
							\includegraphics[width = 0.11\textwidth]{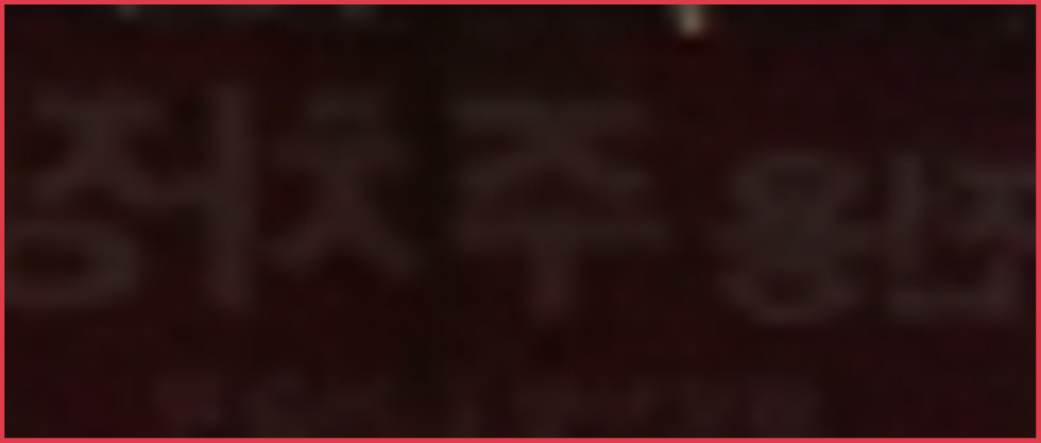}                         &
							\includegraphics[width = 0.11\textwidth]{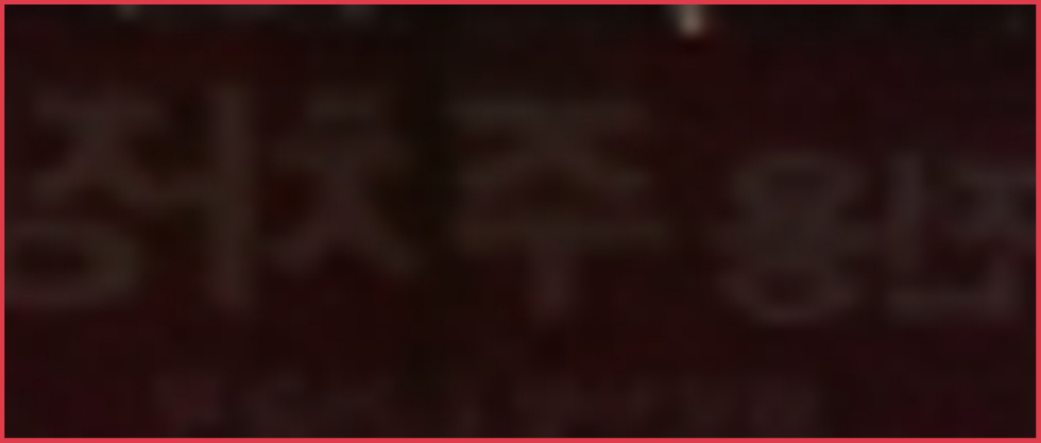}                         &   
							\includegraphics[width = 0.11\textwidth]{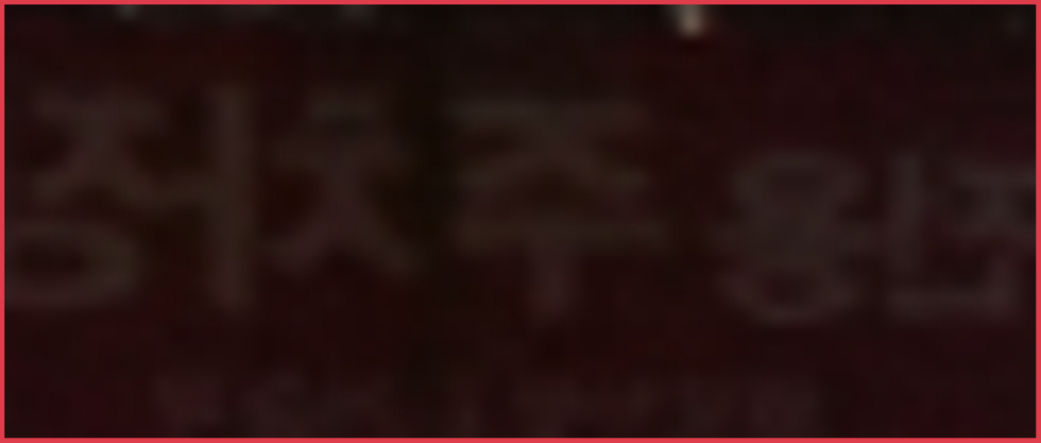}                         & 
							\includegraphics[width = 0.11\textwidth]{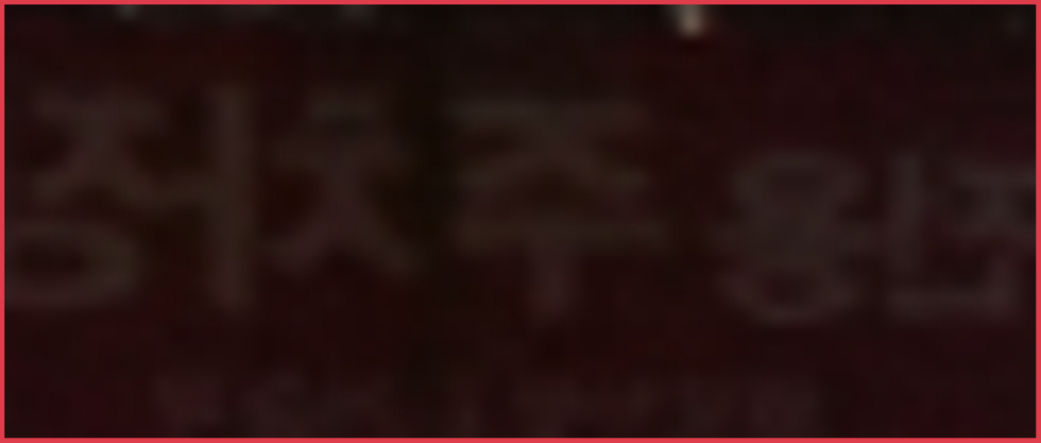}                         &
							\includegraphics[width = 0.11\textwidth]{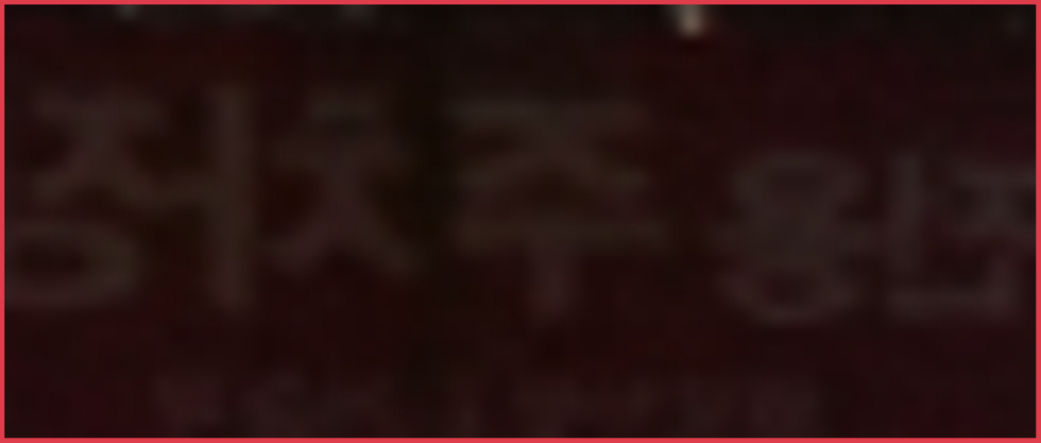}                         &
							\includegraphics[width = 0.11\textwidth]{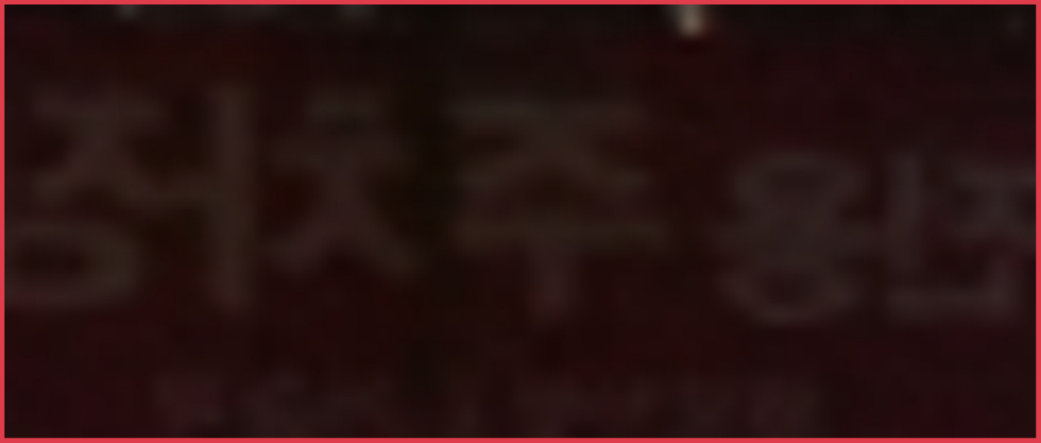}  &
							\includegraphics[width = 0.11\textwidth]{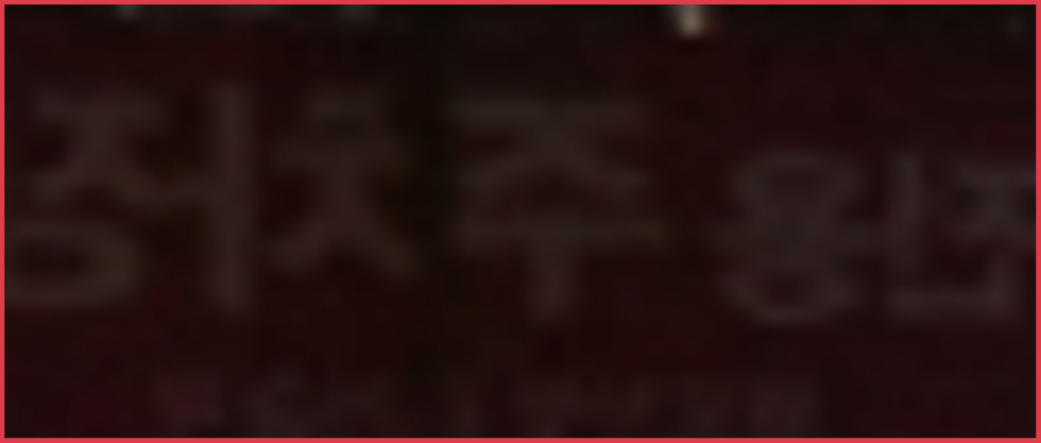}          
							\\
							
							(a) PSNR/SSIM (R-J)  &
							(b) 28.77/0.870     &
							(c) 28.95/0.878     &  
							(d) 28.61/0.866    & 
							(e) 32.49/\textbf{0.930}    & 
							(f) 32.15/0.899    & 
							(g) 30.22/0.910   & 
							(h) \textbf{32.50}/0.927    &
							(i) ($+\infty$/1)     \\

							\includegraphics[width = 0.11\textwidth]{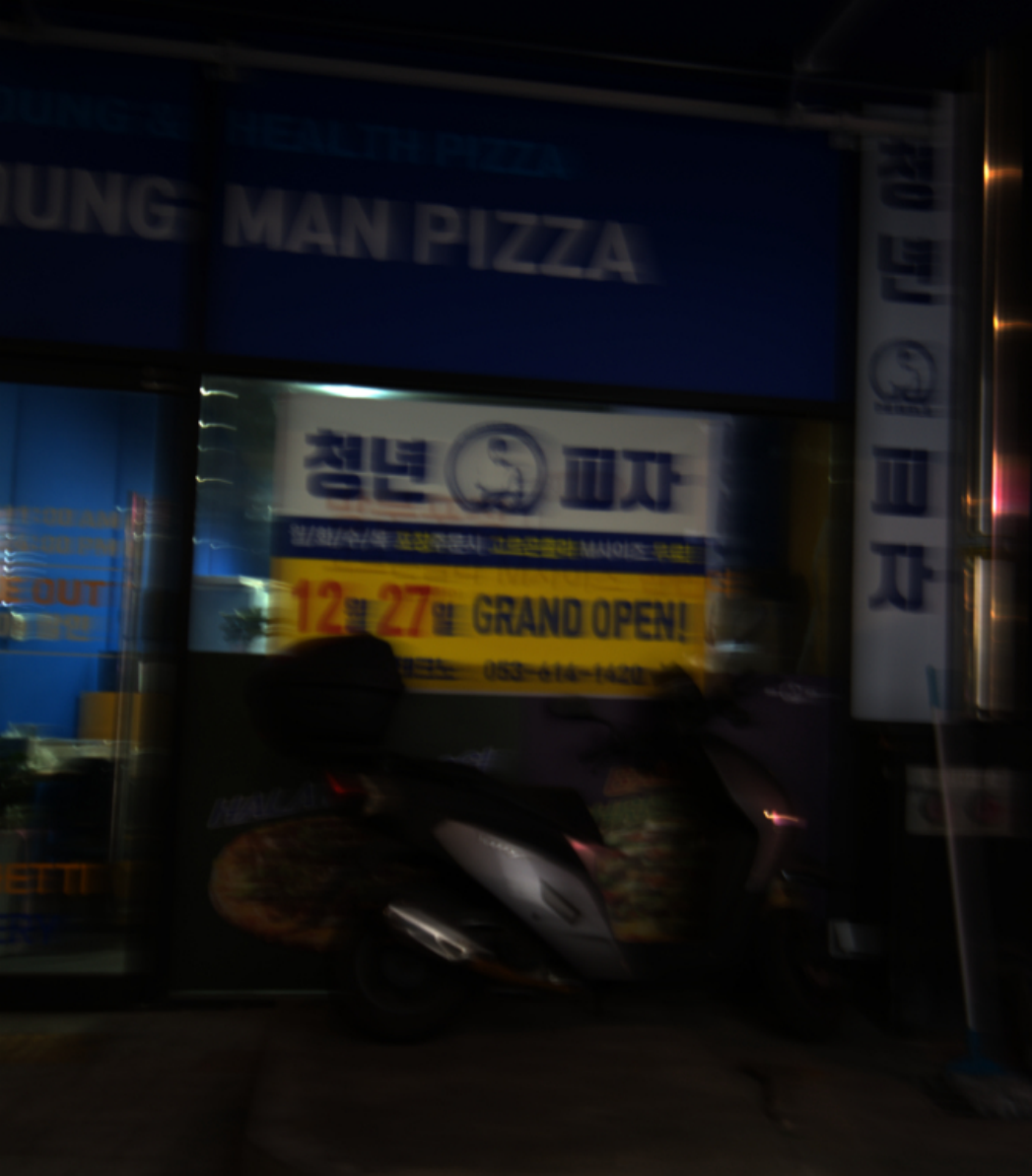}                &
							\includegraphics[width = 0.11\textwidth]{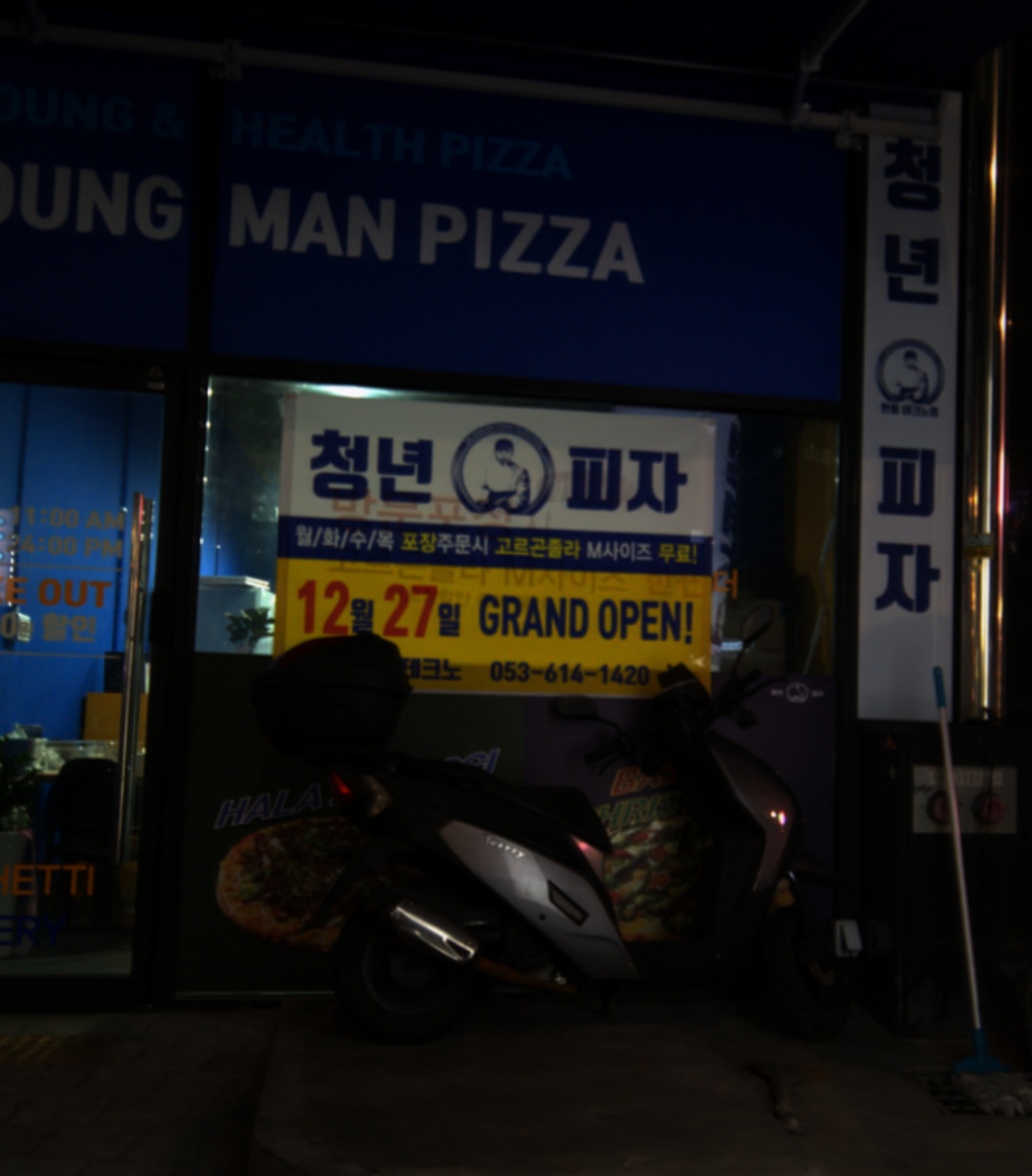}            &
							\includegraphics[width = 0.11\textwidth]{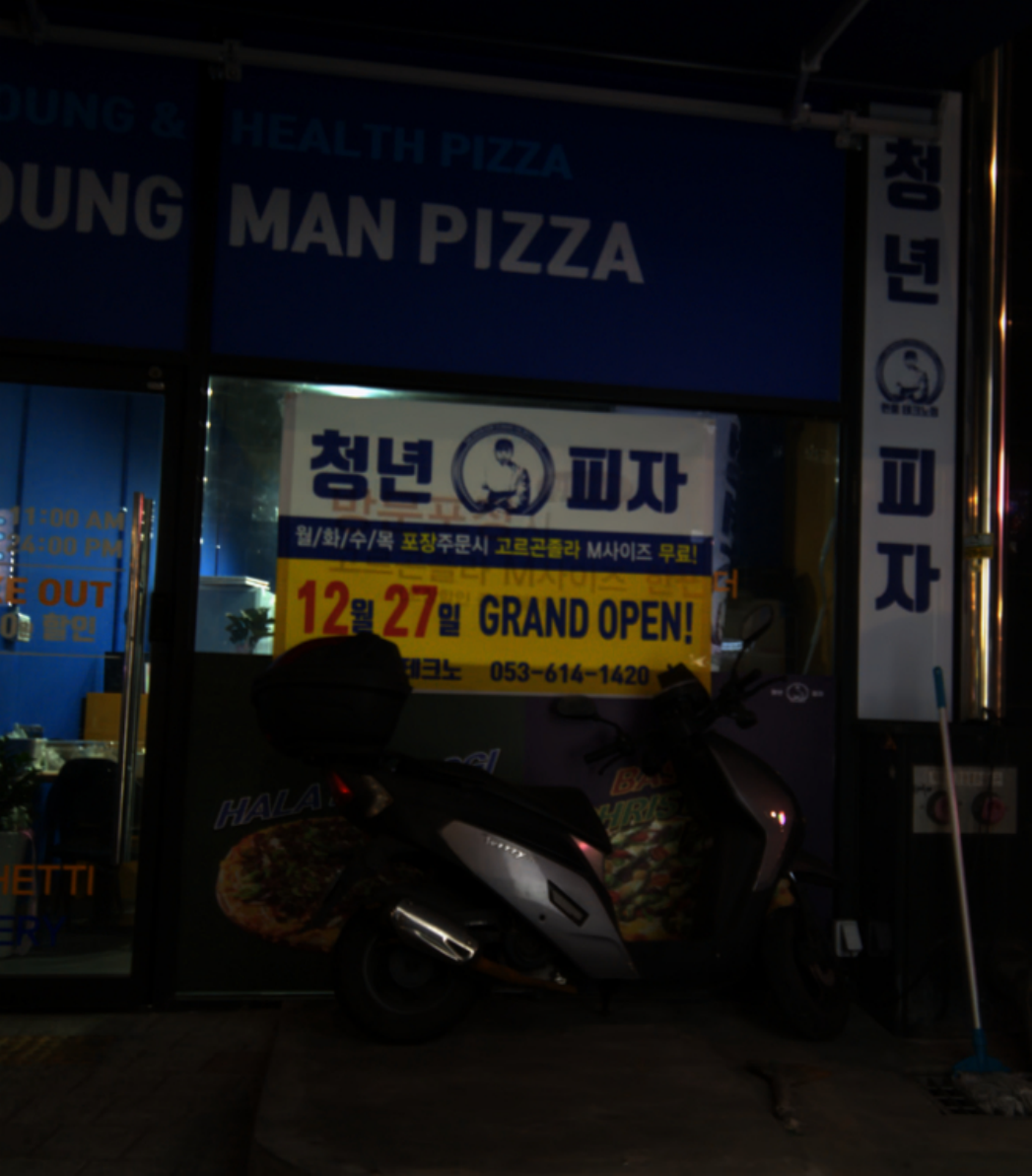}                         &
							\includegraphics[width = 0.11\textwidth]{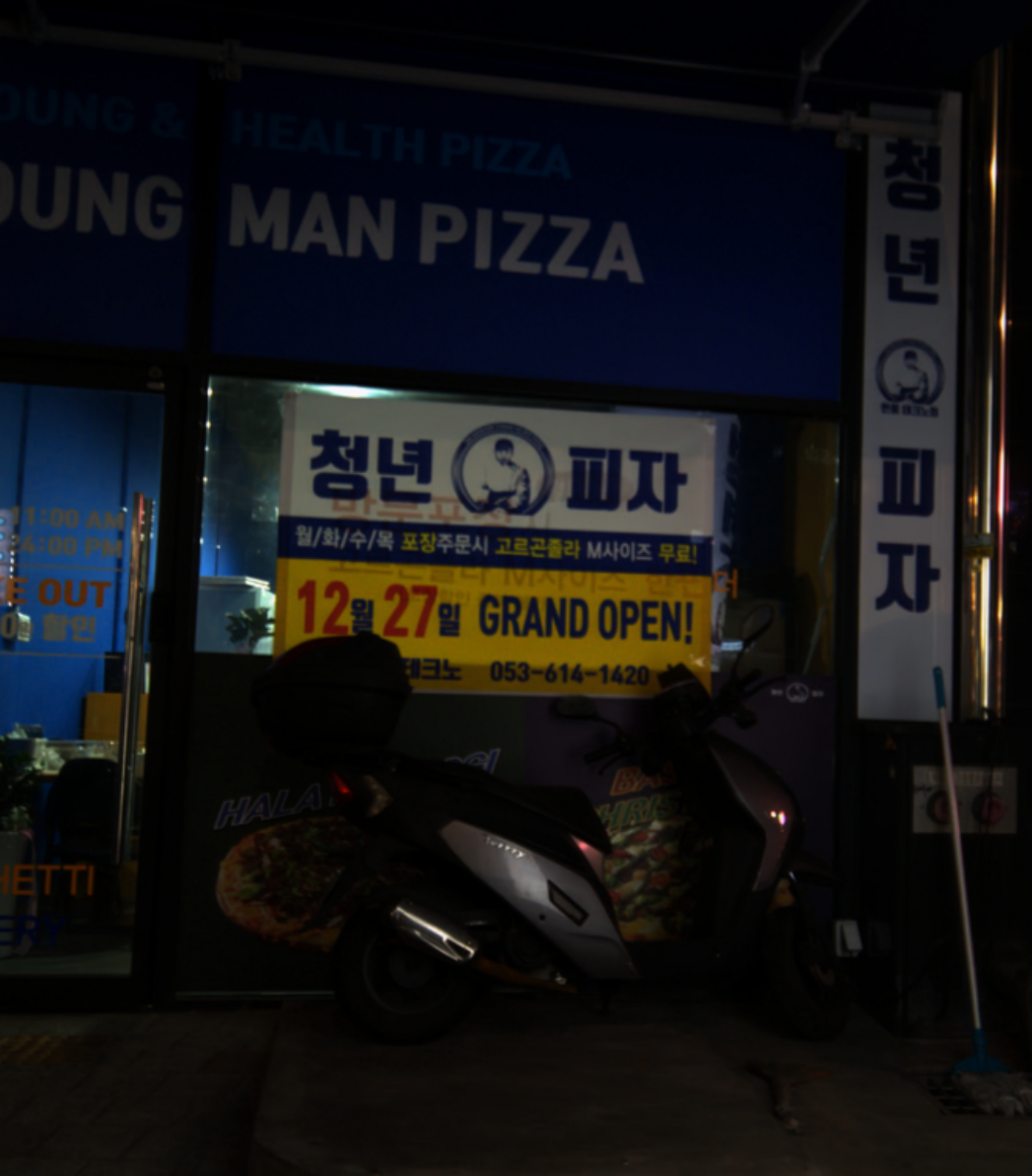}                         &   
							\includegraphics[width = 0.11\textwidth]{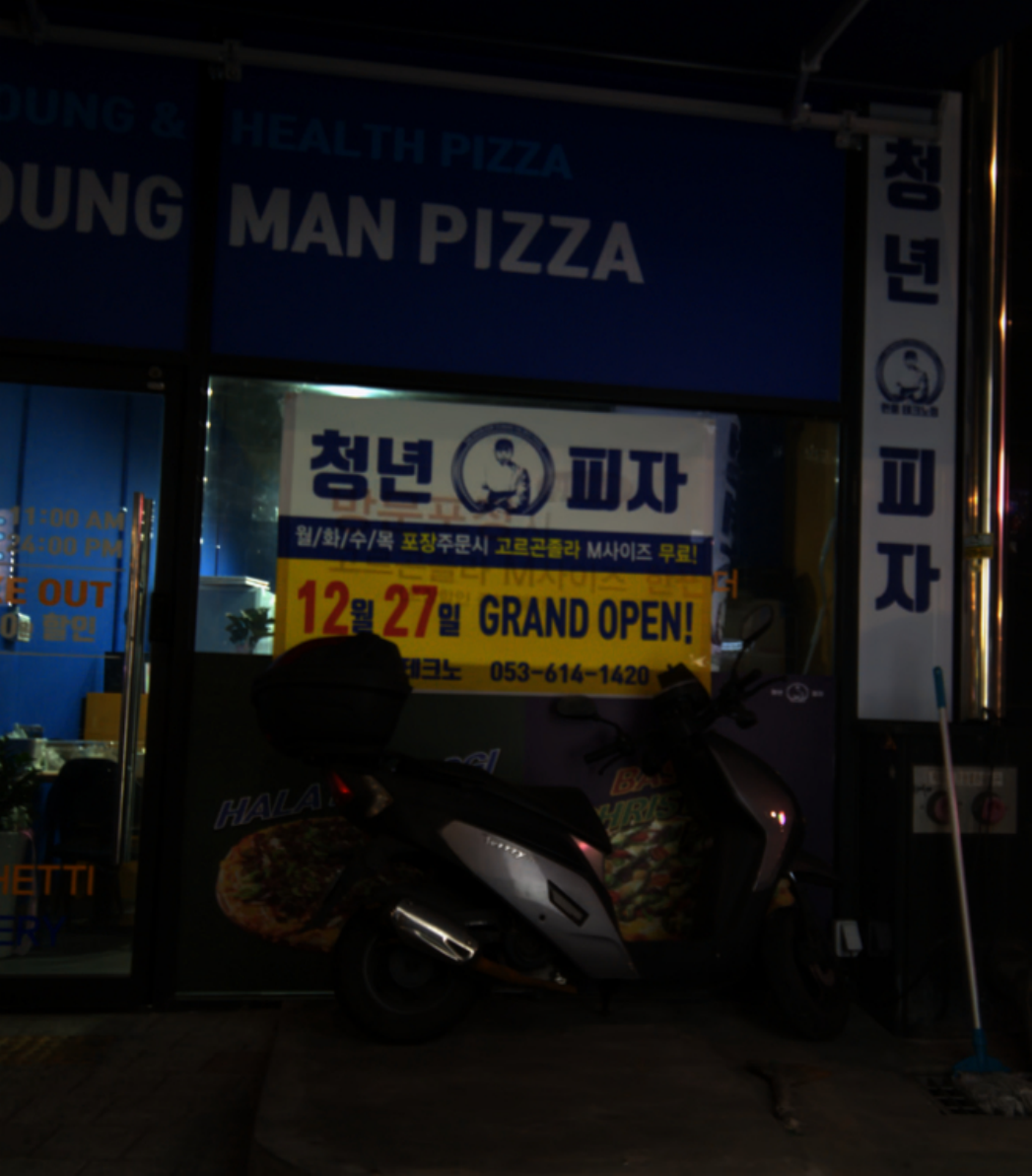}                         & 
							\includegraphics[width = 0.11\textwidth]{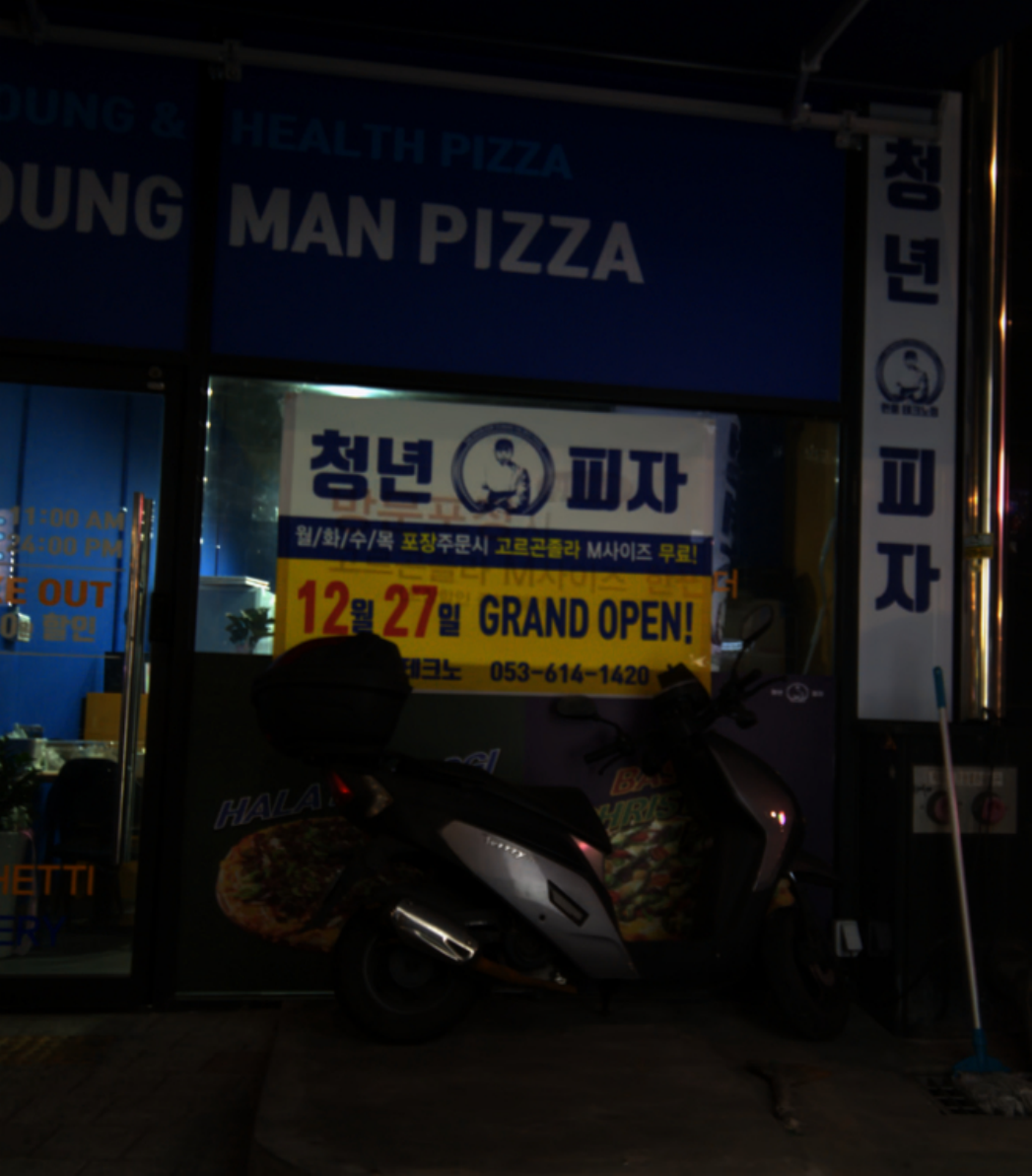}                         & 
							\includegraphics[width = 0.11\textwidth]{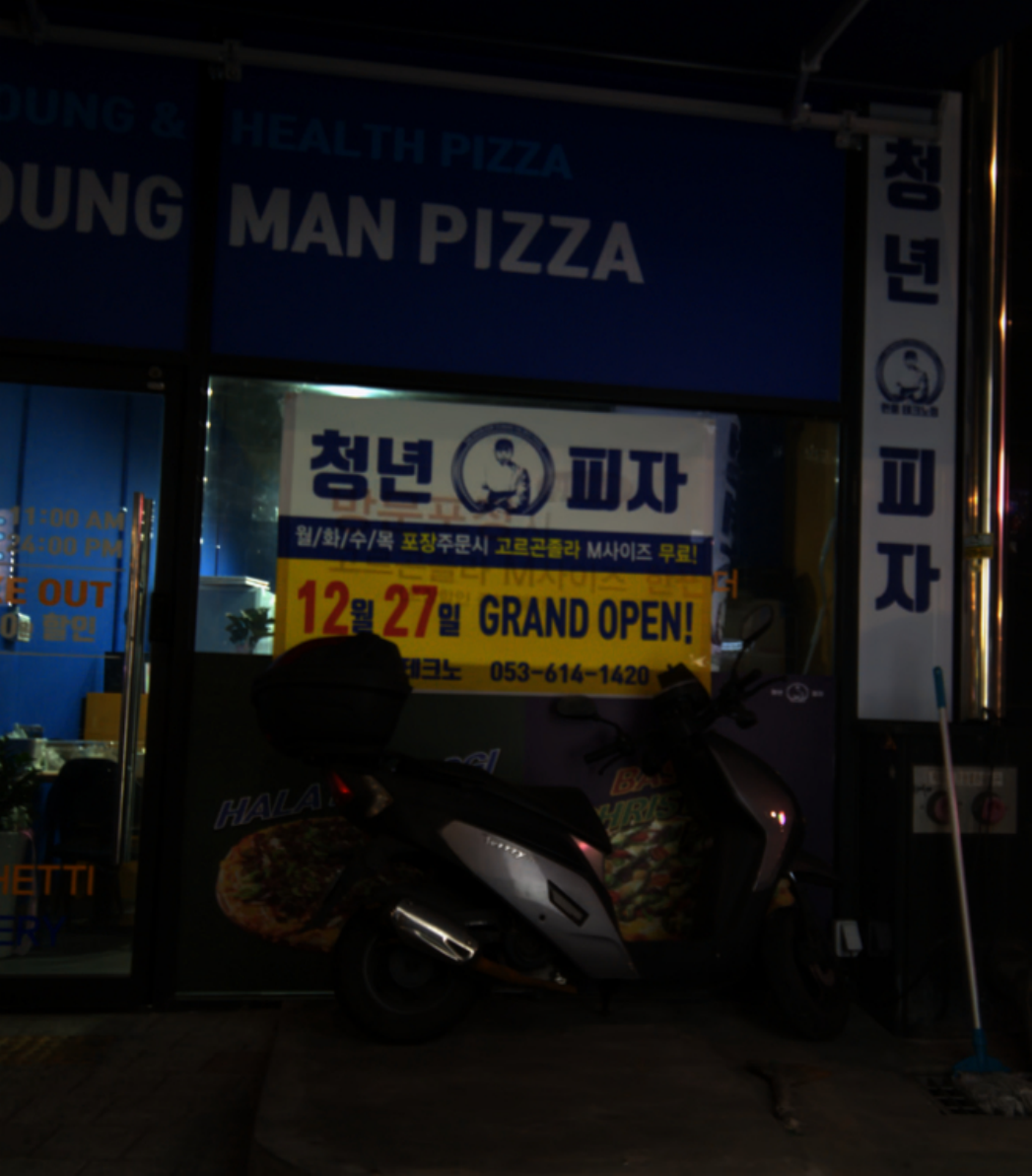}                         & 
							\includegraphics[width = 0.11\textwidth]{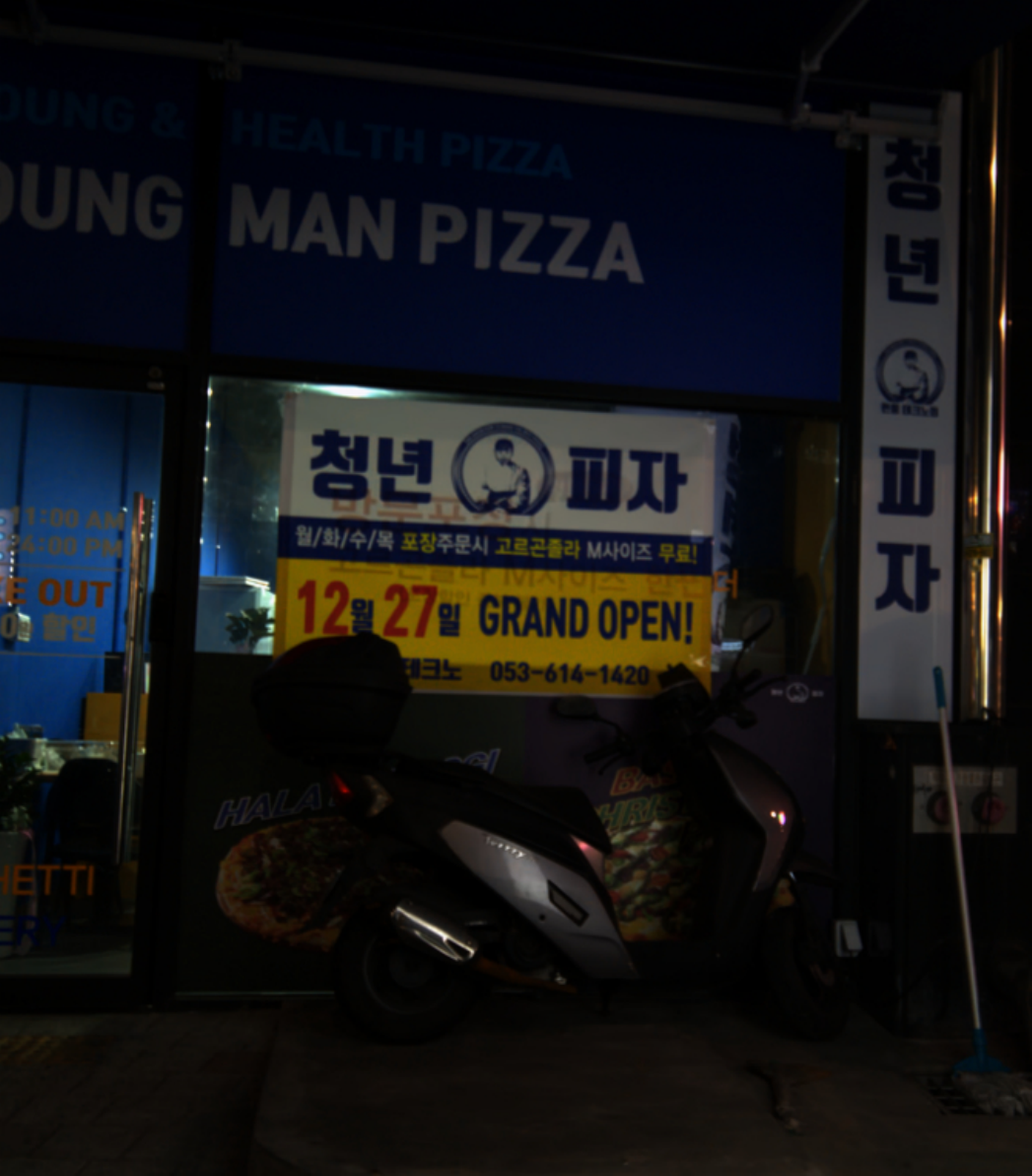}  &
							\includegraphics[width = 0.11\textwidth]{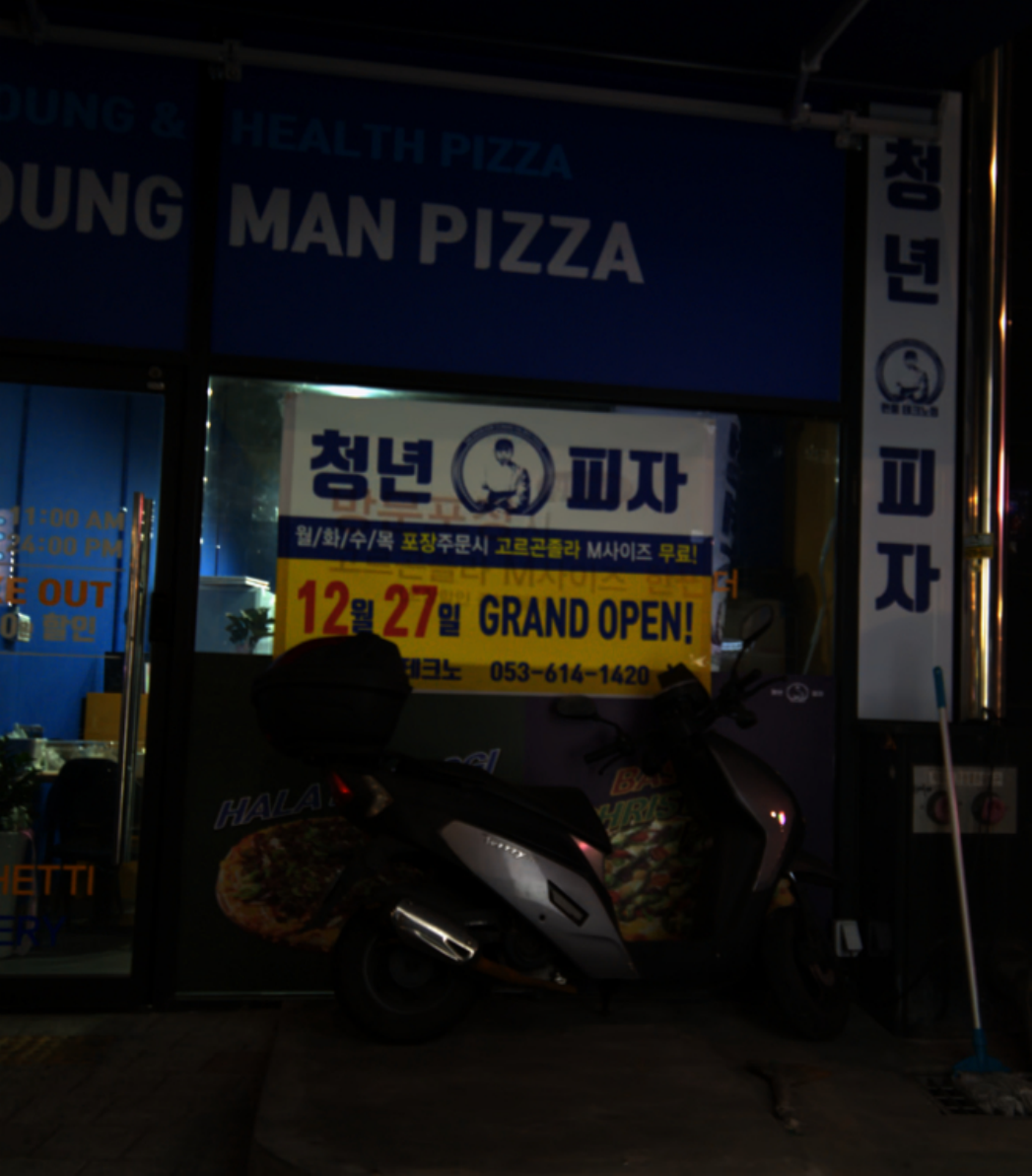}      
							\\
							
							\includegraphics[width = 0.11\textwidth]{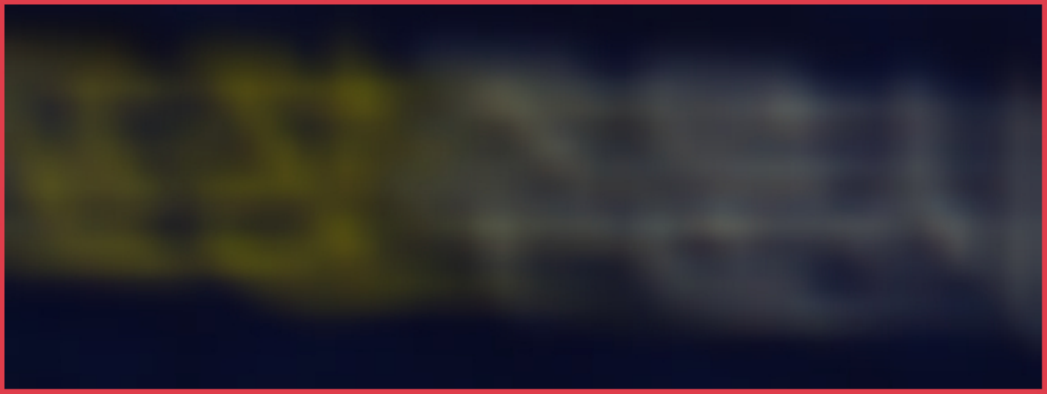}                &
							\includegraphics[width = 0.11\textwidth]{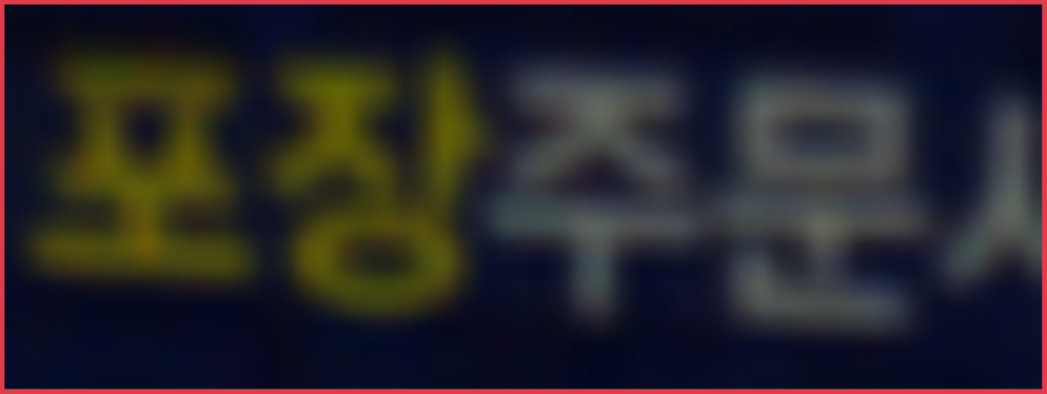}            &
							\includegraphics[width = 0.11\textwidth]{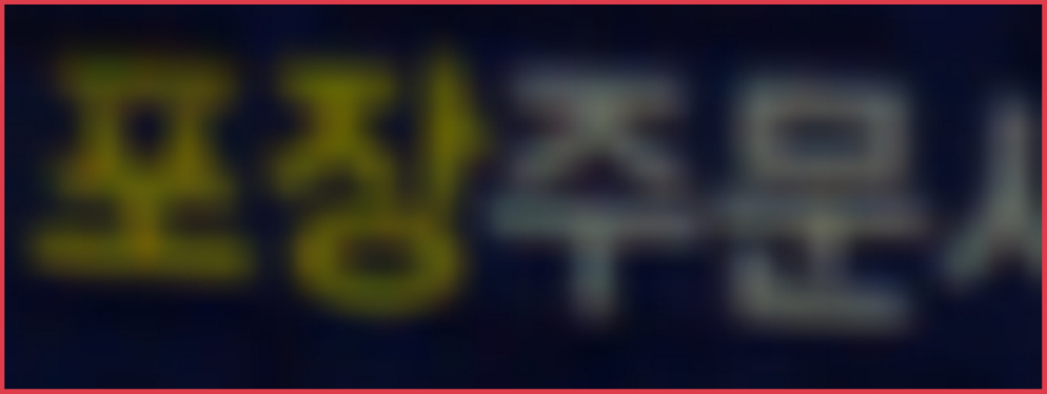}                         &
							\includegraphics[width = 0.11\textwidth]{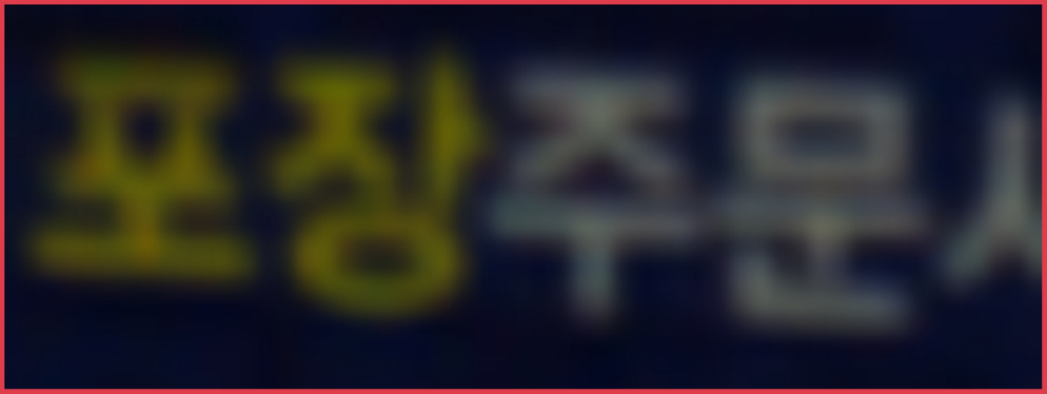}                         &   
							\includegraphics[width = 0.11\textwidth]{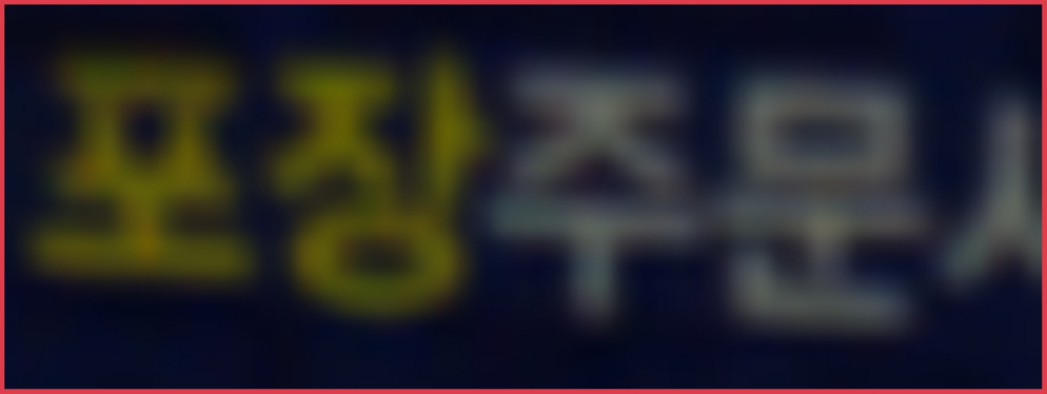}                         & 
							\includegraphics[width = 0.11\textwidth]{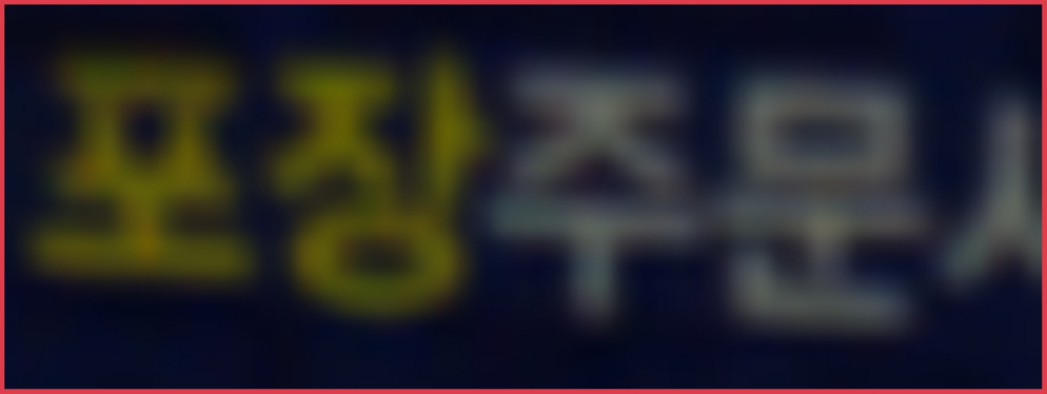}                         & 
							\includegraphics[width = 0.11\textwidth]{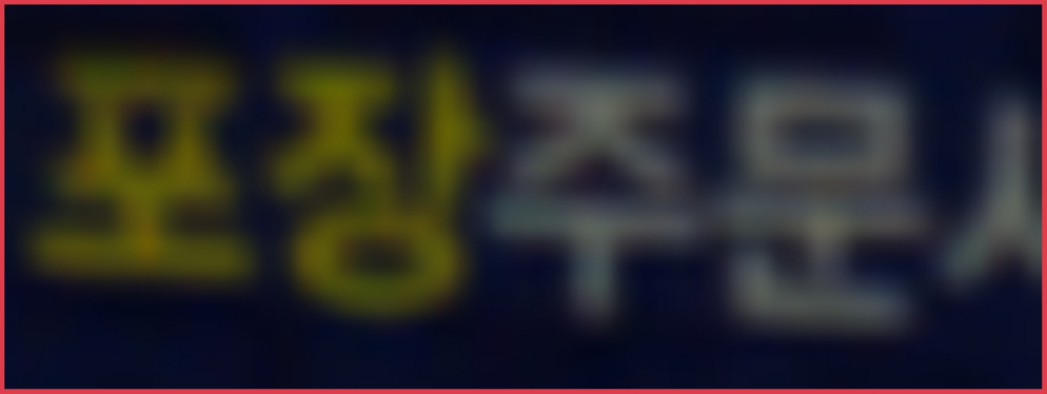}                         & 
							\includegraphics[width = 0.11\textwidth]{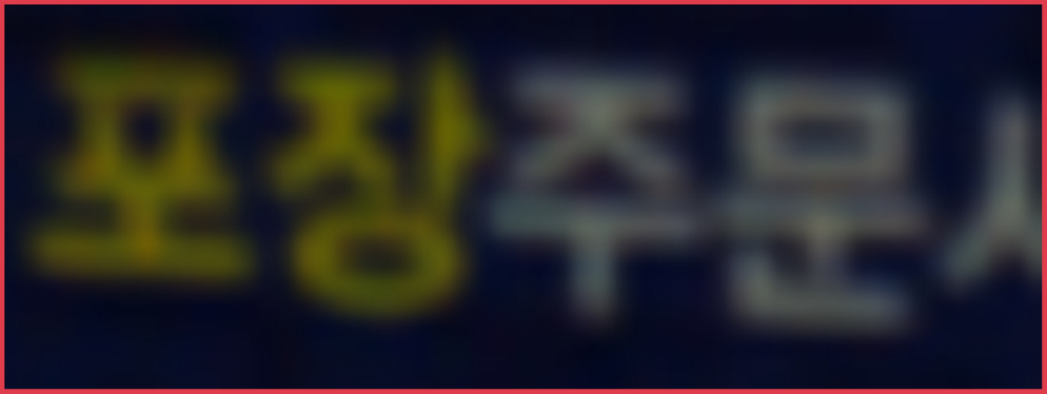}  &
							\includegraphics[width = 0.11\textwidth]{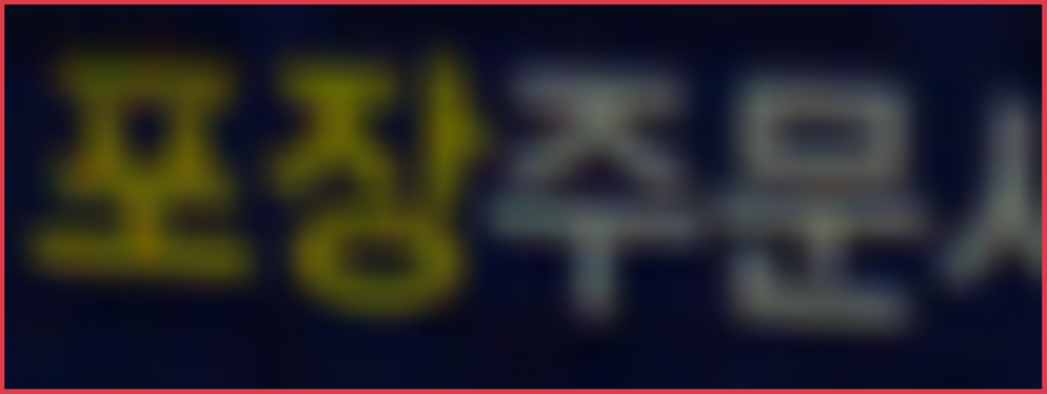}          
							\\
							
							(a) PSNR/SSIM (R-R)   &
							(b) 35.96/0.950     &
							(c) 36.20/0.955     &  
							(d) 35.87/0.950    & 
							(e) \textbf{39.03}/0.962   & 
							(f) 35.44/0.930   & 
							(g) 33.20/0.915   & 
							(h) 35.99/\textbf{0.967}    &
							(i) ($+\infty$/1)     \\
							
							Input &
							MPRNet & 
							Restormer&  
							NAFNet&    
							MAXIM& 
							MC-Blur& 
							ASTL& 
							Ours    &
							GT   \\
							
						\end{tabular}
					\end{center}
					\vspace{-4mm}
					\caption{Qualitative evaluations on G-P (GoPro), R-J (RealBlur-R), R-R (RealBlur-J) and 4K (4KRD) datasets. Our proposed multi-scale cubic-mixer generates much clearer details and sharp edges. (Zoom in for best view)}
					%\vspace{-2mm}
					\label{fig-ex1}
				\end{figure*}
				%
				%\begin{figure*}[h]
				%	\centering
				%	\includegraphics[width=0.8\textwidth]{plot/main_plot/Speed_network.pdf}
				%	\vspace{-2mm}
				%	\caption{Trade-off of speed and accuracy between our proposed deblurring method and state-of-the-art methods on the 4K Resolution Deblurring (4KRD) dataset. The red line denotes the real-time approach for UHD image deblurring at 30 fps. The \textbf{yellow region} represents the methods that cannot process UHD images directly and needs to use the downsampling-deblurring-upsampling strategy. 
					%For instance, the maximum resolution can be handled by \citep  {KupynMWW19, SuinPR20} is around 2K. 
					%	The proposed algorithm generate deblurred images efficiently and accurately at UHD resolution (4K).}
				%\vspace{-4mm}
				%	\label{speed}
				%\end{figure*}

	\subsection{Dataset}

Although Deng et al.~\citep  {deng2021multi} build a blurry dataset with 4K resolution, there are less than 100 scenes, for which we expand this dataset.
%Since there is no public UHD datasets to evaluate deblurring methods, we use the multi-frame fusion method~\citep  {NahBHMSTL19} to generate a 4K Resolution Deblurring (4KRD) dataset. deng2021multi
%The proposed dataset covers a variety natural scenarios with widely varying data distributions.
We expand the new 4KRD dataset with two main schemes: frame interpolation and dataset synthesis. 
The video capturing devices are three mainstream mobile phones (e.g., iPhone 11 Pro Max, Samsung S20 Ultra, and HUAWEI Mate 30 Pro).
%We also use a DJI Osmo Mobile 3 to stabilize mobile phone so that captured videos as clear as possible.
High frame rates are necessary for the subsequent multi-frame fusion to ensure the fluidity and stability of frames in the synthetic dataset. 
Due to hardware configuration limitations, we cannot directly capture 4K videos at high-frame-rates with smartphones. 
Therefore, we use the frame interpolation method~\citep  {NiklausML17} to interpolate the recorded 4K video from 30/60 fps to 480 fps as the scheme in~\citep  {NahBHMSTL19}.
Then we generate blurry frames by averaging a series of successive sharp frames.
We synthesized a total of 40 different scene data sets, each scene has 100 images.
In addition to our 4K resolution dataset, we also use the public GoPro~\citep  {NahKL17}, RealBlur-J~\citep  {rim_2020_ECCV}, and RealBlur-R~\citep  {rim_2020_ECCV} datasets to evaluate the effectiveness of our model. 

\subsection{Performance Comparisons}
\label{PC}
We compare our algorithm against four state-of-the-art deblurring methods
of MPRNet~\citep  {zamir2021multi}, Restormer~\citep  {zamir2021restormer}, NaFNet~\citep  {chen2022simple}, MAXIM~\citep  {tu2022maxim}.
%\citep  {Chakrabarti16,KupynMWW19,NahKL17,SuinPR20,TaoGSWJ18,ZhangDLK19}. 

{\flushleft\textbf{Implementation details.}} All the experiments are implemented in PyTorch 1.10 and evaluated on a single NVIDIA Titan RTX 3090 GPU with 24GB RAM. 
The batch size is set to 16 during training. 
The Adam optimizer is used to train our models with patch size of $1152 \times 648$. 
The initial learning rate is set to $10^{-4}$. 
Due to the recent deburring models~\citep  {chen2022simple,tu2022maxim,zamir2021restormer,zamir2021multi} cannot directly deblur UHD images on a single Tian RTX GPU, we design the splitting-deblurring-stitching (SDS) scheme for the compared deblurring methods on UHD images.
The SDS method applies deblurring approaches at the full-resolution image, then stitches these patches (16$\sim$32 patches) to generate a clear 4K image.
%
%For low-resolution input, we retain the maximum resolution that these models can handle. 
%For low-resolution input, we retain the maximum resolution that these models can handle. 
%So far, we adopt the DDU strategy for these six deep models~\citep  {Chakrabarti16, KupynMWW19, NahKL17,  SuinPR20, TaoGSWJ18, ZhangDLK19}. 
All the compared methods are fine-tuned on our 4KRD training dataset, we uniformly use SGD optimizer with learning rate set to 0.001 and epoch set to 16.
Note that we do not use any data augmentation methods.
%
%In addition,  the large amount of parameters of our model and limited memory loading, we can only directly process images with a resolution of less than 2K (GoPro dataset) without downsample.
%For the restoration of ultra-high-definition images, pre-downsampling of the images is necessary.
%For ultra-high-definition images, down-sampling is required.

%
\begin{figure*}[t]\scriptsize
	\begin{center}
		\tabcolsep 1pt
		\begin{tabular}{@{}cccccccc@{}}
			
			\includegraphics[width=0.122\textwidth]{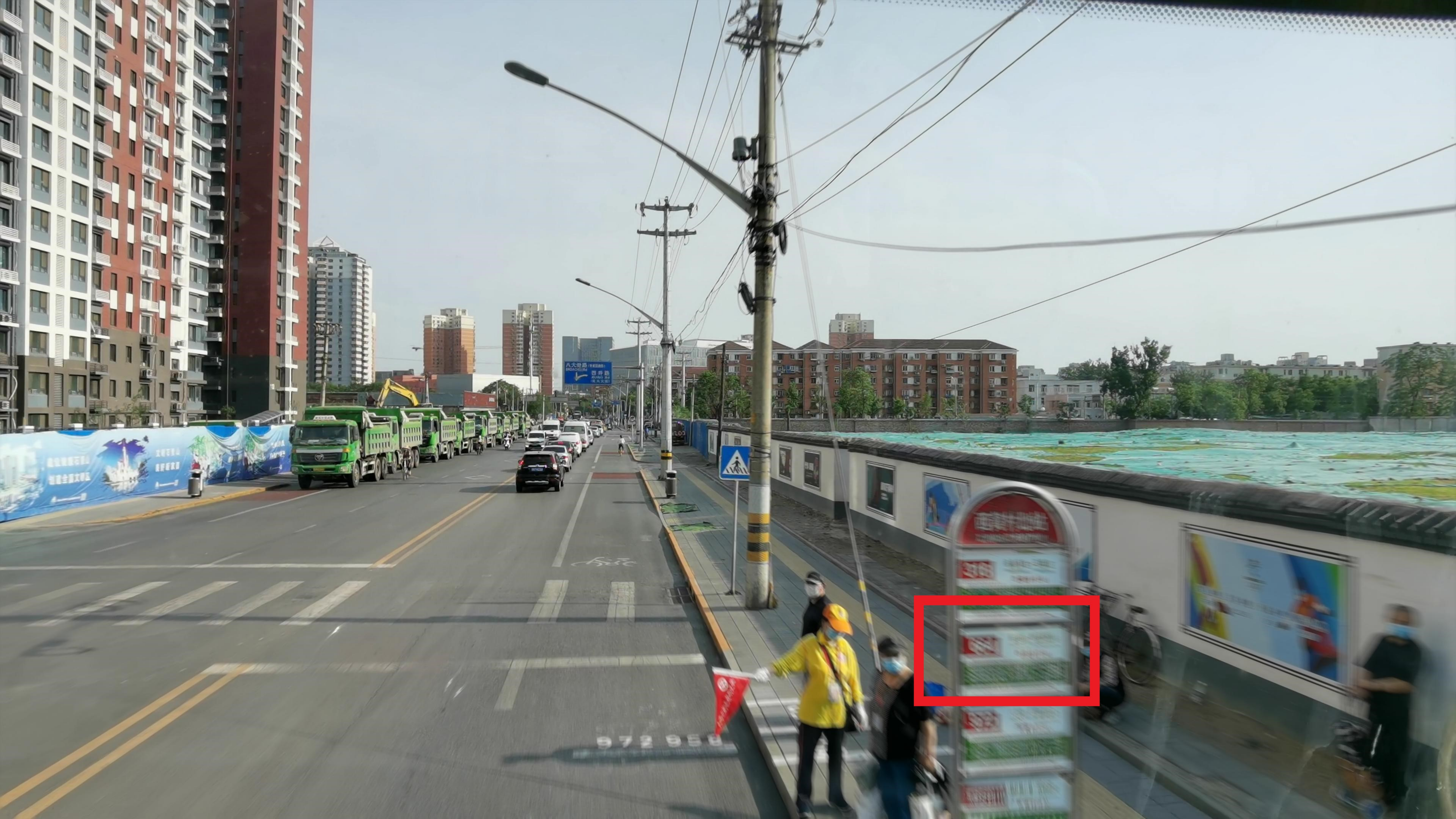}  &
			\includegraphics[width=0.122\textwidth]{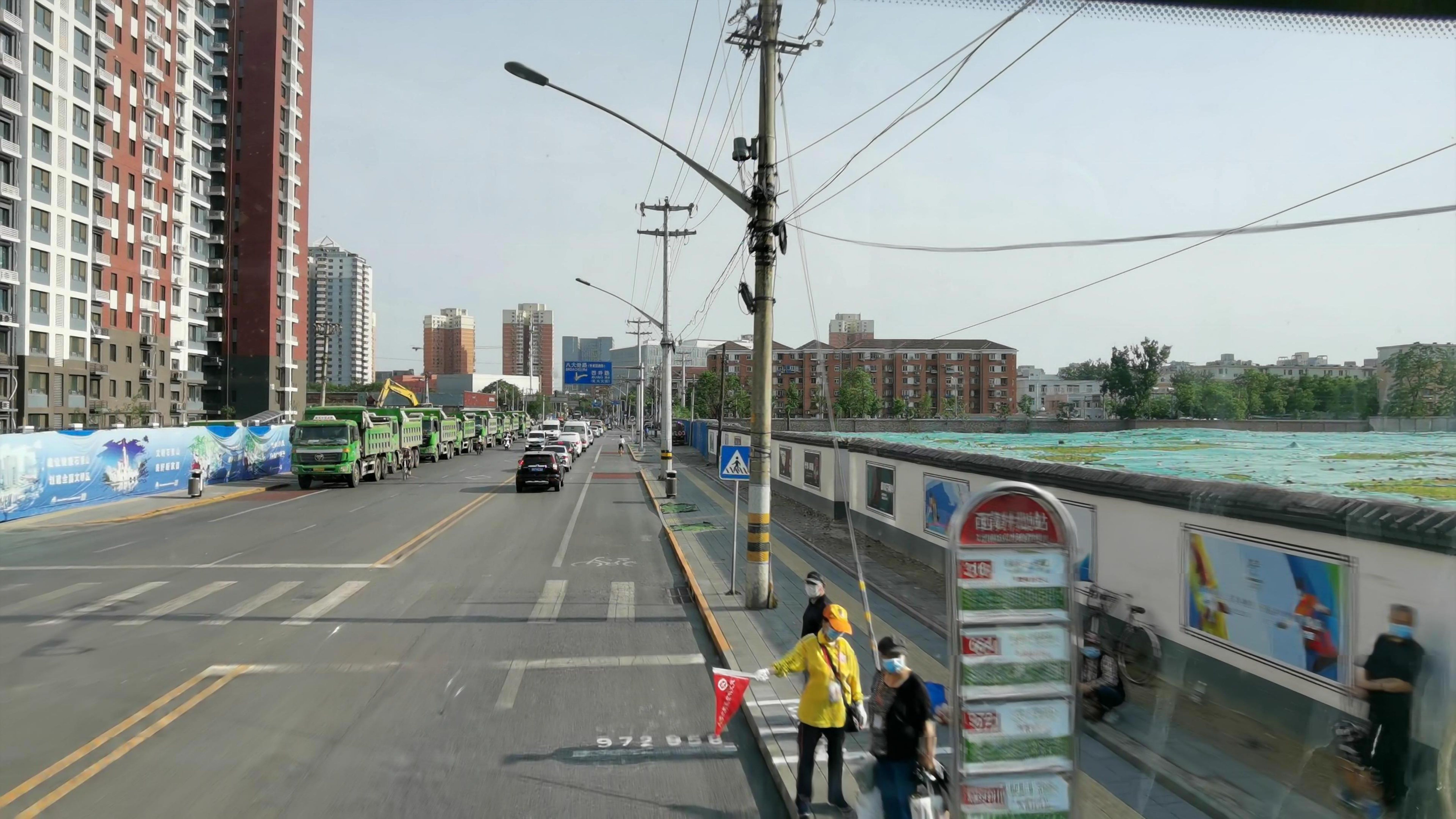}   &
			\includegraphics[width=0.122\textwidth]{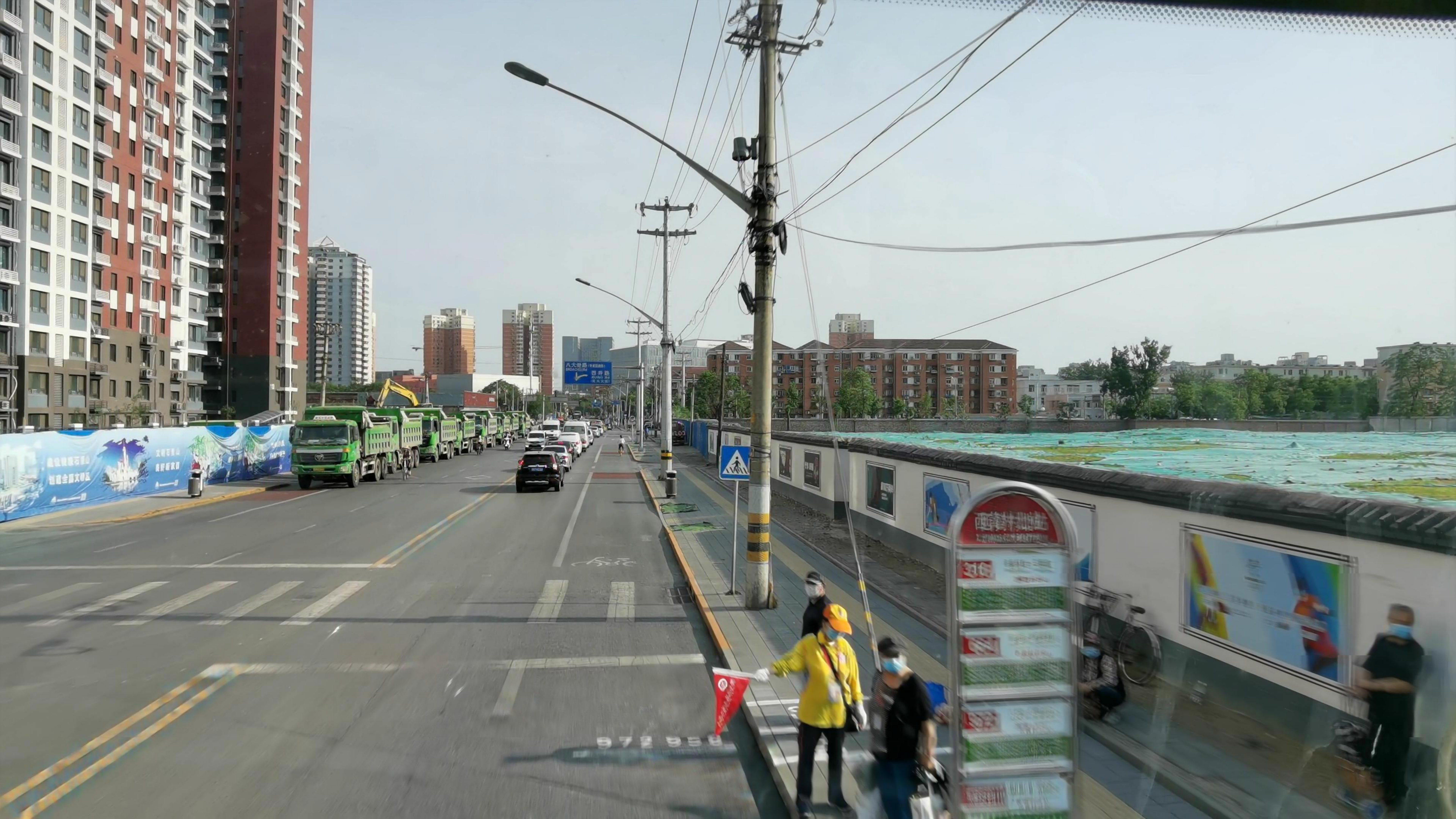}  &
			\includegraphics[width=0.122\textwidth]{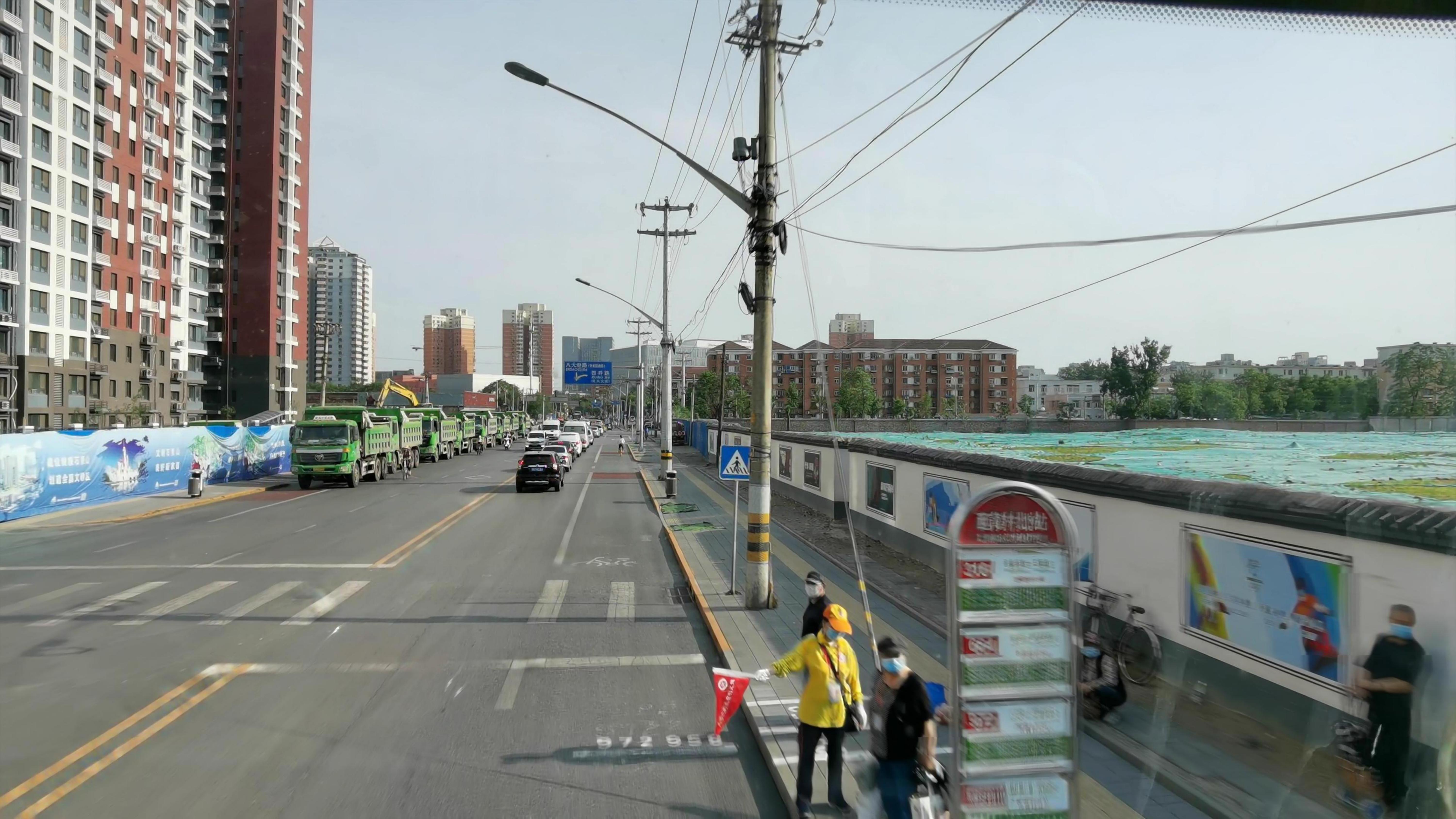}      &   
			\includegraphics[width=0.122\textwidth]{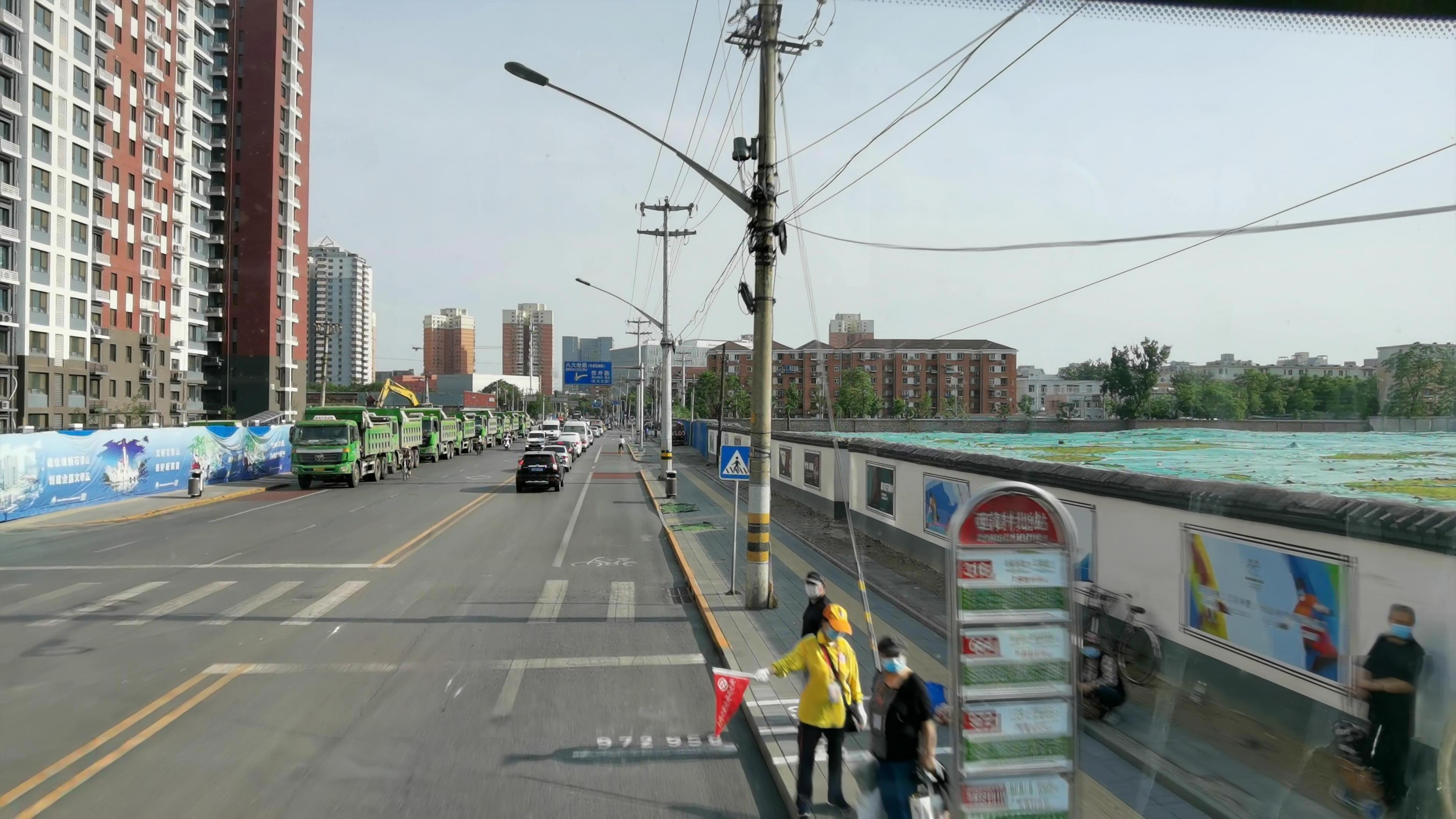}      & 
			\includegraphics[width=0.122\textwidth]{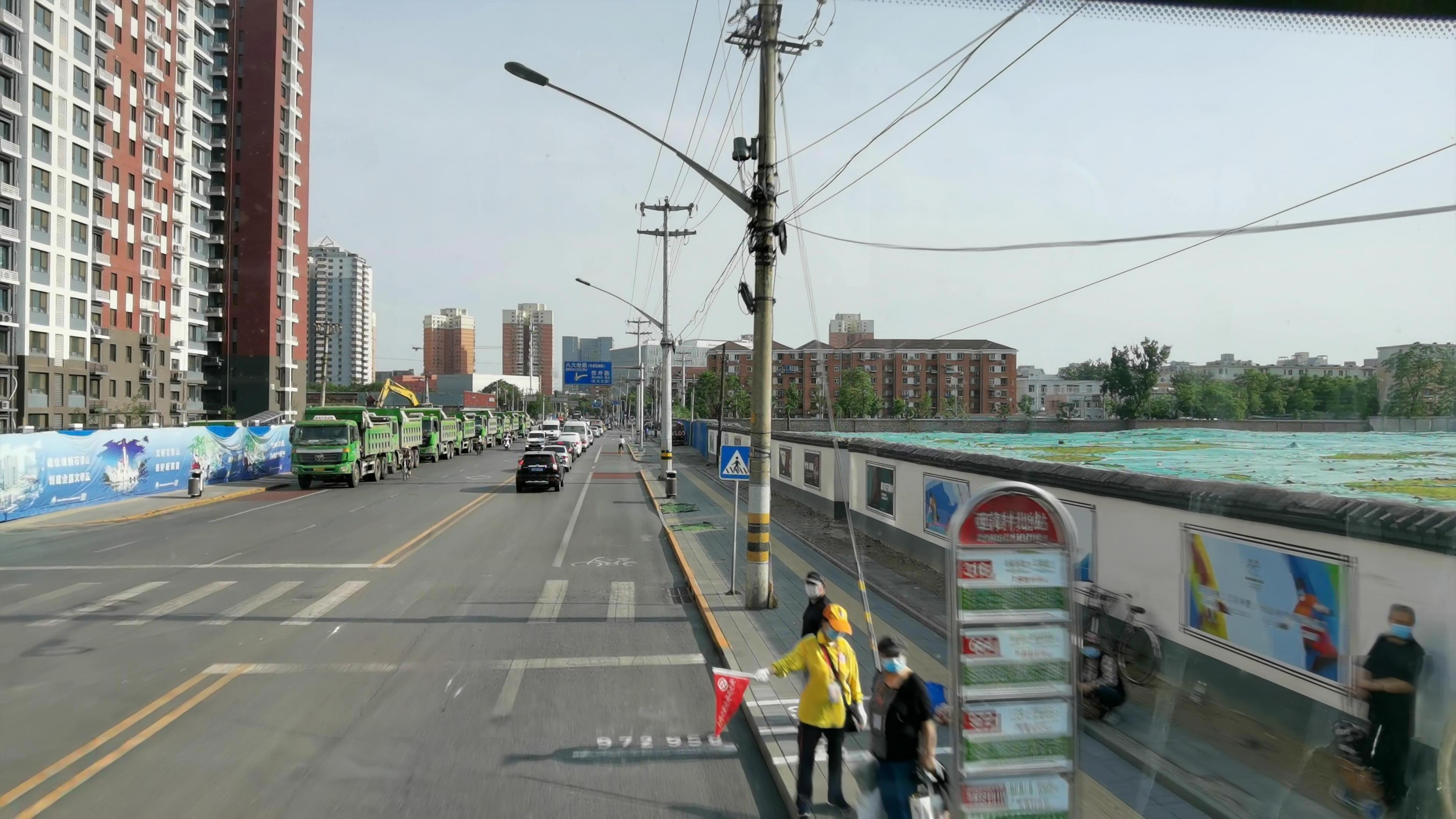}      & 
			\includegraphics[width=0.122\textwidth]{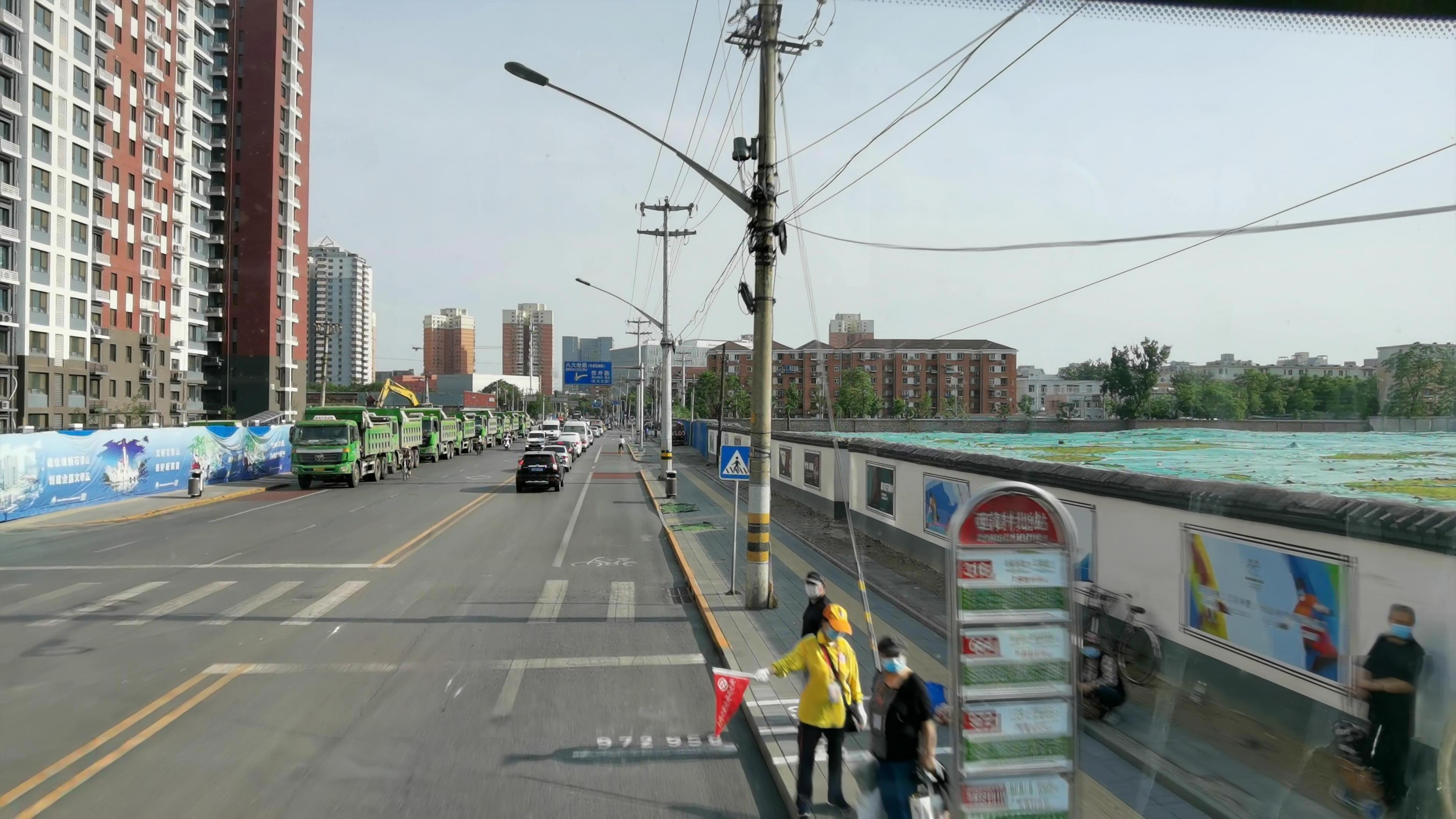}      & 
			\includegraphics[width = 0.122\textwidth]{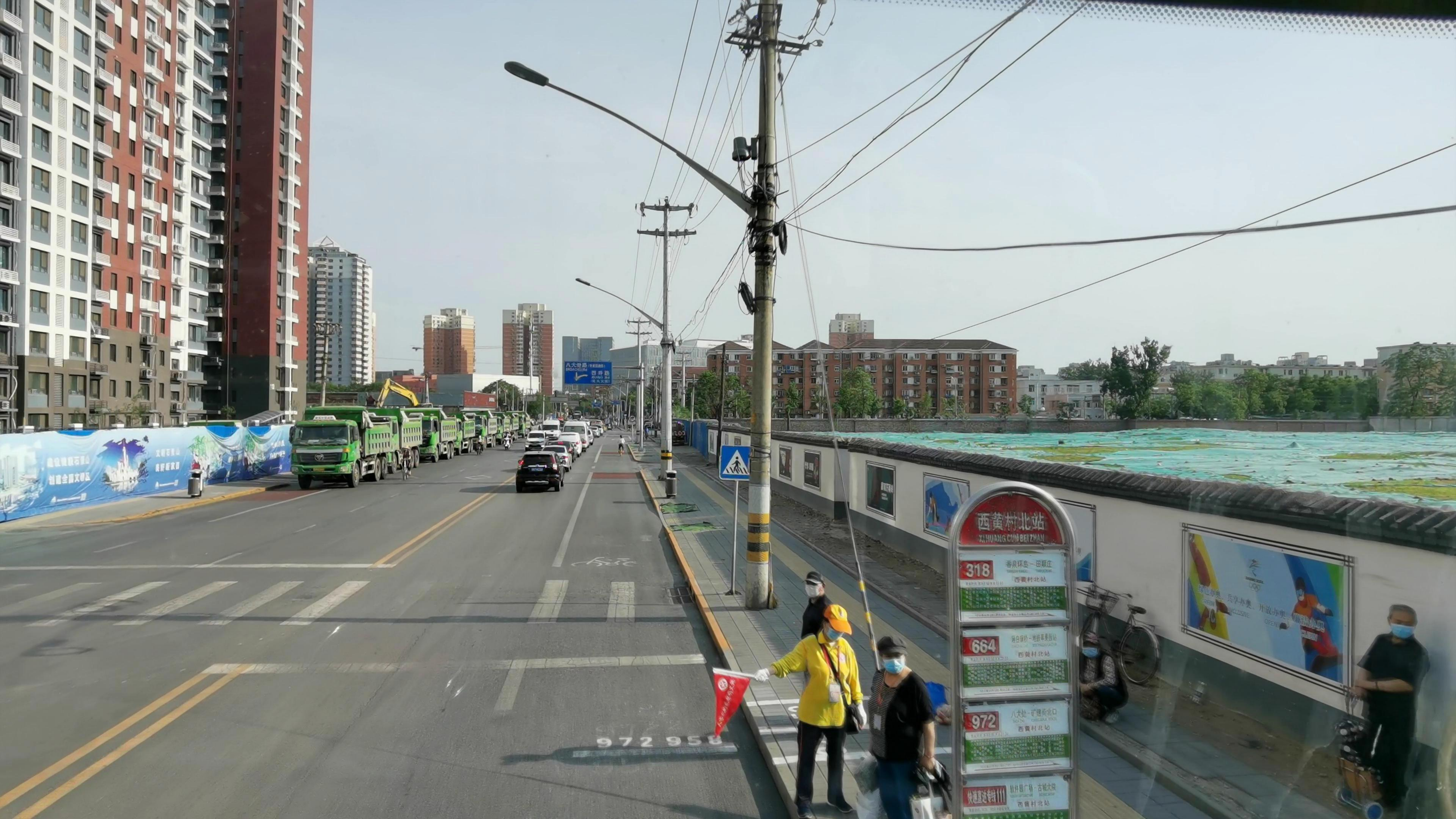}      \\
			
			\includegraphics[width=0.122\textwidth]{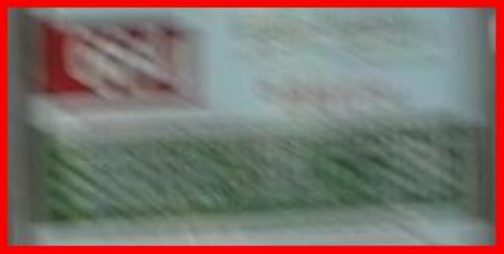}  &
			\includegraphics[width=0.122\textwidth]{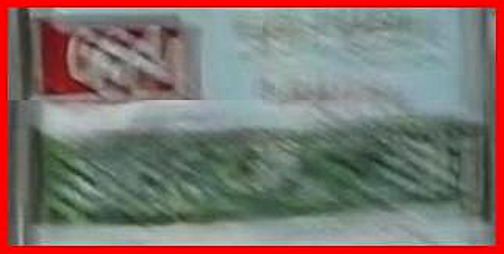}   &
			\includegraphics[width=0.122\textwidth]{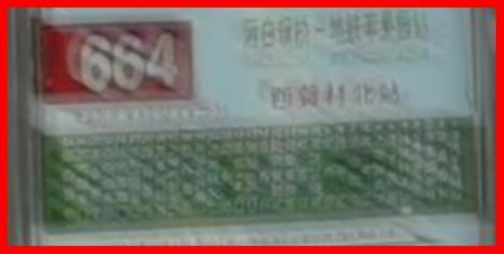}  &
			\includegraphics[width=0.122\textwidth]{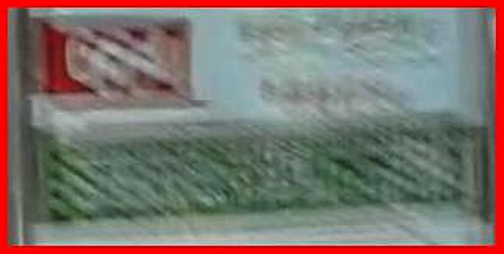}      &   
			\includegraphics[width=0.122\textwidth]{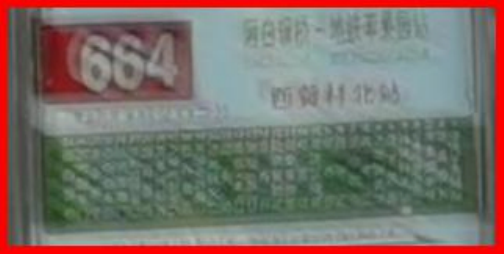}      & 
			\includegraphics[width=0.122\textwidth]{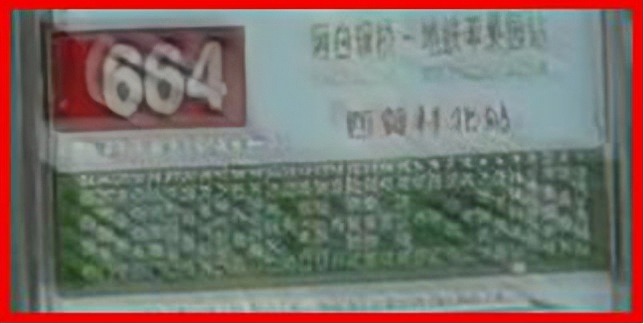}      & 
			\includegraphics[width=0.122\textwidth]{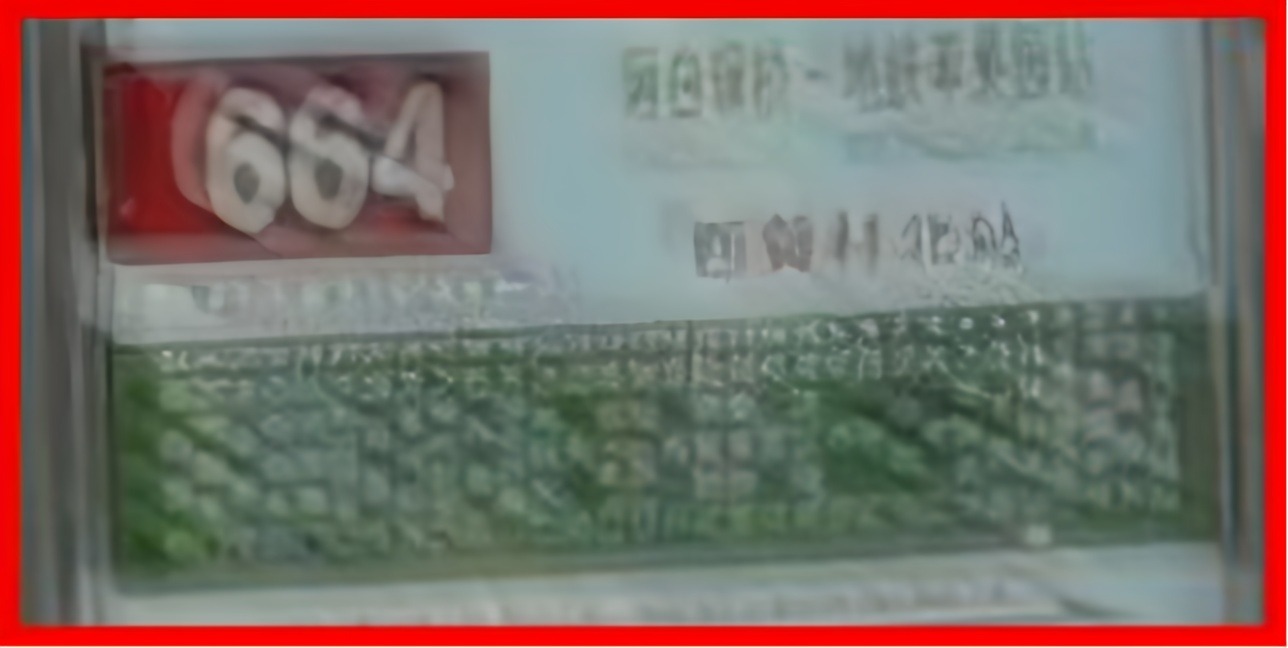}      & 
			\includegraphics[width = 0.122\textwidth]{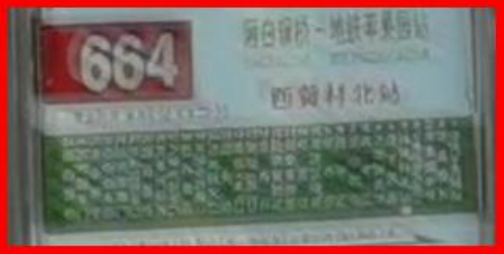}      \\

			\includegraphics[width=0.122\textwidth]{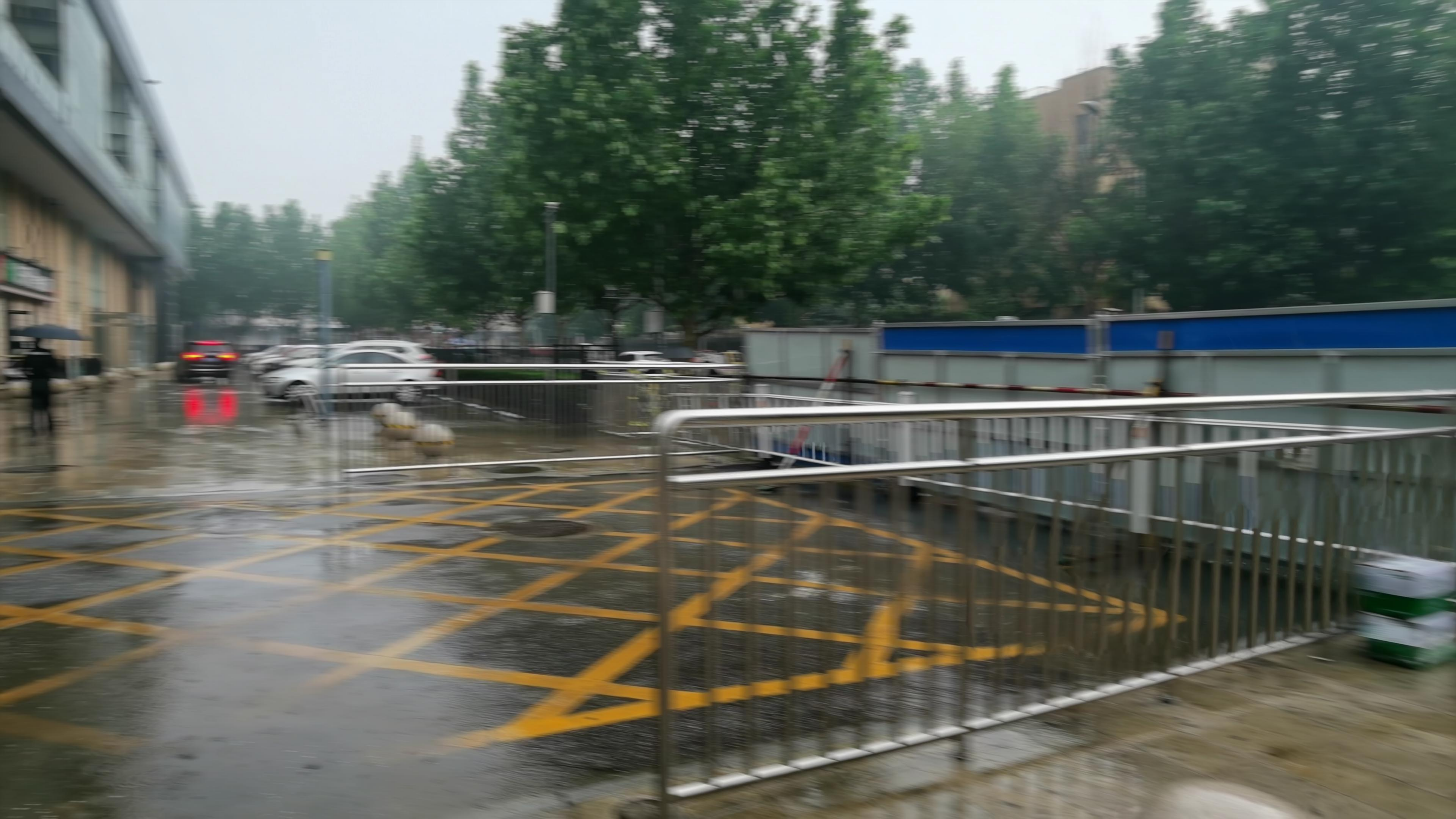}  &
			\includegraphics[width=0.122\textwidth]{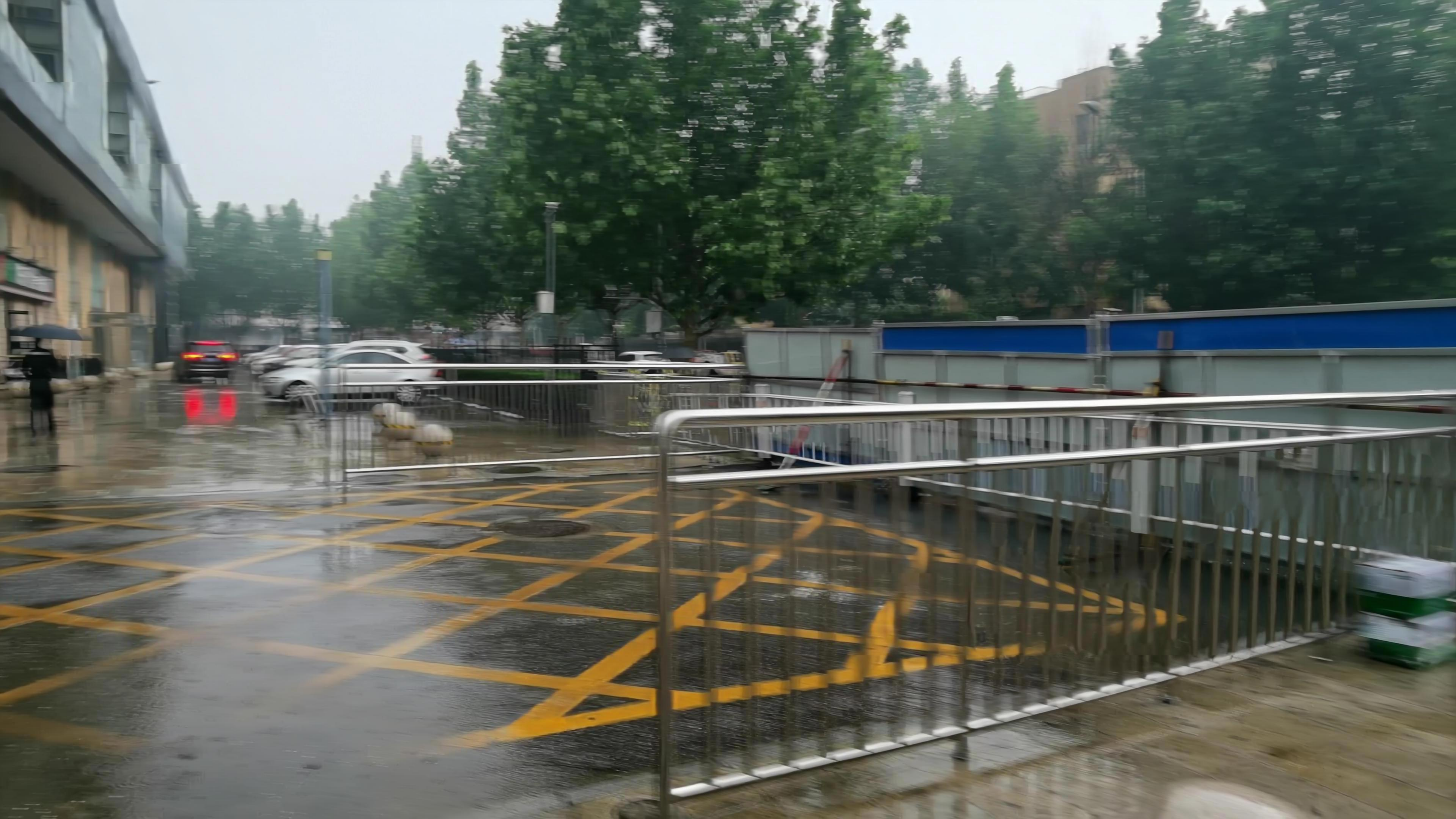}   &
			\includegraphics[width=0.122\textwidth]{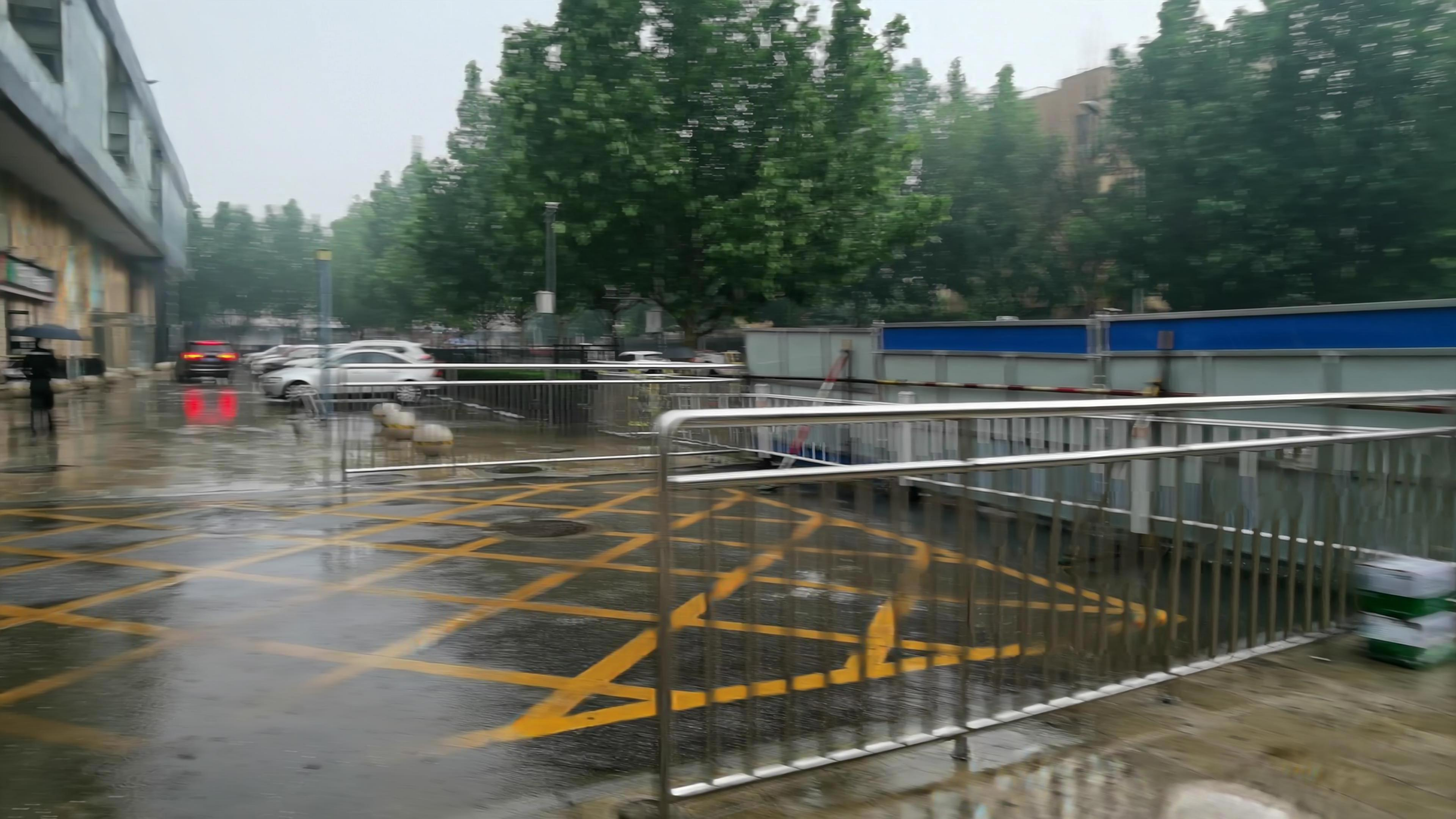}  &
			\includegraphics[width=0.122\textwidth]{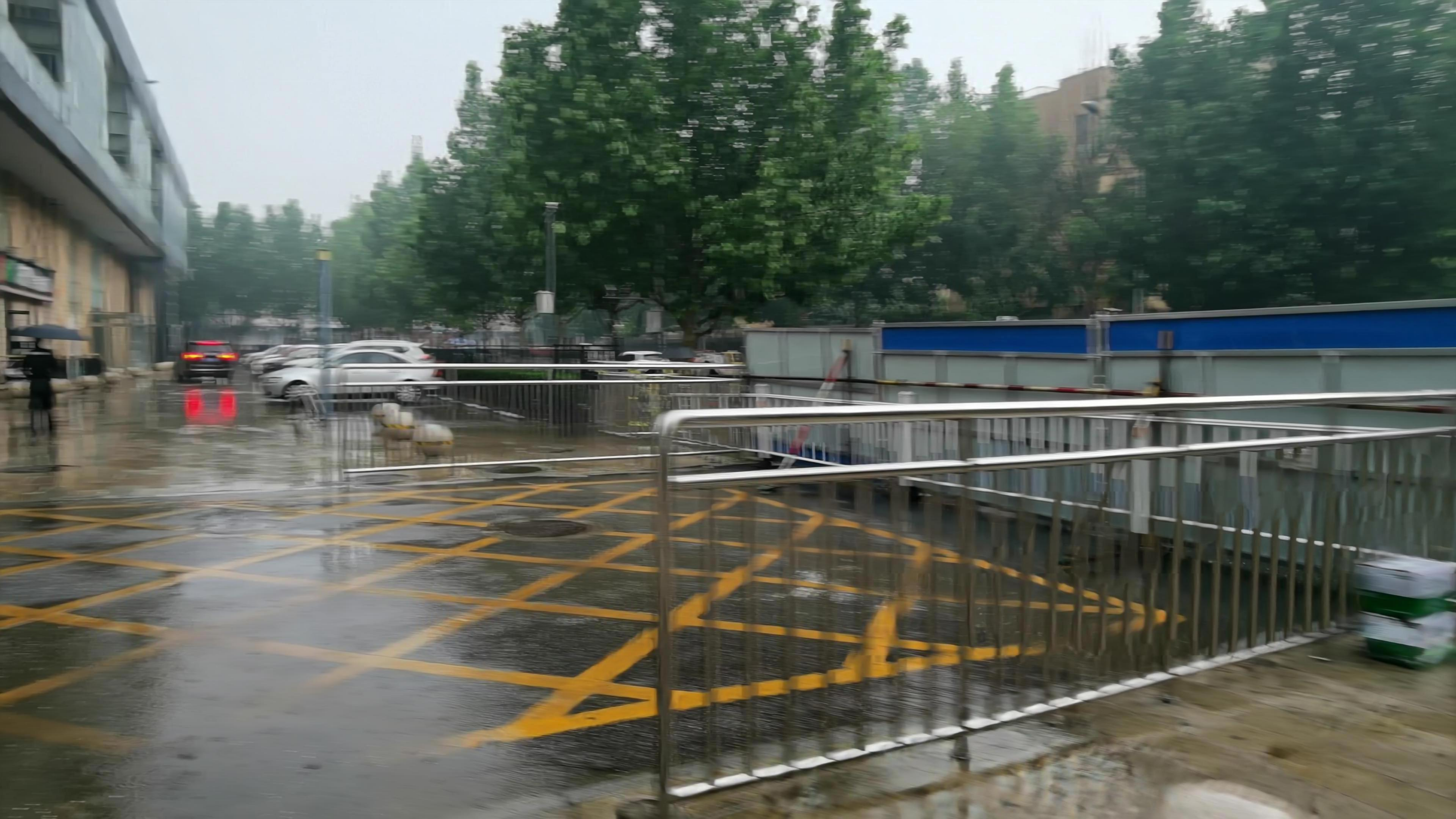}      &   
			\includegraphics[width=0.122\textwidth]{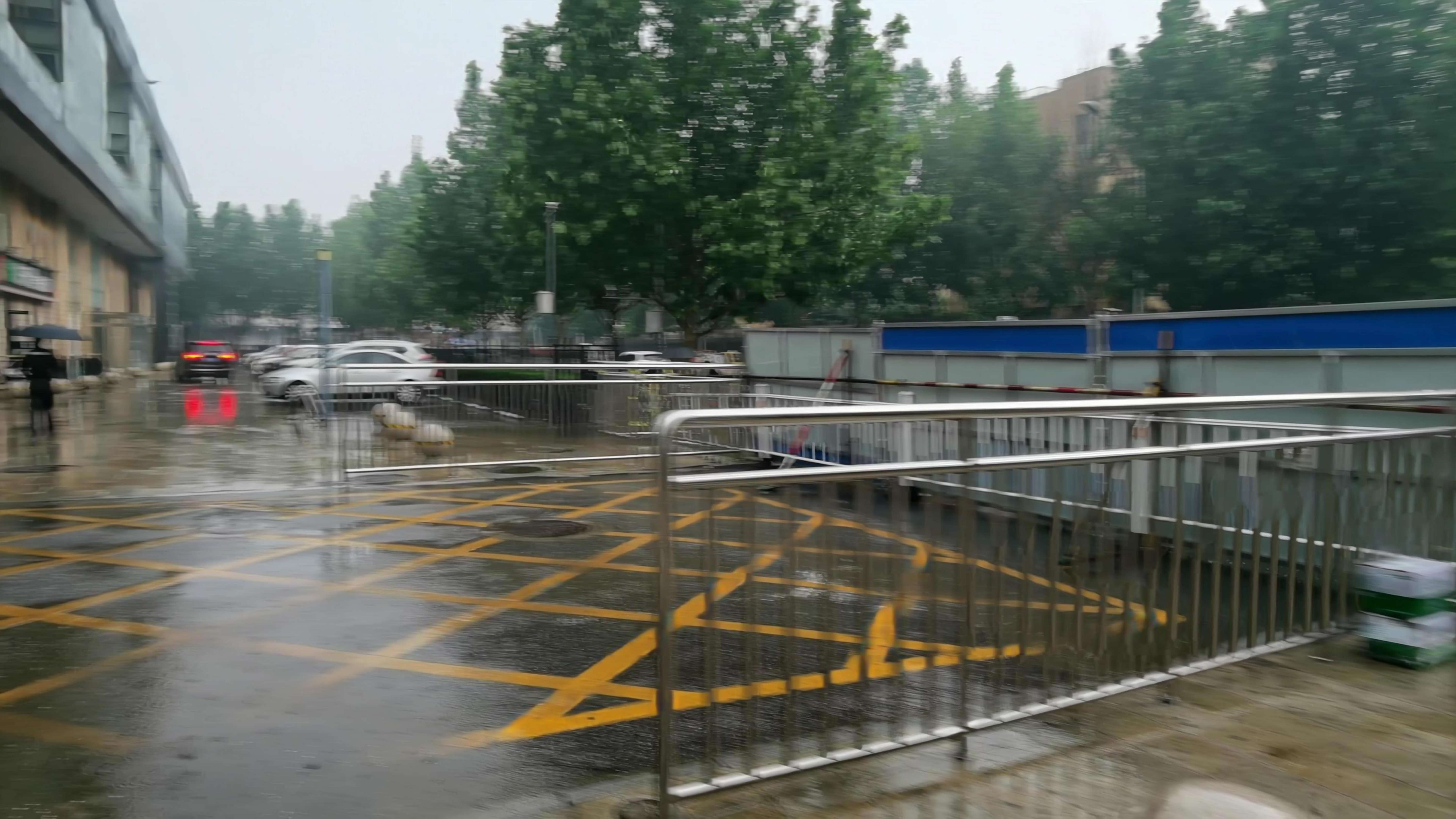}      & 
			\includegraphics[width=0.122\textwidth]{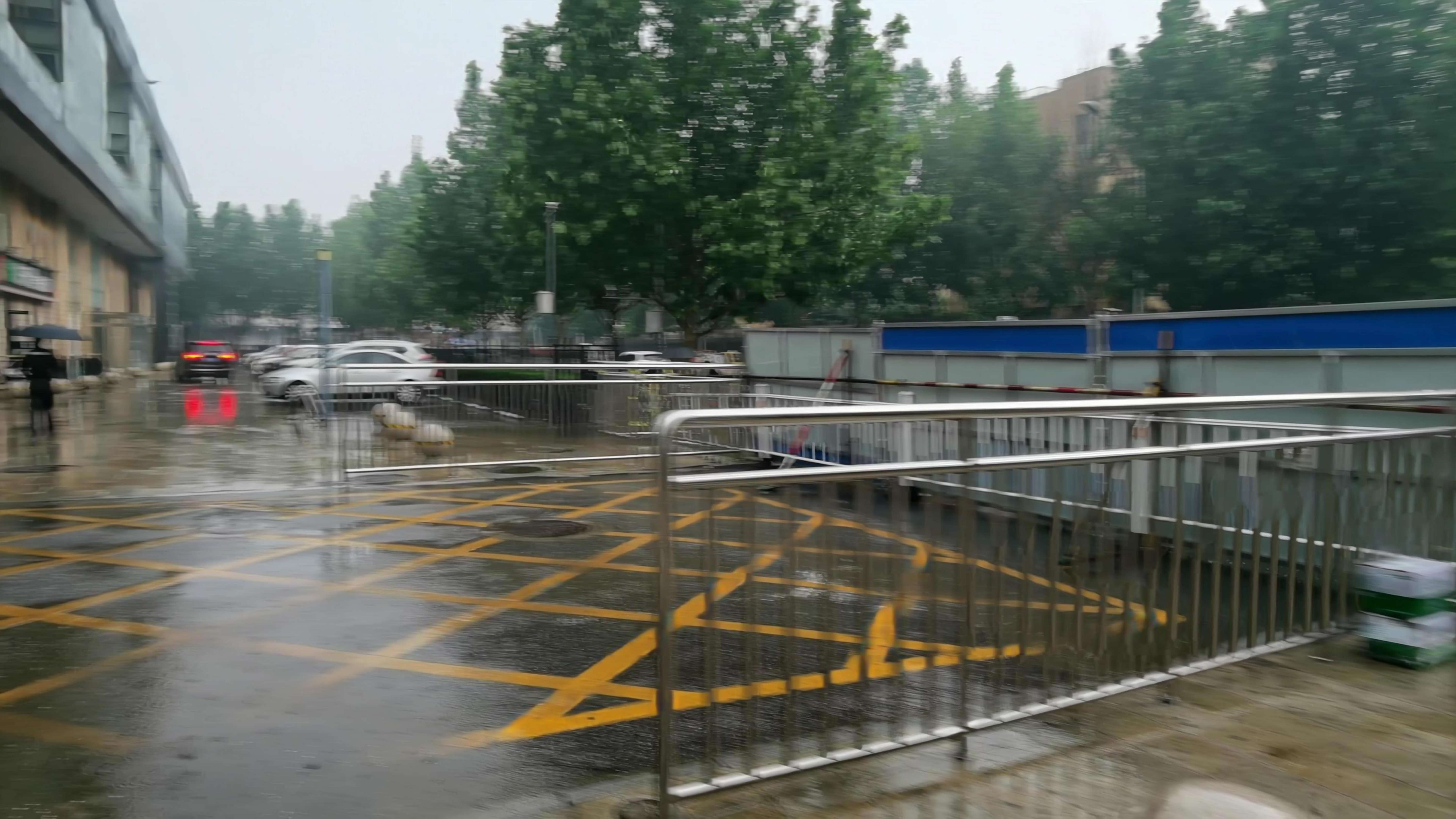}      & 
			\includegraphics[width=0.122\textwidth]{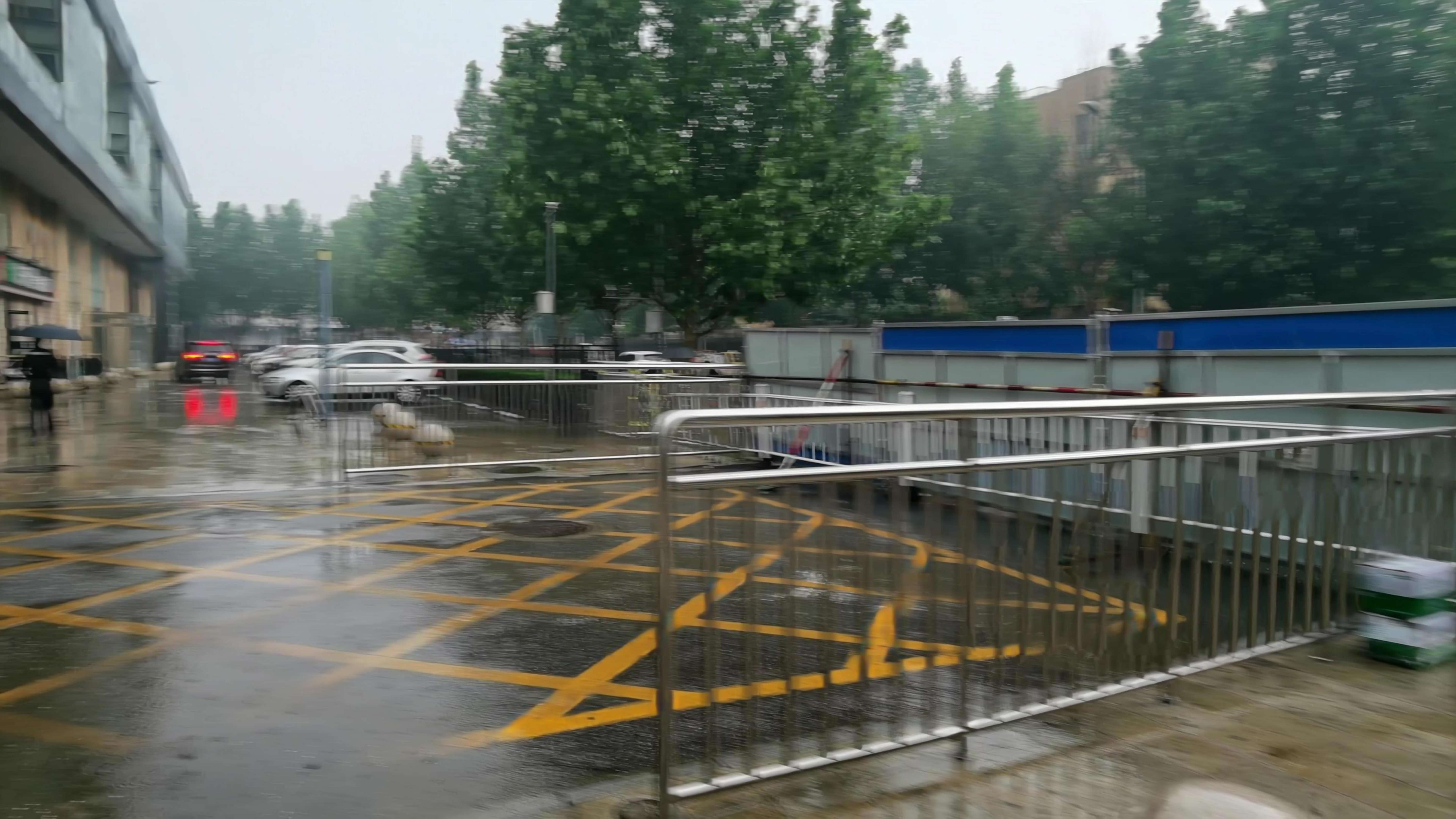}      & 
			\includegraphics[width = 0.122\textwidth]{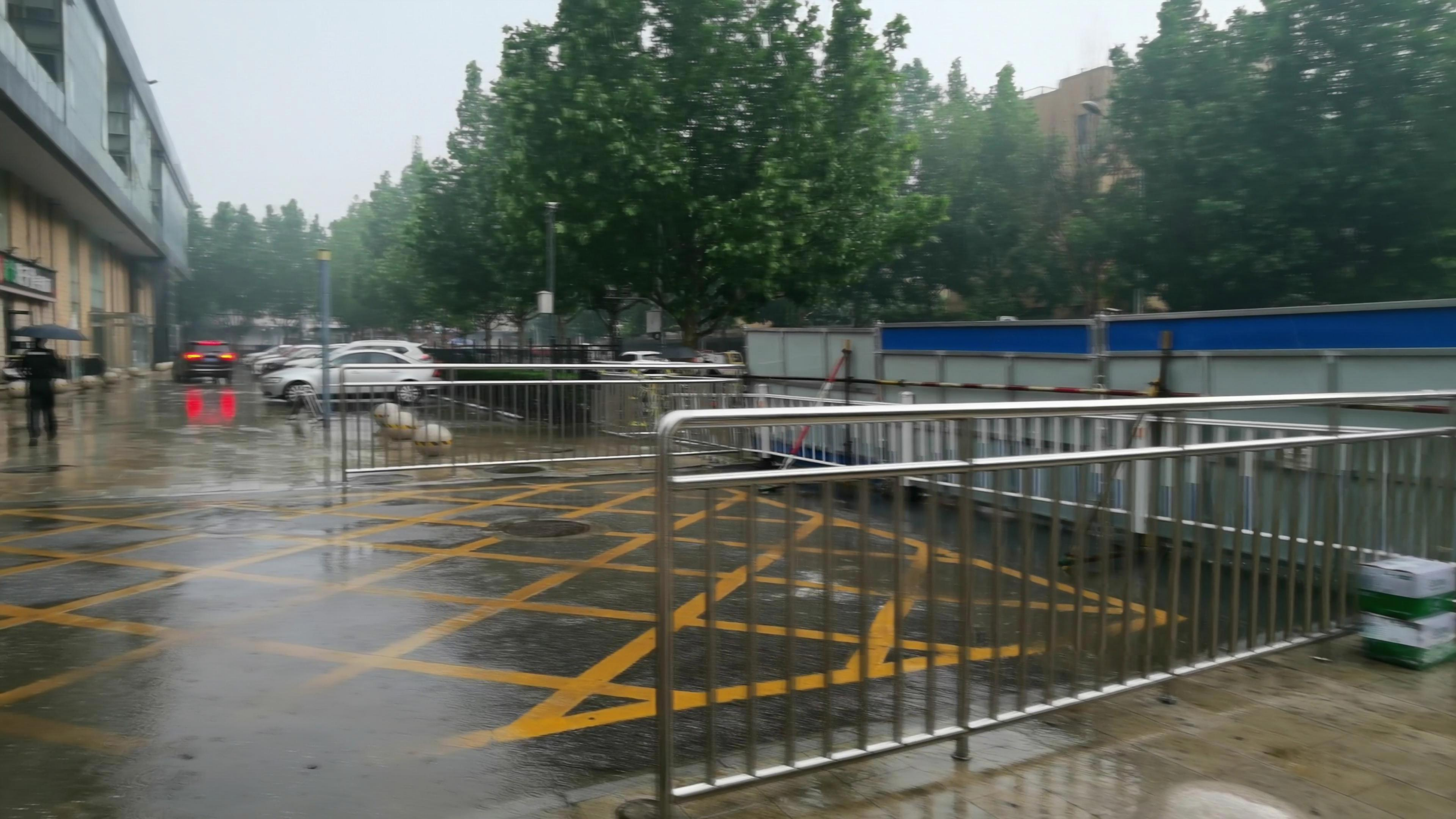}      \\
			
			\includegraphics[width=0.122\textwidth]{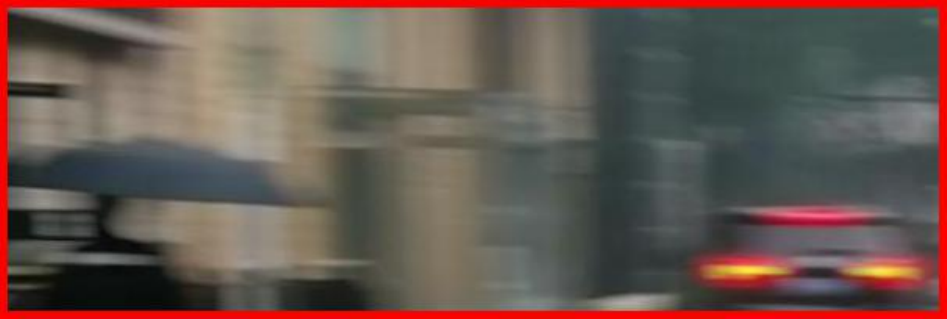}  &
			\includegraphics[width=0.122\textwidth]{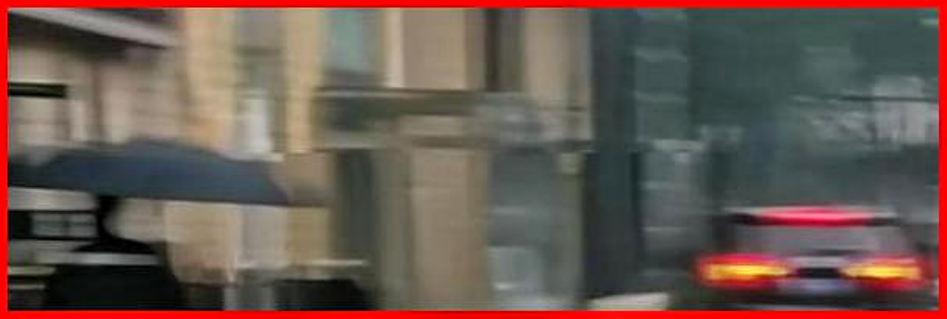}  &
			\includegraphics[width=0.122\textwidth]{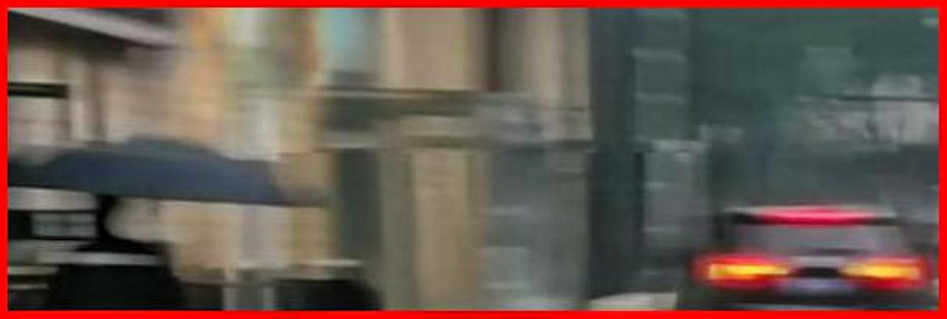}  &
			\includegraphics[width=0.122\textwidth]{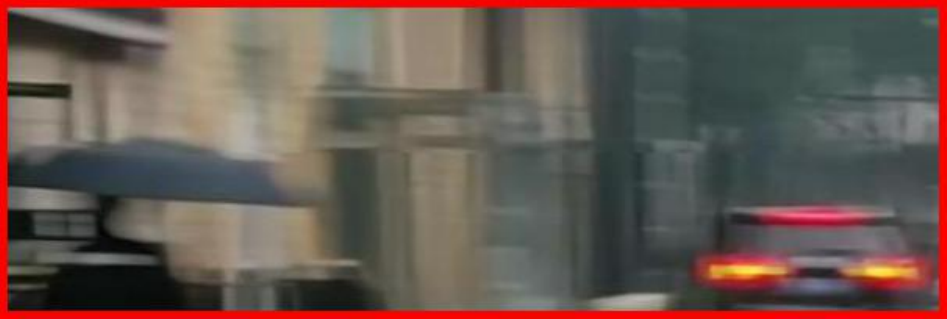}  &
			\includegraphics[width=0.122\textwidth]{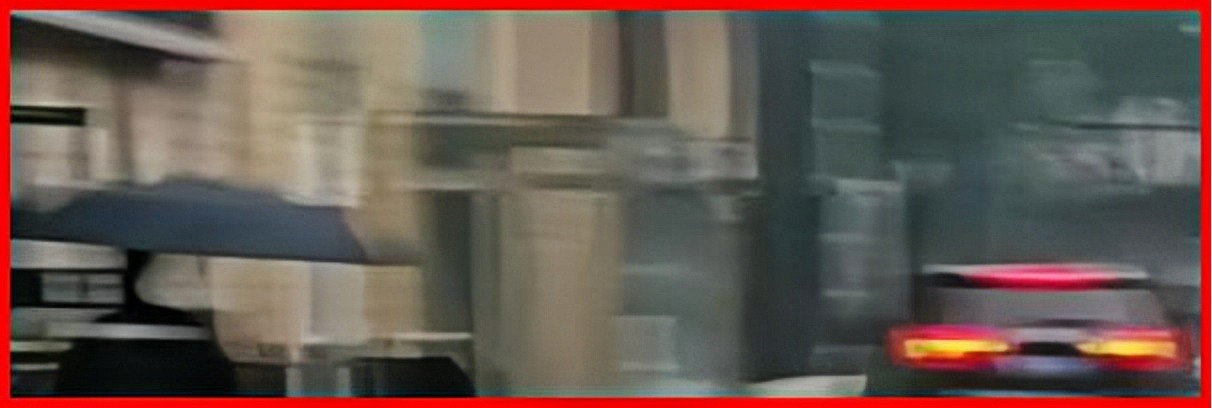}  &
			\includegraphics[width=0.122\textwidth]{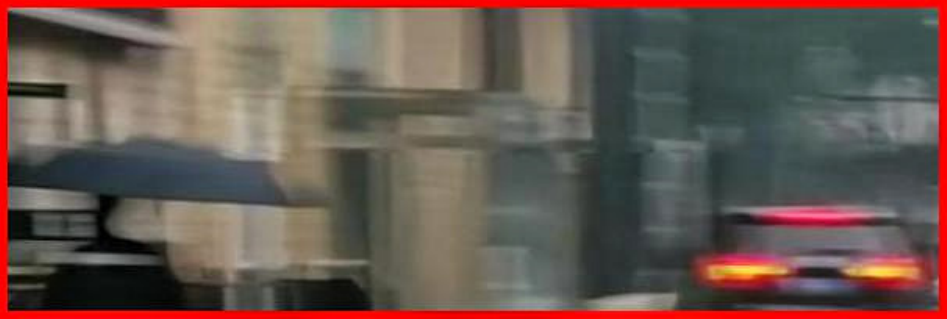}  &
			\includegraphics[width=0.122\textwidth]{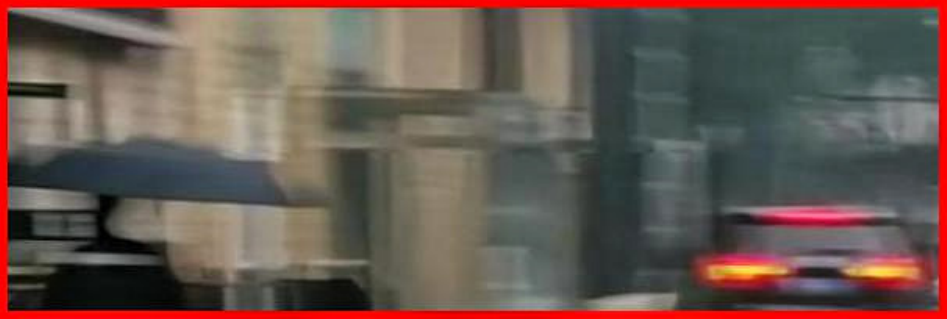}  &
			\includegraphics[width=0.122\textwidth]{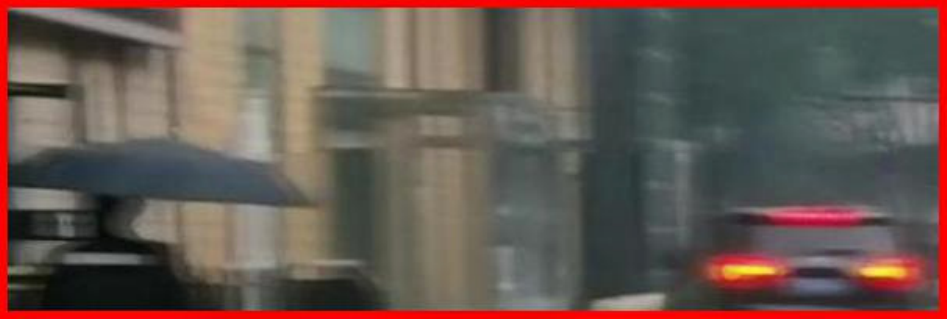}    \\

			(a) Input&
			(b) MPRNet& 
			(c) Restormer&  
			(d) NaFNet&    
			(e) MAXIM& 
			(f) MC-Blur& 
			(g) ASTL& 			
			(h) Ours    \\

		\end{tabular}
	\end{center}
	\vspace{-4mm}
	\caption{Deblurred results on real-world 4K resolution images. Our method recovers more image details compared with other state-of-the-art methods.}
	%\vspace{-2mm}
	\label{fig-ex4}
\end{figure*}
{\flushleft\textbf{Quantitative evaluation.}} 
The proposed method is evaluated on four datasets: test data of 4KRD, GoPro, RealBlur-J, and RealBlur-R. 
Sample results for the proposed method and compared approaches are shown in Figure~\ref{fig-ex1}. 
It can be observed that recent deep models~\citep  {chen2022simple,tu2022maxim,zamir2021restormer,zamir2021multi} still remain some residual blurs in the results. 
In contrast, the deblurred results generated by our algorithm are close to the ground truth images in Figure~\ref{fig-ex1}(f). 
Our method achieves the best performance compared to other state-of-the-art methods on the 4K dataset as shown Figure~\ref{fig-ex4}.
In addition, more detailed quantitative results on the 4KRD, RealBlur-J, RealBlur-R and GoPro datasets are reported in Table~\ref{tab-T-D}, which demonstrates the effectiveness of the proposed method. 
%Note that we selected 200 images with large differences in scene styles as test data for the GoPro dataset. These models usually have their own inductive bias, and a large number of scenarios in the test data are repeated. In order to evaluate the generalization ability of each model, we constructed this more balanced distributed test dataset (avoid the long tail problem).

{\flushleft\textbf{Qualitative evaluation.}} We also compare our algorithm against other methods on two real-world 4K blurry images, shown in Figure~\ref{fig-ex4}. While other state-of-the-art	methods produce varying degrees of residual blurs and artifacts, our algorithm yields sharper edges and contain fewer artifacts, shown in Figure~\ref{fig-ex4}(f).

%
%	\begin{table*}[!ht] \footnotesize 
	%		\begin{center}
		%		\caption{Quantitative evaluations on the 4KRD test data and the GoPro dataset in terms of PSNR, SSIM, and run time. The time consumption covers the time sampled on LR images.}
		%			\label{tab-T-D}
		%\resizebox{\textwidth}{!}{%
			%			\begin{tabular}{ccccccccc}
				%				\hline
				%				&      & \citep  {Chakrabarti16}  & \citep  {KupynMWW19}    & \citep  {NahKL17} & \citep  {SuinPR20}    & \citep  {TaoGSWJ18}     & \citep  {ZhangDLK19}    & Ours    \\ \hline
				%				\multirow{2}{*}{4KRD}          & PSNR & 22.3  & 25.64    & 25.22    & 28.90   & 25.13        & 24.98    & \textbf{30.08}         \\
				%				& SSIM & 0.708       & 0.763   & 0.759    & 0.831  & 0.741      & 0.757   & \textbf{0.842}   \\
				%				& Time(ms)& 1550      & 150       & 330        & 149       &127             & 40      & \textbf{21}         \\
				
				%				\multirow{2}{*}{GoPro}     & PSNR & 26.88  & 29.55    & 28.45     & 30.21   & 30.10        & 23.89    & \textbf{31.01}\\
				%				& SSIM & 0.899 & 0.934   & 0.917    & 0.944  & 0.932       & 0.901   & \textbf{0.945}  \\  
				%				& Time(ms)&    1249  & 89        & 270         & 81       &109               & 22      & \textbf{8} \\
				%				\hline
				%			\end{tabular}%
			%}
		%	\end{center}
	%	\vspace{-5mm}
	%	\end{table*}

% \usepackage{color}

\begin{table*}[!ht] \footnotesize 
	\centering
	%\vspace{-4mm}
	\caption{Quantitative evaluations on the 4KRD test data and other datasets in terms of PSNR, SSIM. 
		Although our method does not perform best on publicly available low-resolution datasets (GoPro, RealBlur-R, and RealBlur-J), it has a highly remarkable advantage on 4KRD datasets.
		\textbf{Bolded black} indicates first rank, \textcolor{red}{red} and \textcolor{blue}{blue} indicate second and third rank.	
		Our method is performed 5 $\times$ with cross-validation to obtain an accuracy result.}
	%\vspace{1mm}
	\label{tab-T-D}
	\begin{tabular}{ccccccccc} 
		\toprule
		%\hline
		& \multicolumn{2}{c|}{GoPro}                                   & \multicolumn{2}{c|}{RealBlur-R}                                  & \multicolumn{2}{c|}{RealBlur-J}                           & \multicolumn{2}{c}{4KRD}                                                              \\
		~ Method  & \multicolumn{1}{c|}{PSNR $\uparrow$}       & \multicolumn{1}{c|}{SSIM $\uparrow$}       & PSNR $\uparrow$                           & \multicolumn{1}{c|}{SSIM $\uparrow$}      & PSNR $\uparrow$                            & \multicolumn{1}{c|}{SSIM $\uparrow$} & PSNR $\uparrow$                            & SSIM $\uparrow$
		\\  
		\midrule%\hline
		
		~MPRNet~\citep  {zamir2021multi}   & 32.66                           & 0.959                           & \textcolor{blue}{\textbf{39.31}} & \textbf{0.972}                  & \textcolor{blue}{\textbf{31.76}}                           & \textcolor{blue}{\textbf{0.922}}                     & \textcolor{blue}{\textbf{28.64}}                           & \textcolor{blue}{\textbf{0.865}}                                                     \\ 
		
		Restormer~\citep  {zamir2021restormer} & \textcolor{blue}{\textbf{32.92}}                           & \textcolor{blue}{\textbf{0.961}}                           & 36.19                           & 0.957                           & 28.96                           & 0.879                     & 26.88                           & 0.830                                                     \\ 
		
		~ NaFNet~\citep  {chen2022simple}  & \textcolor{red}{\textbf{33.69}}                  & \textbf{0.967}                  & 37.69                           & \textcolor{blue}{\textbf{0.966}}                           & 29.39                           & 0.910                     & 27.19                           & 0.839                                                     \\ 
		
		~ MAXIM~\citep  {tu2022maxim}   & 32.86                           & \textcolor{blue}{\textbf{0.961}}                           & \textcolor{red}{\textbf{39.45}}                           & 0.935                           & \textcolor{red}{\textbf{36.15}}                  & \textcolor{red}{\textbf{0.949}}                     & \textcolor{red}{\textbf{30.19}} & \textcolor{red}{\textbf{0.891}}                           \\  
		~ MC-Blur~\citep{9857414} & 31.80 & 0.948 & 35.85 & 0.950 & 28.70 & 0.870 & 26.50 & 0.820 \\ 
		~ ASTL~\citep{8449842} & 32.25 & 0.936 & 37.80 & 0.964 & 30.25 & 0.910 & 27.80 & 0.845 \\
		\midrule  
		~ ~ Ours  & \textbf{33.79} & \textcolor{red}{\textbf{0.962}} & \textbf{39.66}                  & \textcolor{red}{\textbf{0.969}} & \textbf{36.75} & \textbf{0.953}            & \textbf{33.89}                  & \begin{tabular}[c]{@{}l@{}}\textbf{0.916}\\\end{tabular}  \\ 
		%\hline
		\bottomrule
	\end{tabular}
	%\vspace{-4mm}
\end{table*} 

\begin{table*}[!ht] \footnotesize
	\begin{center}
		\caption{Effectiveness of the real/imag. part, cubic-mixer, multi-scale, slicing scheme, and the local feature extraction. Quantitative results demonstrate the effectiveness of each module.}
		%\vspace{1mm}
		\label{Withouts}
		\begin{tabular}{ccccccc}
			\toprule
			& d-real 	& d-imag.	& w/o CM  &w/o MS &w/o SS &w/o LFE             \\ 
			\midrule
			PSNR   & 27.65 $\pm$ \textbf{\scriptsize{0.11}} & 25.87 $\pm$ \textbf{\scriptsize{0.08}}  & 31.44 $\pm$ \textbf{\scriptsize{0.15}}   &32.49  $\pm$  \textbf{\scriptsize{0.12}}  &31.98 $\pm$  \textbf{\scriptsize{0.09}}   &33.52 $\pm$  \textbf{\scriptsize{0.009}}  \\
			SSIM  & 0.877 $\pm$ \textbf{\scriptsize{0.009}}  & 0.859 $\pm$ \textbf{\scriptsize{0.010}}   & 0.939 $\pm$ \textbf{\scriptsize{0.008}}   &0.942  $\pm$ \textbf{\scriptsize{0.006}} &0.953  $\pm$  \textbf{\scriptsize{0.007}} &0.960  $\pm$  \textbf{\scriptsize{0.002}}  \\ 
			\bottomrule
		\end{tabular}%
		%\vspace{-4mm}
	\end{center}
\end{table*}

\subsection{Ablation Study}
\begin{figure*}[t]\footnotesize
	\begin{center}
		\tabcolsep 1pt
		\begin{tabular}{@{}ccccc@{}}
			\includegraphics[width = 0.2\textwidth]{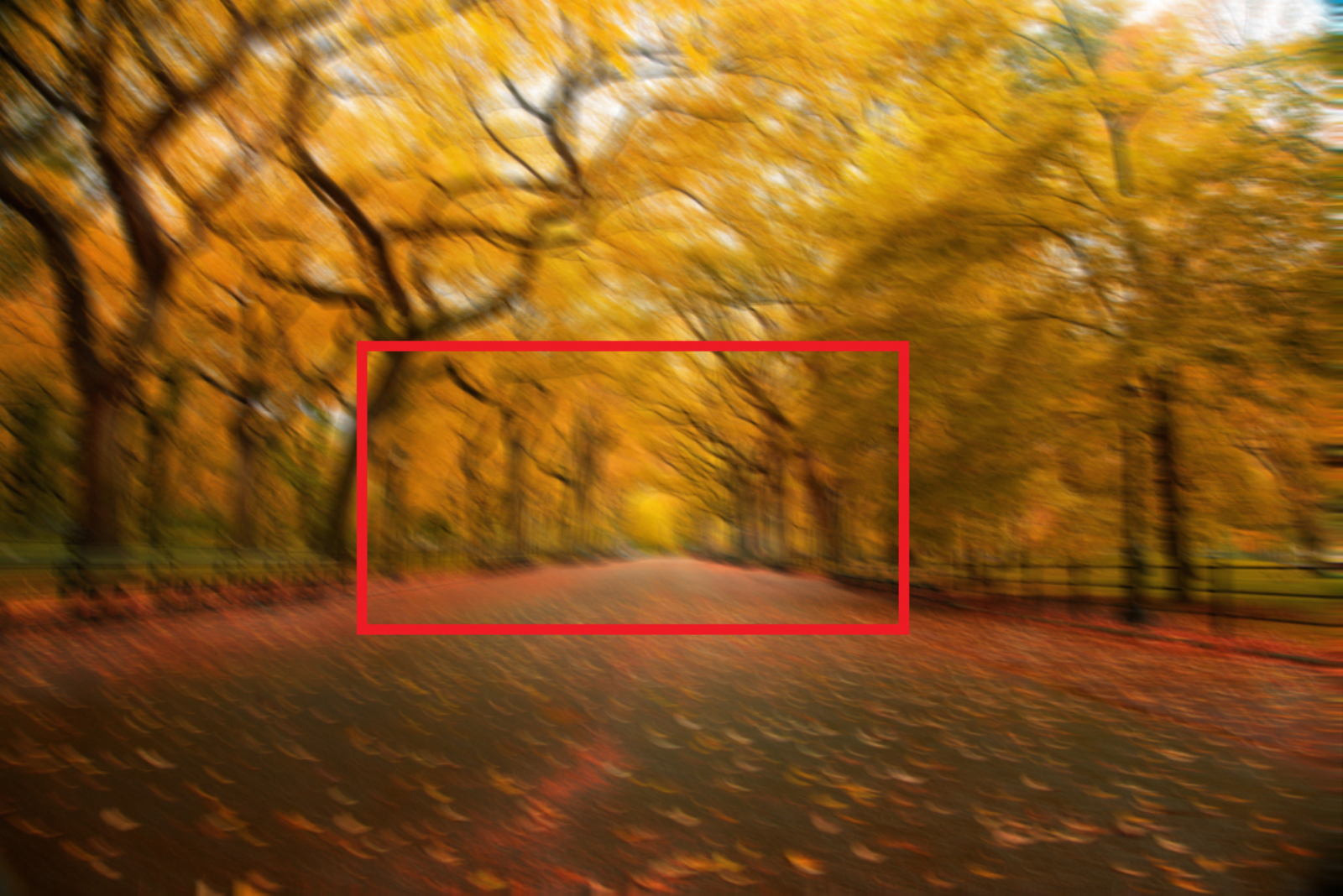}                 &
			\includegraphics[width = 0.2\textwidth]{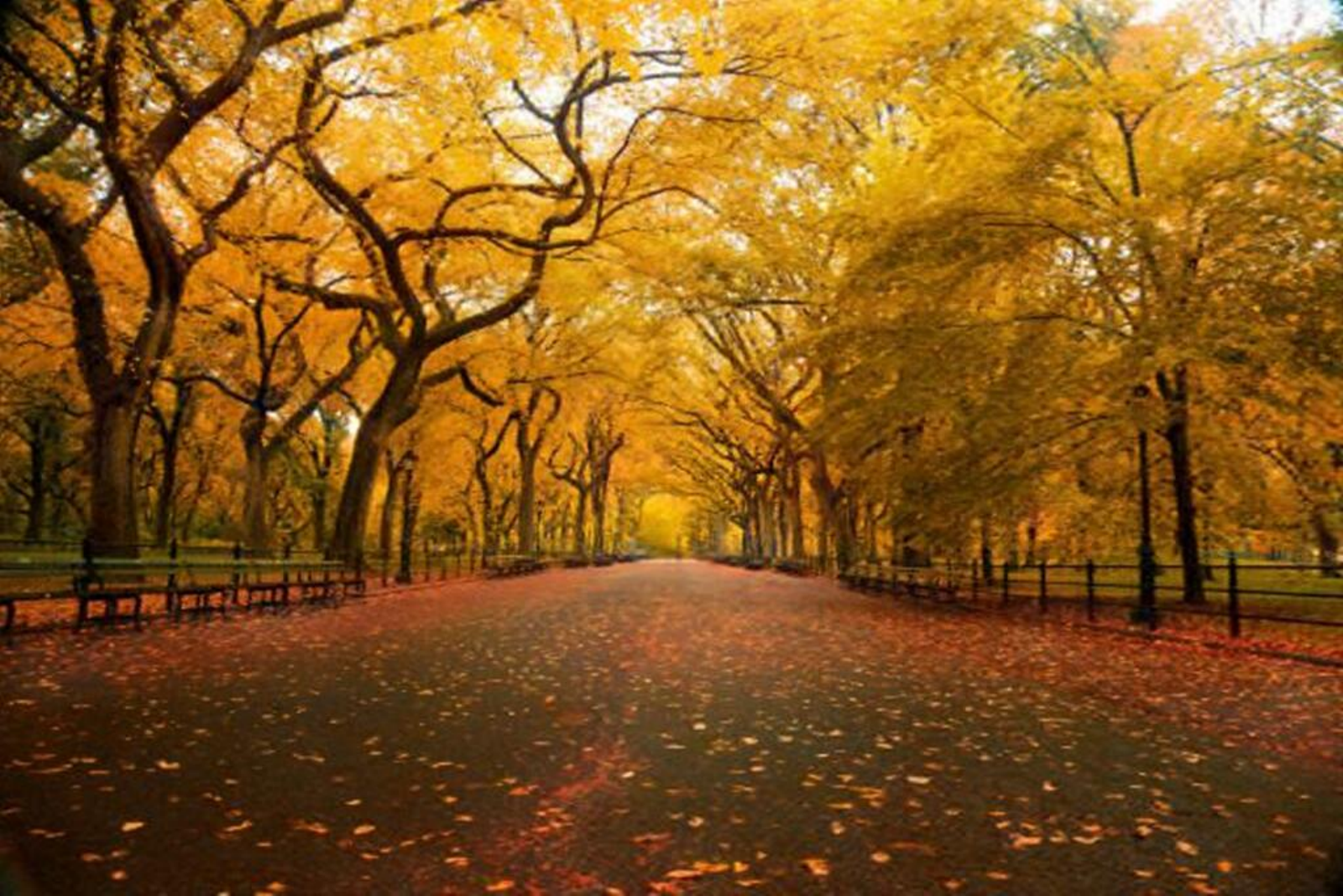}                &
			\includegraphics[width = 0.2\textwidth]{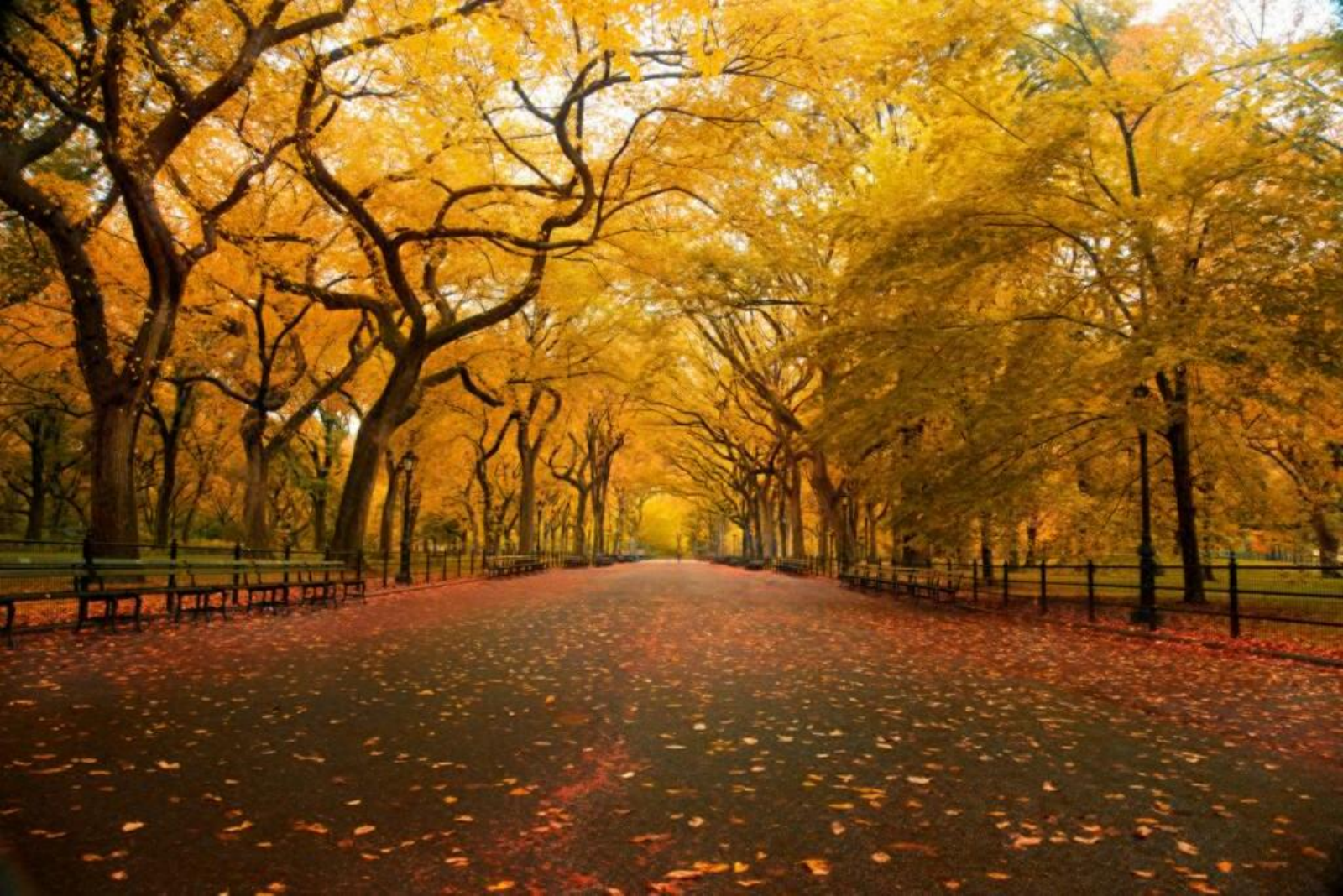}                 &
			\includegraphics[width = 0.2\textwidth]{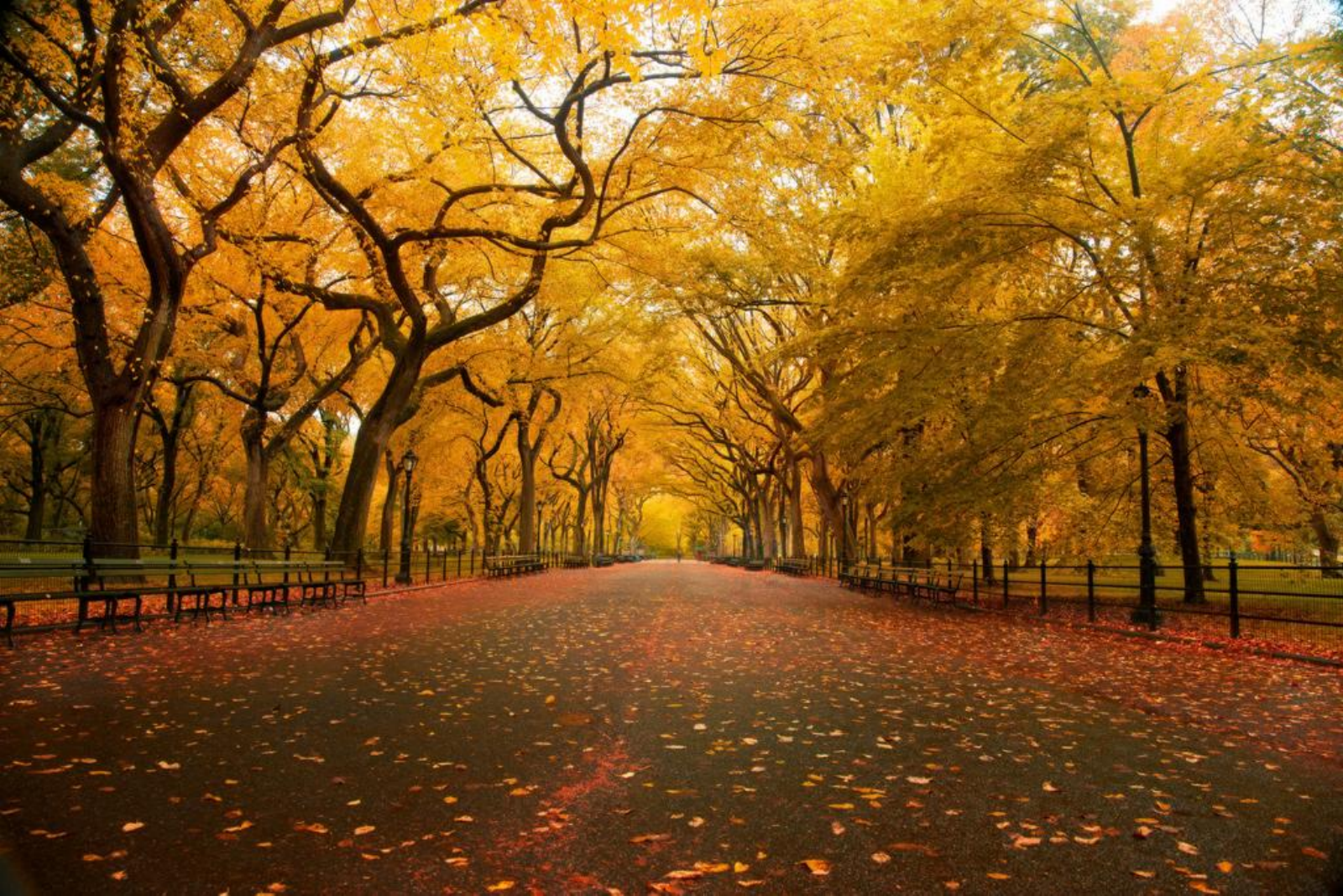}                 &
			\includegraphics[width = 0.2\textwidth]{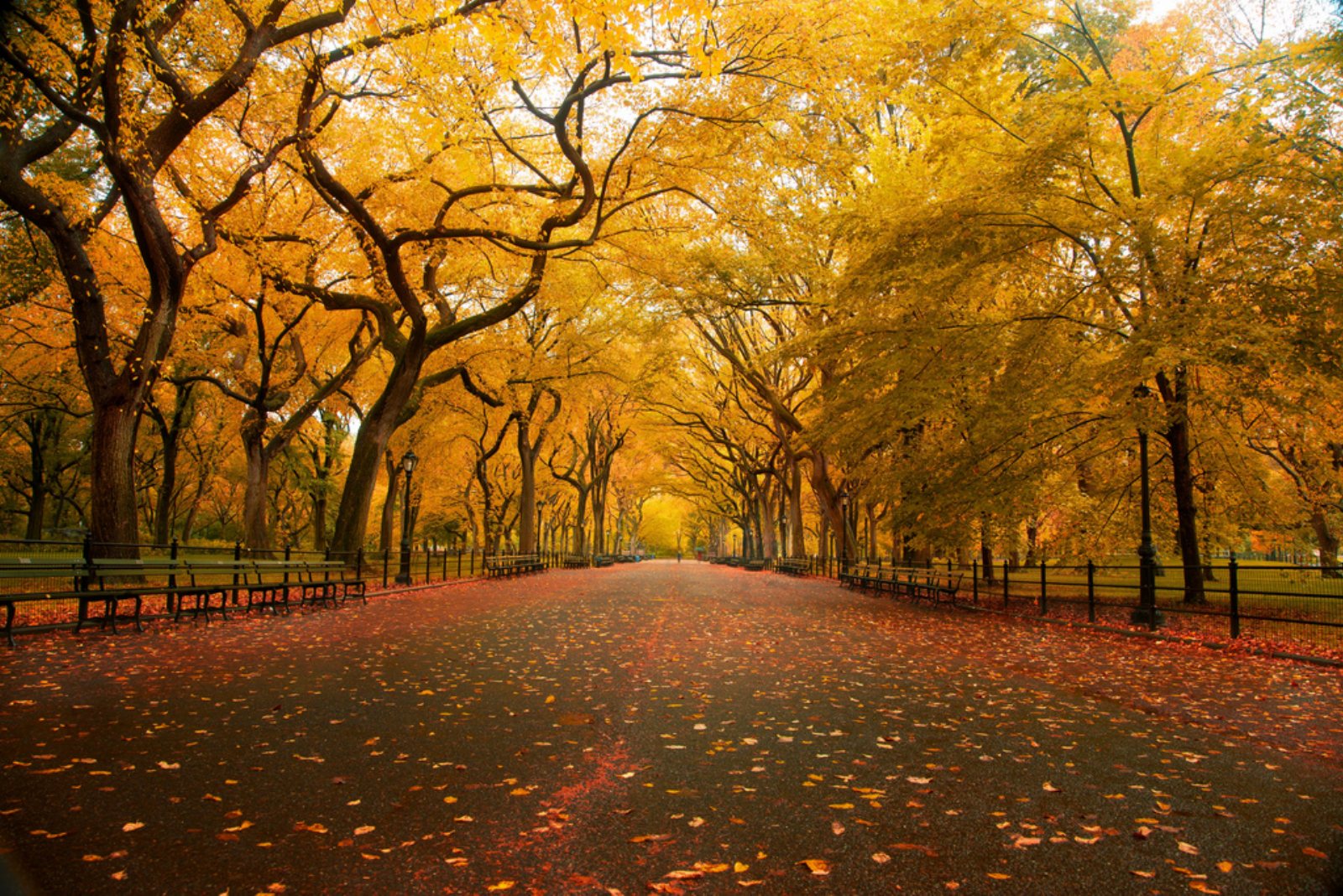}                                                    \\
			
			\includegraphics[width = 0.2\textwidth]{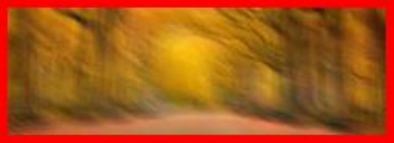}                 &
			\includegraphics[width = 0.2\textwidth]{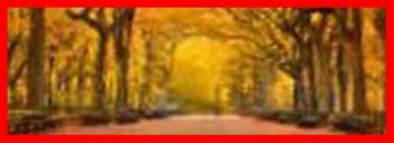}                &
			\includegraphics[width = 0.2\textwidth]{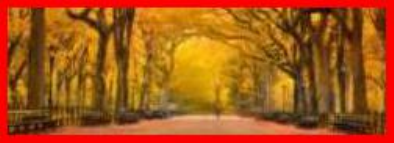}                 &
			\includegraphics[width = 0.2\textwidth]{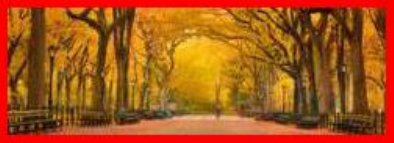}                 &
			\includegraphics[width = 0.2\textwidth]{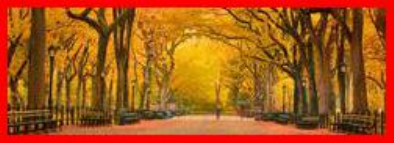}               \\				
			
			(a)  Input  &
			(b)  w/o CM   &
			(c)  w/o SS   &  
			(d)  Ours     & 
			(e)  GT  \\
		\end{tabular}
	\end{center}
	\vspace{-4mm}
	\caption{Our method obtains better visual quality and recovers more image details compared with other ablation experiments.
		Note that we do not show  w/o MS and  w/o LFE because the quantitative results are very close and almost visually indistinguishable.}
	\vspace{-2mm}
	\label{fig-data_D}
\end{figure*}

To demonstrate the effectiveness of each module in the proposed network, we perform ablation studies involving the following five experiments.
In addition, as shown in Figure~\ref{fig-data_D}, we select a low-resolution image for the ablation experiment to demonstrate.

\noindent \textbf{Effectiveness of real/imaginary part.}
To fairly verify the role of real and imaginary parts with the WFP framework, we run the input of double the real part information (named d-real) or double the imaginary part information (named d-imag.) with the WFP on the GoPro dataset.
Table~\ref{Withouts} demonstrates that the effectiveness of only relying on the information of the real or imaginary parts of the Fourier coefficients to reconstruct the blurred image is discounted.

\noindent \textbf{Effectiveness of cubic-mixer.}
%As shown in Figure~\ref{time_cost}, 
To evaluate the effectiveness of the proposed cubic-mixer, we use the U-Net to replace the proposed cubic-mixer using the WFP framework (named w/o CM). 
Table~\ref{Withouts} demonstrates the effectiveness of the proposed cubic-mixer on the GoPro dataset, which demonstrates that cubic-mixer generates richer textural and color.

\noindent \textbf{Effectiveness of multi-scale.}
We retain only the top path in the network (named w/o MS), and to be fair, we add some cubic-mixers to the top path to maintain the balance of computational complexity.
Note that we can only add 6 cubic-mixers to the top path due to the large resolution of the image.
Table~\ref{Withouts} demonstrates the effectiveness of the proposed multi-scale on the GoPro dataset, which demonstrates that multi-scale techniques can boost the performance of the model.

\noindent \textbf{Effectiveness of slicing scheme.} 
We replace the slicing scheme (named w/o SS) with bilateral grid learning and compare it with our proposed algorithm.
Specifically, after local feature extraction we use a downsampling method to generate the biggest bilateral grid ($12 \times 64 \times 16 \times 16$) that can be processed on a single GPU with 24G RAM, and then apply and slice the raw image.
Table~\ref{Withouts} also demonstrates the effectiveness of the proposed slicing scheme.

\noindent \textbf{Effectiveness of local feature extraction.}
We use $1 \times 1$ convolution to replace the $3 \times 3$ convolution (named w/o LFE) in the part of local feature extraction. 
Table~\ref{Withouts} demonstrates the effectiveness of the $3 \times 3$ convolution in the local feature extraction.

\begin{figure*}[t]\footnotesize
	\begin{center}
		\tabcolsep 1pt
		\begin{tabular}{@{}ccccc@{}}
			\includegraphics[width = 0.185\textwidth]{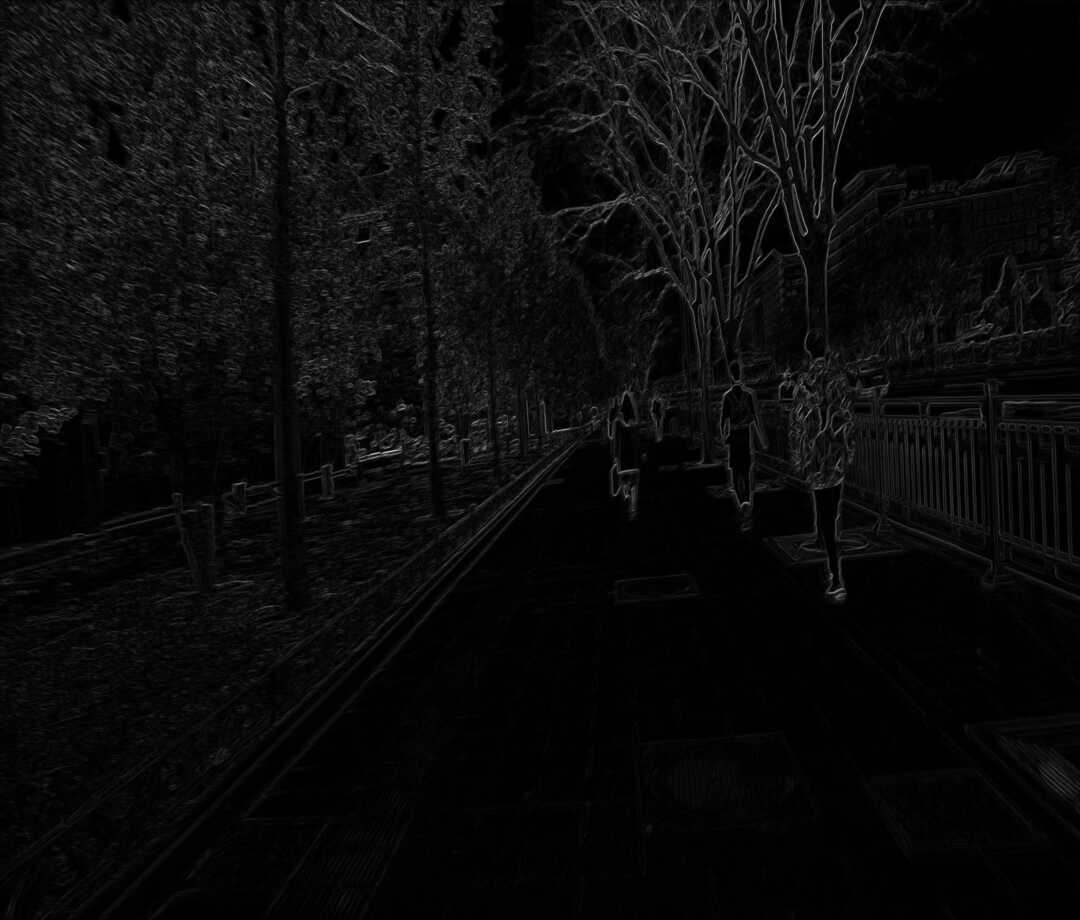}                 &
			\includegraphics[width = 0.185\textwidth]{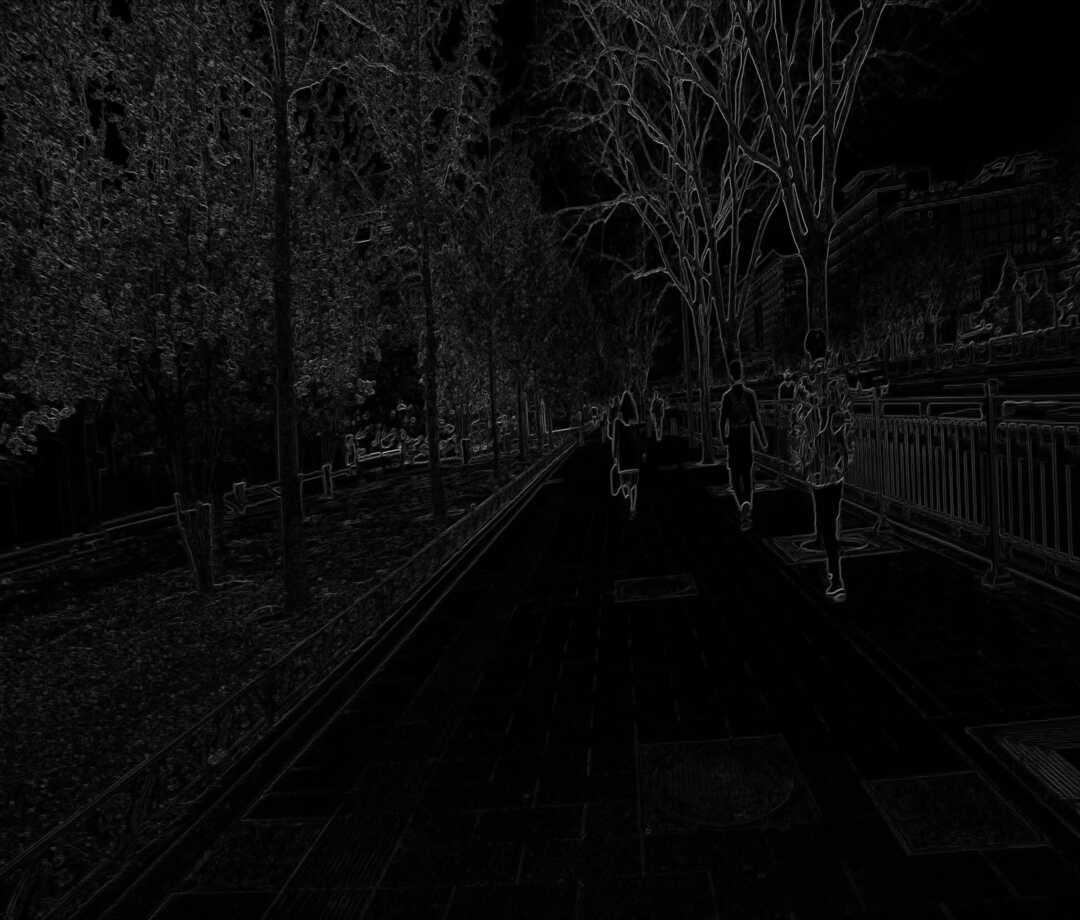}                &
			\includegraphics[width = 0.185\textwidth]{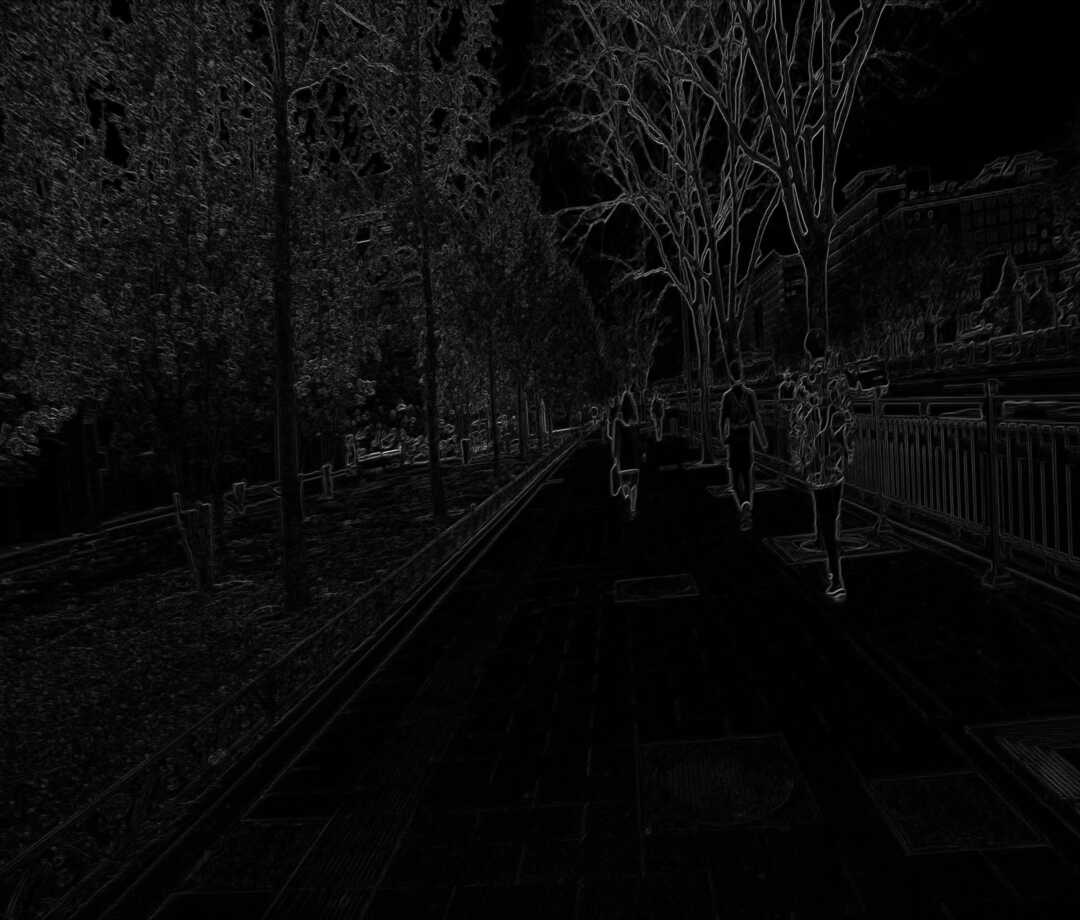}                 &
			\includegraphics[width = 0.185\textwidth]{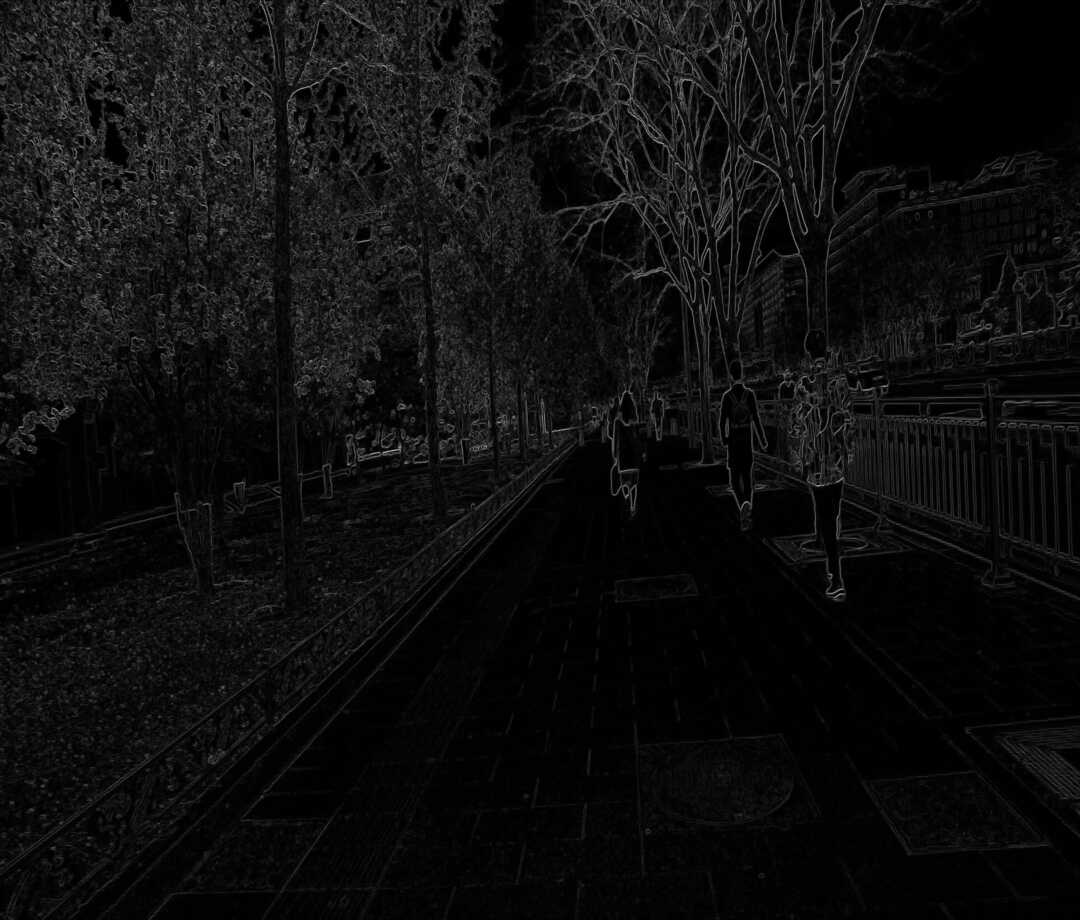}                 &
			\includegraphics[width = 0.185\textwidth]{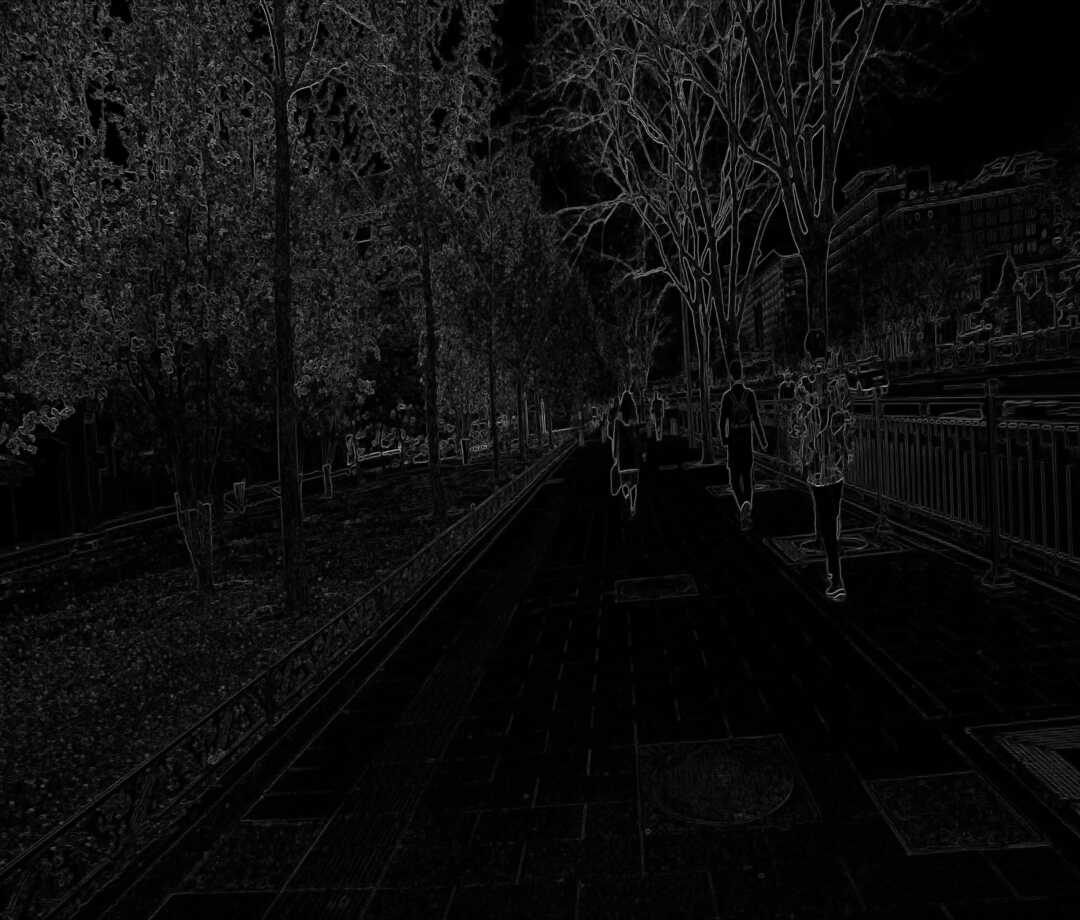}        \\
			(a)  Input  &
			(b)  w/o SS   &
			(c)  Grided Slicing   &  
			(d)  Ours     & 
			(e)  GT  \\
		\end{tabular}
	\end{center}
	\vspace{-4mm}
	\caption{Our method produces sharper edges and restores finer details compared to other approaches.}
	\label{fig-slicing}
\end{figure*}
\vspace{-4mm}

Furthermore, we combine the comparison methods~\citep  {chen2022simple,tu2022maxim,zamir2021restormer,zamir2021multi} with the slicing scheme to verify the effectiveness on the 4KRD dataset.
As shown in Figure~\ref{fig-slicing}, our approach produces sharper edges, fewer artifacts, and better fine detail preservation compared to grid slicing schemes in the restored images.

%\begin{figure*}[t]
%	\centering
%	\includegraphics[width=0.8\textwidth]{plot/main_plot/time_cost.pdf}
%	\vspace{-2mm}
%	\caption{(a) demonstrates that using the proposed MC-Mixer in Siamese-Mixer has a higher SSIM value than other deblurring methods in the SFDP framework. (b) shows the improvement in prediction accuracy of these networks using SFDP over DDU on 4KRD dataset. Our model achieves the best performance on both GoPro and 4KRD datasets.}
%	\label{time_cost}
%\end{figure*}
%
%

%We resize the 4K blurry input to the resolution close to the corresponding maximum size of each deblurring model can handle. 
%We find that these networks show significant improvement in UHD image deblurring using the DFDP framework compared to the DDU.
%In addition, these networks also show significant improvement in HR image (GoPro dataset) deblurring using the DFDP framework compared to full-resolution deblurring (i.e., without DDU). 

%The LR deblurred images obtained using the DFDP framework on the 4KRD dataset are directly trilinearly interpolated to recover the original resolution. 
%In Figure~\ref{fig-data_D}, we visualize the image effects of our method comparing other network methods for deblurring in the DFDP framework.
\begin{table}[!ht]
	\scriptsize
	\centering
	\caption{Performance on DPDD (defocus blur) and Rain100L (derain) datasets.}
	\label{tab:additional_experiments}\vspace{-2mm}
	\begin{tabular}{lccc}
		\toprule
		\textbf{Method} & \textbf{Dataset} & \textbf{PSNR} ($\uparrow$) & \textbf{SSIM} ($\uparrow$) \\
		\midrule
		IFAN~\citep{Lee2021Iterative}   & DPDD  & 25.37 & 0.789 \\
		Restormer~\citep{zamir2021restormer} & DPDD  & 25.98 & 0.811 \\
		\textbf{Ours}  & DPDD  & \textbf{27.11} & \textbf{0.855} \\
		\midrule
		MPRNet~\citep{zamir2021multi}   & Rain100L & 36.40 & 0.965 \\
		Restormer~\citep{zamir2021restormer} & Rain100L & 38.99 & 0.978 \\
		\textbf{Ours}   & Rain100L & \textbf{38.82} & \textbf{0.977} \\
		\bottomrule
	\end{tabular}\vspace{-2mm}
\end{table}

%\vspace{-2mm}
\subsection{Run Time and potential of the model}
%\vspace{-2mm}
In terms of run time, the proposed multi-scale cubic-mixer model performs favorably against all the comparison deblurring approaches~\citep  {chen2022simple,tu2022maxim,zamir2021restormer,zamir2021multi}. 
All the approaches are evaluated on the same machine with an Inter(R) Xeon(R) CPU and an NVIDIA Tian RTX 3090 GPU. 
Note that the time is only the processing time of the GPU without considering the run time of I/O operations. 
With the SDS method, it takes 100 ms for MPRNet, 124 ms for Restormer, 98 ms for NaFNet and 279 ms for MAXIM to run a 4K resolution image. 
These deep models are clearly less efficient than our algorithm.
%\vspace{-2mm}

\noindent \textbf{Other task.}
To assess the generalization ability of our method, we conducted experiments on the DPDD dataset (defocus blur) and Rain100L (derain). These datasets evaluate the robustness of our approach beyond UHD image deblurring. 
As shown in Table~\ref{tab:additional_experiments}, our method outperforms existing approaches in both deraining and defocus blur removal tasks.  

\noindent \textbf{Cubic-Mixer vs. Wave-MLP on ImageNet.}
%
%\vspace{-2mm}
So far, we attempt to use cubic-mixer in an image classification task to verify its effectiveness.
We conduct image classification experiments on ImageNet, which contains about 1.3M training images and 50k validation images from 1000 classes.
Our model is trained for 300 epochs with Adam optimizer, whose learning rate is initialized as 0.002 and declines with a cosine decay scheme.
In addition, we use the data augmentation strategies.
To be fair, we use a training technique consistent with Wave-MLP, and four frameworks with close parameter numbers (T, S, M, and B).
\begin{table*} [!ht] \scriptsize
	\centering
	%\vspace{-4mm}
	\caption{Comparison of the proposed cubic-mixer architecture with Wave-MLP model on ImageNet.}
	%\vspace{1mm}
	\begin{tabular}{cccc|cccc} 
		\toprule
		Model      & Params. & FLOPs & Top-1 acc. (\%) & Model         & Params & FLOPs & Top-1 acc. (\%)~  \\ 
		\midrule
		Wave-MLP-T & 17M     & 2.4G  & \textbf{80.6}   & Cubic-Mixer-T & 16.5M  & 2.2G  & 80.5              \\
		Wave-MLP-S & 30M     & 4.5G  & 82.6            & Cubic-Mixer-S & 31M    & 4.4G  & \textbf{82.7}     \\
		Wave-MLP-M & 44M     & 7.9G  & 83.4            & Cubic-Mixer-M & 40M    & 7.6G  & \textbf{83.5}     \\
		Wave-MLP-B & 63M     & 10.2G & \textbf{83.6}   & Cubic-Mixer-B & 59M    & 9.9G  & \textbf{83.6}     \\
		\bottomrule
	\end{tabular}
	\label{imagenet}
\end{table*}
%
%All the experiments are conducted with PyTorch 1.10  on the RTX 3090 GPU.
As reported in Table~\ref{imagenet}, our approach exhibits competitiveness, as described in detail in the supplementary material.

\section{Limitations, Social Implications and Application}
%\vspace{-2mm}
Our model is focused on the image deblurring task, and we have tried to perform it on other image restoration tasks (such as dehazing, deraining, etc.), but it does not perform well when it comes to color restoration requirements.
This may be due to the slicing scheme biasing towards the inductive bias property of the texture.
In addition, our dataset does not involve sensitive information, and portrait information is solicited.
In the end, we conduct a test task of object detection on the comparison method with the help of Google's API, and the experimental results show that our method helps to improve the performance of the downstream task.

\section{Related Work}
%\vspace{-2mm}
%Deblurring of a single 4K image is an ill-posed problem, and most of the current solutions are implemented in the frequency domain and the spatial domain. However, the current model solving strategy relies on deep convolutional layers with a large number of pooling layers, which obviously costs a lot of computational resources. Recently, Transformer has performed well on low-level image processing tasks and does not require deep network stacking.

{\flushleft \textbf{Single image deblurring in the spatial domain.}} 
%Currently, most image deblurring methods directly act on the spatial domain of the image. 
%To solve the ill-posedness of the image sharpening problem, traditional methods have made different assumptions to model the sharp image prior.
%Perrone et al.~\citep  {PerroneF14} propose to use of sparse gradient priors via total variation.  Dong et al.~\citep  {DongZSW11} introduces two adaptive regularization terms in the sparse representation framework. 
%This method introduces linear operators to calculate the local maximum gradient and an effective optimization scheme, which can be applied to a variety of complex scenarios.
%
Many assumptions and priors for image deblurring have been developed including gradient priors~\citep  {ChenFWZ19,McCloskeyL09,PerroneF14}, sparse representation priors~\citep  {DongZSW11,KrishnanTF11,XuZJ13}, $l_{0}$-norm regularizers~\citep  {LiPLGS018,XuZJ13}, patch priors~\citep  {MichaeliI14,SunCWH13}.
However, these prior-based methods are not always applicable in dynamic scenes that contain depth variations and moving objects.

Recently, due to the success of deep learning in computer vision, many CNN-based approaches have also been proposed for image deblurring~\citep  {chen2022simple,GaoTSJ19,KimLSH17,KupynBMMM18,KupynMWW19,SuinPR20,TaoGSWJ18,zamir2021multi,ZhangDLK19,ZhangPRSBL018}. 
%
%CNN-based approaches with multiple scales are also frequently proposed, 
%Nah et al.~\citep  {NahKL17} propose to start with deblurring the input at a coarse-scale and then gradually deblurring it at a finer scale until a clear image is recovered.
%
%Tao et al.~\citep  {TaoGSWJ18} build a scale-recurrent network (SRN) that uses shared gradients in the training process. 
%The method can incorporate recurrent modules and benefits restoration across scales.
%
%Gao et al.~\citep  {GaoTSJ19} propose an effective network by sharing parameters on each scale with skip connections. 
%
%Although the above methods achieve reliable accuracy, the speed of model inference is limited.
%
%Recently, two methods of learning from local features and finally converging to the global perspective is proposed to achieve image deblurring~\citep  {SuinPR20, ZhangDLK19}. These methods performs computation acceleration in processing a single blurred image.  
%
In addition to CNNs, Transformer-based approaches~\citep  {chen2021pre,liang2021swinir,tu2022maxim,wang2021uformer,zamir2021restormer} are also used for image deblurring. 
%However, these models usually involve a large number of network parameters and a long processing time, which cannot meet the increasing demand for real-time deblurring. 
However, these methods pay attention only to the spatial domain and ignore the frequency domain information in image deblurring. 
In addition, all these spatial domain based deblurring methods are computationally expensive.
%In addition, these methods do not consider the nature of the deconvolution and neglect the problem that the blurred areas of the image are the high frequency details lost.

%\vspace{-2mm}
{\flushleft \textbf{Single image deblurring in the frequency domain.}} 
There are also some methods achieve image deblurring in the frequency domain.
%Anarim et al.~\citep  {AnarimUI96} proposes to achieve image deblurring in the frequency domain of an image without prior knowledge using the FT combined with the maximum expectation method, which is a significant improvement over the methods that uses prior knowledge as a constraint.
%
%Recently, there are also some methods that combine deep networks with discrete Fourier transforms.
%
%
%Xu et al.~\citep  {XuRLJ14} propose to let
Wiener deconvolution is a commonly used non-blind linear image recovery algorithm that performs deblurring by converting spatial domain to frequency domain~\citep  {LevinWDF09}.
In addition, several CNN-based deep network methods are proposed to act on the image frequency domain by FFT.
Delbracio et al.~\citep  {DelbracioS15} proposed a Fourier aggregation method based on \citep  {DelbracioS15_cvpr} to recover a clear image in videos with multiple adjacent frames.
Chakrabarti~\citep  {Chakrabarti16} used a CNN to learn an affine matrix in the frequency domain and then multiplied the affine matrix with the frequency domain features to recover a clear image by inverse Fourier transform (IFT). 
% 卷积的天然问题
These methods have difficulty in establishing long dependencies on the Fourier coefficients due to the limited capacity of the model.
% Recently a new alternative has been proposed that is to use only the MLP network to perform computer vision tasks.
%This method requires only a small number of fully connected layers to focus on global information.
%In this work we focus on MLP to solve the natural limitations of convolutional operators.

%\vspace{-2mm}
{\flushleft \textbf{MLP on the visual task.}} 
Recently, there are many attempts to explore the benefits of MLP in computer vision tasks including image classification~\citep  {ChenRC0JLS20,DosovitskiyICLR,KolesnikovBZPYG20,TolstikhinCoRR}, object detection~\citep  {ZhuCoRR}, recognition~\citep  {ZhaoJK20}, and low-level image tasks~\citep  {chen2021pre,liang2021swinir,tu2022maxim,wang2021uformer,Wu0LCW20,zamir2021restormer}.
%
%However, few related works focus on low-level vision tasks by using long dependencies in MLP.
However, these methods usually need a fine-grained CNN stem allowing the subsequent model to build long-range dependencies, which can easily cause visible streaks in the enhanced image. 
In this paper, we explore a novel MLP-based image deblurring approach without CNN stem to establish long-range dependencies on the image frequency domain.

%\vspace{-2mm}
{\flushleft \textbf{UHD image enhancement.}}
Some approaches \citep  {GharbiCBHD17,HeWSD20,WangZFSZJ19,wang2021real,XiaZXSFKC20,yuan2021hrformer,ZengPAMI,ZhangT20} have been proposed to reconstruct high-resolution images in real-time.
The bilateral filter/grid has attracted long-term attention in its acceleration~\citep  {BarronP16,Adams16a,ChenXK17,zheng2021ultra}, which is an edge-aware manipulation of images in the bilateral space~\citep  {BarronASH15,WangZFSZJ19,XiaZXSFKC20}. 
In this paper,  we regress a full-resolution affine transformation tensor directly from the tail of the network.

\vspace{-2mm}
{\flushleft \textbf{Multi-scale fusion.}}
Recently multiscale image fusion schemes~\citep  {chen2021crossvit,ke2021musiq,li2021infrared,zhang2021multi} are still widely adopted due to its thorough consideration of the texture and color of the image on different perceptual fields.
In contrast, the multiscale network is used to recover a tensor with attention characteristics instead of recovering an image directly.
%

%\vspace{-2mm}
\section{Discussion and Conclusion}
%\vspace{-2mm}
Executing an image by multi-scale cubic-mixer can be regarded as learning one dimension of the image information with attention characteristics, and its efficiency advantage comes from the computational resources applied on low density images.
In this paper, we propose a multi-scale cubic-mixer for arbitrary size image deblurring. 
Our algorithm learns the mapping relationship of real and imaginary components between blurry input and sharp output using multi-scale cubic-mixer. 
Local feature extraction, slicing strategie and multi-scale scheme help us to improve the deblurring effect.
Quantitative and qualitative results show that the proposed algorithm performs favorably against the state-of-the-art deblurring methods in terms of accuracy and inference speed, and can generate a pleasing-visually image on real-world UHD images in real time.
%
%\clearpage

% References
\bibliography{uai2023-template}

\begin{thebibliography}{69}
\providecommand{\natexlab}[1]{#1}
\providecommand{\url}[1]{\texttt{#1}}
\expandafter\ifx\csname urlstyle\endcsname\relax
  \providecommand{\doi}[1]{doi: #1}\else
  \providecommand{\doi}{doi: \begingroup \urlstyle{rm}\Url}\fi

\bibitem[Barron and Poole(2016)]{BarronP16}
Jonathan~T. Barron and Ben Poole.
\newblock The fast bilateral solver.
\newblock In \emph{ECCV}, 2016.

\bibitem[Barron et~al.(2015)Barron, Adams, Shih, and
  Hern{\'{a}}ndez]{BarronASH15}
Jonathan~T. Barron, Andrew Adams, YiChang Shih, and Carlos Hern{\'{a}}ndez.
\newblock Fast bilateral-space stereo for synthetic defocus.
\newblock In \emph{CVPR}, 2015.

\bibitem[Chakrabarti(2016)]{Chakrabarti16}
Ayan Chakrabarti.
\newblock A neural approach to blind motion deblurring.
\newblock In \emph{ECCV}, 2016.

\bibitem[Chen et~al.(2021{\natexlab{a}})Chen, Fan, and Panda]{chen2021crossvit}
Chun-Fu~Richard Chen, Quanfu Fan, and Rameswar Panda.
\newblock Crossvit: Cross-attention multi-scale vision transformer for image
  classification.
\newblock In \emph{CVPR}, 2021{\natexlab{a}}.

\bibitem[Chen et~al.(2021{\natexlab{b}})Chen, Wang, Guo, Xu, Deng, Liu, Ma, Xu,
  Xu, and Gao]{chen2021pre}
Hanting Chen, Yunhe Wang, Tianyu Guo, Chang Xu, Yiping Deng, Zhenhua Liu, Siwei
  Ma, Chunjing Xu, Chao Xu, and Wen Gao.
\newblock Pre-trained image processing transformer.
\newblock In \emph{CVPR}, 2021{\natexlab{b}}.

\bibitem[Chen et~al.(2016)Chen, Adams, Wadhwa, and Hasinoff]{Adams16a}
Jiawen Chen, Andrew Adams, Neal Wadhwa, and Samuel~W. Hasinoff.
\newblock Bilateral guided upsampling.
\newblock \emph{ACM TOG}, 35\penalty0 (6):\penalty0 203:1--203:8, 2016.

\bibitem[Chen et~al.(2019)Chen, Fang, Wang, and Zhang]{ChenFWZ19}
Liang Chen, Faming Fang, Tingting Wang, and Guixu Zhang.
\newblock Blind image deblurring with local maximum gradient prior.
\newblock In \emph{CVPR}, 2019.

\bibitem[Chen et~al.(2022)Chen, Chu, Zhang, and Sun]{chen2022simple}
Liangyu Chen, Xiaojie Chu, Xiangyu Zhang, and Jian Sun.
\newblock Simple baselines for image restoration.
\newblock \emph{arXiv preprint arXiv:2204.04676}, 2022.

\bibitem[Chen et~al.(2020)Chen, Radford, Child, Wu, Jun, Luan, and
  Sutskever]{ChenRC0JLS20}
Mark Chen, Alec Radford, Rewon Child, Jeffrey Wu, Heewoo Jun, David Luan, and
  Ilya Sutskever.
\newblock Generative pretraining from pixels.
\newblock In \emph{ICML}, 2020.

\bibitem[Chen et~al.(2017)Chen, Xu, and Koltun]{ChenXK17}
Qifeng Chen, Jia Xu, and Vladlen Koltun.
\newblock Fast image processing with fully-convolutional networks.
\newblock In \emph{ICCV}, 2017.

\bibitem[Delbracio and Sapiro(2015{\natexlab{a}})]{DelbracioS15}
Mauricio Delbracio and Guillermo Sapiro.
\newblock Hand-held video deblurring via efficient fourier aggregation.
\newblock \emph{{IEEE} Trans. Computational Imaging}, 1\penalty0 (4):\penalty0
  270--283, 2015{\natexlab{a}}.

\bibitem[Delbracio and Sapiro(2015{\natexlab{b}})]{DelbracioS15_cvpr}
Mauricio Delbracio and Guillermo Sapiro.
\newblock Burst deblurring: Removing camera shake through fourier burst
  accumulation.
\newblock In \emph{CVPR}, 2015{\natexlab{b}}.

\bibitem[Deng et~al.(2021)Deng, Ren, Yan, Wang, Song, and Cao]{deng2021multi}
Senyou Deng, Wenqi Ren, Yanyang Yan, Tao Wang, Fenglong Song, and Xiaochun Cao.
\newblock Multi-scale separable network for ultra-high-definition video
  deblurring.
\newblock In \emph{ICCV}, 2021.

\bibitem[Dong et~al.(2011)Dong, Zhang, Shi, and Wu]{DongZSW11}
Weisheng Dong, Lei Zhang, Guangming Shi, and Xiaolin Wu.
\newblock Image deblurring and super-resolution by adaptive sparse domain
  selection and adaptive regularization.
\newblock \emph{IEEE TIP}, 20\penalty0 (7):\penalty0 1838--1857, 2011.

\bibitem[Dosovitskiy et~al.(2021)Dosovitskiy, Beyer, Kolesnikov, Weissenborn,
  Zhai, Unterthiner, Dehghani, Minderer, Heigold, Gelly, Uszkoreit, and
  Houlsby]{DosovitskiyICLR}
Alexey Dosovitskiy, Lucas Beyer, Alexander Kolesnikov, Dirk Weissenborn,
  Xiaohua Zhai, Thomas Unterthiner, Mostafa Dehghani, Matthias Minderer, Georg
  Heigold, Sylvain Gelly, Jakob Uszkoreit, and Neil Houlsby.
\newblock An image is worth 16x16 words: Transformers for image recognition at
  scale.
\newblock In \emph{ICLR}, 2021.

\bibitem[Gao et~al.(2019)Gao, Tao, Shen, and Jia]{GaoTSJ19}
Hongyun Gao, Xin Tao, Xiaoyong Shen, and Jiaya Jia.
\newblock Dynamic scene deblurring with parameter selective sharing and nested
  skip connections.
\newblock In \emph{CVPR}, 2019.

\bibitem[Gharbi et~al.(2017)Gharbi, Chen, Barron, Hasinoff, and
  Durand]{GharbiCBHD17}
Micha{\"{e}}l Gharbi, Jiawen Chen, Jonathan~T. Barron, Samuel~W. Hasinoff, and
  Fr{\'{e}}do Durand.
\newblock Deep bilateral learning for real-time image enhancement.
\newblock \emph{ACM TOG}, 36\penalty0 (4):\penalty0 118:1--118:12, 2017.

\bibitem[Ghiglia and Pritt(1998)]{1998Two}
D.~C. Ghiglia and MD~Pritt.
\newblock \emph{Two-Dimensional Phase Unwrapping: Theory, Algorithms, and
  Software}.
\newblock Two-Dimensional Phase Unwrapping: Theory, Algorithms, and Software,
  1998.

\bibitem[Gong et~al.(2017)Gong, Yang, Liu, Zhang, Reid, Shen, van~den Hengel,
  and Shi]{gong2017motion}
Dong Gong, Jie Yang, Lingqiao Liu, Yanning Zhang, Ian~D. Reid, Chunhua Shen,
  Anton van~den Hengel, and Qinfeng Shi.
\newblock From motion blur to motion flow: {A} deep learning solution for
  removing heterogeneous motion blur.
\newblock In \emph{CVPR}, 2017.

\bibitem[He et~al.(2020)He, Wang, Shi, and Duan]{HeWSD20}
Bin He, Ce~Wang, Boxin Shi, and Ling{-}Yu Duan.
\newblock Fhde\({}^{2}\)net: Full high definition demoireing network.
\newblock In \emph{ECCV}, 2020.

\bibitem[Jaesung et~al.(2020)Jaesung, Haeyun, Jucheol, and
  Sunghyun]{rim_2020_ECCV}
Rim Jaesung, Lee Haeyun, Won Jucheol, and Cho Sunghyun.
\newblock Real-world blur dataset for learning and benchmarking deblurring
  algorithms.
\newblock In \emph{ECCV}, 2020.

\bibitem[Johnson et~al.(2016)Johnson, Alahi, and Li]{JohnsonAF16}
Justin Johnson, Alexandre Alahi, and Fei{-}Fei Li.
\newblock Perceptual losses for real-time style transfer and super-resolution.
\newblock In \emph{ECCV}, 2016.

\bibitem[Kaufman and Fattal(2020)]{KaufmanF20}
Adam Kaufman and Raanan Fattal.
\newblock Deblurring using analysis-synthesis networks pair.
\newblock In \emph{CVPR}, 2020.

\bibitem[Ke et~al.(2021)Ke, Wang, Wang, Milanfar, and Yang]{ke2021musiq}
Junjie Ke, Qifei Wang, Yilin Wang, Peyman Milanfar, and Feng Yang.
\newblock Musiq: Multi-scale image quality transformer.
\newblock In \emph{ICCV}, 2021.

\bibitem[Kim et~al.(2017)Kim, Lee, Sch{\"{o}}lkopf, and Hirsch]{KimLSH17}
Tae~Hyun Kim, Kyoung~Mu Lee, Bernhard Sch{\"{o}}lkopf, and Michael Hirsch.
\newblock Online video deblurring via dynamic temporal blending network.
\newblock In \emph{ICCV}, 2017.

\bibitem[Kolesnikov et~al.(2020)Kolesnikov, Beyer, Zhai, Puigcerver, Yung,
  Gelly, and Houlsby]{KolesnikovBZPYG20}
Alexander Kolesnikov, Lucas Beyer, Xiaohua Zhai, Joan Puigcerver, Jessica Yung,
  Sylvain Gelly, and Neil Houlsby.
\newblock Big transfer (bit): General visual representation learning.
\newblock In \emph{ECCV}, 2020.

\bibitem[Krishnan et~al.(2011)Krishnan, Tay, and Fergus]{KrishnanTF11}
Dilip Krishnan, Terence Tay, and Rob Fergus.
\newblock Blind deconvolution using a normalized sparsity measure.
\newblock In \emph{CVPR}, 2011.

\bibitem[Kupyn et~al.(2018)Kupyn, Budzan, Mykhailych, Mishkin, and
  Matas]{KupynBMMM18}
Orest Kupyn, Volodymyr Budzan, Mykola Mykhailych, Dmytro Mishkin, and Jiri
  Matas.
\newblock Deblurgan: Blind motion deblurring using conditional adversarial
  networks.
\newblock In \emph{CVPR}, 2018.

\bibitem[Kupyn et~al.(2019)Kupyn, Martyniuk, Wu, and Wang]{KupynMWW19}
Orest Kupyn, Tetiana Martyniuk, Junru Wu, and Zhangyang Wang.
\newblock Deblurgan-v2: Deblurring (orders-of-magnitude) faster and better.
\newblock In \emph{ICCV}, 2019.

\bibitem[Lee et~al.(2021)Lee, Son, Rim, Cho, and Lee]{Lee2021Iterative}
Junyong Lee, Hyeongseok Son, Jaesung Rim, Sunghyun Cho, and Seungyong Lee.
\newblock Iterative filter adaptive network for single image defocus
  deblurring.
\newblock In \emph{CVPR}, 2021.

\bibitem[Levin et~al.(2009)Levin, Weiss, Durand, and Freeman]{LevinWDF09}
Anat Levin, Yair Weiss, Fr{\'{e}}do Durand, and William~T. Freeman.
\newblock Understanding and evaluating blind deconvolution algorithms.
\newblock In \emph{CVPR}, 2009.

\bibitem[Li et~al.(2021)Li, Lin, and Qu]{li2021infrared}
Guofa Li, Yongjie Lin, and Xingda Qu.
\newblock An infrared and visible image fusion method based on multi-scale
  transformation and norm optimization.
\newblock \emph{Information Fusion}, 71:\penalty0 109--129, 2021.

\bibitem[Li et~al.(2018)Li, Pan, Lai, Gao, Sang, and Yang]{LiPLGS018}
Lerenhan Li, Jinshan Pan, Wei{-}Sheng Lai, Changxin Gao, Nong Sang, and
  Ming{-}Hsuan Yang.
\newblock Learning a discriminative prior for blind image deblurring.
\newblock In \emph{CVPR}, 2018.

\bibitem[Liang et~al.(2021)Liang, Cao, Sun, Zhang, Van~Gool, and
  Timofte]{liang2021swinir}
Jingyun Liang, Jiezhang Cao, Guolei Sun, Kai Zhang, Luc Van~Gool, and Radu
  Timofte.
\newblock {SwinIR}: Image restoration using swin transformer.
\newblock In \emph{CVPR}, 2021.

\bibitem[McCloskey and Langer(2009)]{McCloskeyL09}
Scott McCloskey and Michael~S. Langer.
\newblock Planar orientation from blur gradients in a single image.
\newblock In \emph{CVPR}, 2009.

\bibitem[Michaeli and Irani(2014)]{MichaeliI14}
Tomer Michaeli and Michal Irani.
\newblock Blind deblurring using internal patch recurrence.
\newblock In \emph{ECCV}, 2014.

\bibitem[Nah et~al.(2017)Nah, Kim, and Lee]{NahKL17}
Seungjun Nah, Tae~Hyun Kim, and Kyoung~Mu Lee.
\newblock Deep multi-scale convolutional neural network for dynamic scene
  deblurring.
\newblock In \emph{CVPR}, 2017.

\bibitem[Nah et~al.(2019)Nah, Baik, Hong, Moon, Son, Timofte, and
  Lee]{NahBHMSTL19}
Seungjun Nah, Sungyong Baik, Seokil Hong, Gyeongsik Moon, Sanghyun Son, Radu
  Timofte, and Kyoung~Mu Lee.
\newblock {NTIRE} 2019 challenge on video deblurring and super-resolution:
  Dataset and study.
\newblock In \emph{CVPRW}, 2019.

\bibitem[Niklaus et~al.(2017)Niklaus, Mai, and Liu]{NiklausML17}
Simon Niklaus, Long Mai, and Feng Liu.
\newblock Video frame interpolation via adaptive convolution.
\newblock In \emph{CVPR}, 2017.

\bibitem[Perrone and Favaro(2014)]{PerroneF14}
Daniele Perrone and Paolo Favaro.
\newblock Total variation blind deconvolution: The devil is in the details.
\newblock In \emph{CVPR}, 2014.

\bibitem[Qin et~al.(2021)Qin, Li, Cao, Zhu, Zou, Li, Zhang, and
  Xue]{qin2021blind}
Fengqing Qin, Chaorong Li, Lilan Cao, Lihong Zhu, Xuyan Zou, Xiaomei Li, Tianqi
  Zhang, and Yilan Xue.
\newblock Blind image restoration with defocus blur by estimating point spread
  function in frequency domain.
\newblock In \emph{ICAIP}, 2021.

\bibitem[Simonyan and Zisserman(2015)]{SimonyanZ14a}
Karen Simonyan and Andrew Zisserman.
\newblock Very deep convolutional networks for large-scale image recognition.
\newblock In \emph{ICLR}, 2015.

\bibitem[Sui et~al.(2021)Sui, Afacan, Gholipour, and Warfield]{sui2021mri}
Yao Sui, Onur Afacan, Ali Gholipour, and Simon~K Warfield.
\newblock {MRI} super-resolution through generative degradation learning.
\newblock In \emph{MICCAI}, 2021.

\bibitem[Suin et~al.(2020)Suin, Purohit, and Rajagopalan]{SuinPR20}
Maitreya Suin, Kuldeep Purohit, and A.~N. Rajagopalan.
\newblock Spatially-attentive patch-hierarchical network for adaptive motion
  deblurring.
\newblock In \emph{CVPR}, 2020.

\bibitem[Sun et~al.(2013)Sun, Cho, Wang, and Hays]{SunCWH13}
Libin Sun, Sunghyun Cho, Jue Wang, and James Hays.
\newblock Edge-based blur kernel estimation using patch priors.
\newblock In \emph{ICCP}, 2013.

\bibitem[Tang et~al.(2022)Tang, Han, Guo, Xu, Li, Xu, and Wang]{tang2021image}
Yehui Tang, Kai Han, Jianyuan Guo, Chang Xu, Yanxi Li, Chao Xu, and Yunhe Wang.
\newblock An image patch is a wave: {Phase}-aware vision {MLP}.
\newblock In \emph{CVPR}, 2022.

\bibitem[Tao et~al.(2018)Tao, Gao, Shen, Wang, and Jia]{TaoGSWJ18}
Xin Tao, Hongyun Gao, Xiaoyong Shen, Jue Wang, and Jiaya Jia.
\newblock Scale-recurrent network for deep image deblurring.
\newblock In \emph{CVPR}, 2018.

\bibitem[Tolstikhin et~al.(2021)Tolstikhin, Houlsby, Kolesnikov, Beyer, Zhai,
  Unterthiner, Yung, Keysers, Uszkoreit, Lucic, and
  Dosovitskiy]{TolstikhinCoRR}
Ilya Tolstikhin, Neil Houlsby, Alexander Kolesnikov, Lucas Beyer, Xiaohua Zhai,
  Thomas Unterthiner, Jessica Yung, Daniel Keysers, Jakob Uszkoreit, Mario
  Lucic, and Alexey Dosovitskiy.
\newblock {MLP}-mixer: An all-{MLP} architecture for vision.
\newblock \emph{CoRR}, 2021.

\bibitem[Tu et~al.(2022)Tu, Talebi, Zhang, Yang, Milanfar, Bovik, and
  Li]{tu2022maxim}
Zhengzhong Tu, Hossein Talebi, Han Zhang, Feng Yang, Peyman Milanfar, Alan
  Bovik, and Yinxiao Li.
\newblock Maxim: Multi-axis mlp for image processing.
\newblock In \emph{CVPR}, 2022.

\bibitem[Wang et~al.(2019)Wang, Zhang, Fu, Shen, Zheng, and Jia]{WangZFSZJ19}
Ruixing Wang, Qing Zhang, Chi{-}Wing Fu, Xiaoyong Shen, Wei{-}Shi Zheng, and
  Jiaya Jia.
\newblock Underexposed photo enhancement using deep illumination estimation.
\newblock In \emph{CVPR}, 2019.

\bibitem[Wang et~al.(2021{\natexlab{a}})Wang, Li, Peng, Ma, Wang, Song, and
  Yan]{wang2021real}
Tao Wang, Yong Li, Jingyang Peng, Yipeng Ma, Xian Wang, Fenglong Song, and
  Youliang Yan.
\newblock Real-time image enhancer via learnable spatial-aware {3D} lookup
  tables.
\newblock In \emph{ICCV}, 2021{\natexlab{a}}.

\bibitem[Wang et~al.(2021{\natexlab{b}})Wang, Cun, Bao, and
  Liu]{wang2021uformer}
Zhendong Wang, Xiaodong Cun, Jianmin Bao, and Jianzhuang Liu.
\newblock Uformer: A general {U}-shaped transformer for image restoration.
\newblock \emph{arXiv preprint arXiv:2106.03106}, 2021{\natexlab{b}}.

\bibitem[Wu et~al.(2020)Wu, Yu, Liu, Chandraker, and Wang]{Wu0LCW20}
Junru Wu, Xiang Yu, Ding Liu, Manmohan Chandraker, and Zhangyang Wang.
\newblock {DAVID:} {Dual}-attentional video deblurring.
\newblock In \emph{WACV}, 2020.

\bibitem[Xia et~al.(2020)Xia, Zhang, Xue, Sun, Fang, Kulis, and
  Chen]{XiaZXSFKC20}
Xide Xia, Meng Zhang, Tianfan Xue, Zheng Sun, Hui Fang, Brian Kulis, and Jiawen
  Chen.
\newblock Joint bilateral learning for real-time universal photorealistic style
  transfer.
\newblock In \emph{ECCV}, 2020.

\bibitem[Xintian et~al.(2021)Xintian, Yiming, Shen, Qingli, and
  Wang]{Residual_f}
dMao Xintian, Liu Yiming, Wei Shen, Li~Qingli, and Yan Wang.
\newblock Deep residual fourier transformation for single image deblurring.
\newblock In \emph{arXiv:2111.11745}, 2021.

\bibitem[Xu et~al.(2013)Xu, Zheng, and Jia]{XuZJ13}
Li~Xu, Shicheng Zheng, and Jiaya Jia.
\newblock Unnatural {L0} sparse representation for natural image deblurring.
\newblock In \emph{CVPR}, 2013.

\bibitem[Yang and Yamac(2022)]{9857414}
Dan Yang and Mehmet Yamac.
\newblock Motion aware double attention network for dynamic scene deblurring.
\newblock In \emph{2022 IEEE/CVF Conference on Computer Vision and Pattern
  Recognition Workshops (CVPRW)}, pages 1112--1122, 2022.

\bibitem[Yuan et~al.(2021)Yuan, Fu, Huang, Lin, Zhang, Chen, and
  Wang]{yuan2021hrformer}
Yuhui Yuan, Rao Fu, Lang Huang, Weihong Lin, Chao Zhang, Xilin Chen, and
  Jingdong Wang.
\newblock {HRFormer}: High-resolution transformer for dense prediction.
\newblock In \emph{NeurIPS}, 2021.

\bibitem[Zamir et~al.(2021)Zamir, Arora, Khan, Hayat, Khan, Yang, and
  Shao]{zamir2021multi}
Syed~Waqas Zamir, Aditya Arora, Salman Khan, Munawar Hayat, Fahad~Shahbaz Khan,
  Ming-Hsuan Yang, and Ling Shao.
\newblock Multi-stage progressive image restoration.
\newblock In \emph{CVPR}, 2021.

\bibitem[Zamir et~al.(2022)Zamir, Arora, Khan, Hayat, Khan, and
  Yang]{zamir2021restormer}
Syed~Waqas Zamir, Aditya Arora, Salman Khan, Munawar Hayat, Fahad~Shahbaz Khan,
  and Ming-Hsuan Yang.
\newblock Restormer: Efficient transformer for high-resolution image
  restoration.
\newblock In \emph{CVPR}, 2022.

\bibitem[Zeng et~al.(2020)Zeng, Cai, Li, Cao, and Zhang]{ZengPAMI}
Hui Zeng, Jianrui Cai, Lida Li, Zisheng Cao, and Lei Zhang.
\newblock Learning image-adaptive {3D} lookup tables for high performance photo
  enhancement in real-time.
\newblock \emph{IEEE TPAMI}, abs/2009.14468, 2020.

\bibitem[Zhang et~al.(2019{\natexlab{a}})Zhang, Dai, Li, and
  Koniusz]{ZhangDLK19}
Hongguang Zhang, Yuchao Dai, Hongdong Li, and Piotr Koniusz.
\newblock Deep stacked hierarchical multi-patch network for image deblurring.
\newblock In \emph{CVPR}, 2019{\natexlab{a}}.

\bibitem[Zhang et~al.(2018)Zhang, Pan, Ren, Song, Bao, Lau, and
  Yang]{ZhangPRSBL018}
Jiawei Zhang, Jinshan Pan, Jimmy S.~J. Ren, Yibing Song, Linchao Bao, Rynson
  W.~H. Lau, and Ming{-}Hsuan Yang.
\newblock Dynamic scene deblurring using spatially variant recurrent neural
  networks.
\newblock In \emph{CVPR}, 2018.

\bibitem[Zhang and Tao(2020)]{ZhangT20}
Jing Zhang and Dacheng Tao.
\newblock Famed-{Net}: {A} fast and accurate multi-scale end-to-end dehazing
  network.
\newblock \emph{IEEE TIP}, 29:\penalty0 72--84, 2020.

\bibitem[Zhang et~al.(2019{\natexlab{b}})Zhang, Luo, Zhong, Ma, Liu, and
  Li]{8449842}
Kaihao Zhang, Wenhan Luo, Yiran Zhong, Lin Ma, Wei Liu, and Hongdong Li.
\newblock Adversarial spatio-temporal learning for video deblurring.
\newblock \emph{IEEE Transactions on Image Processing}, 28\penalty0
  (1):\penalty0 291--301, 2019{\natexlab{b}}.

\bibitem[Zhang et~al.(2021)Zhang, Dai, Yang, Xiao, Yuan, Zhang, and
  Gao]{zhang2021multi}
Pengchuan Zhang, Xiyang Dai, Jianwei Yang, Bin Xiao, Lu~Yuan, Lei Zhang, and
  Jianfeng Gao.
\newblock Multi-scale vision longformer: A new vision transformer for
  high-resolution image encoding.
\newblock In \emph{CVPR}, 2021.

\bibitem[Zhao et~al.(2020)Zhao, Jia, and Koltun]{ZhaoJK20}
Hengshuang Zhao, Jiaya Jia, and Vladlen Koltun.
\newblock Exploring self-attention for image recognition.
\newblock In \emph{CVPR}, 2020.

\bibitem[Zheng et~al.(2021)Zheng, Ren, Cao, Wang, and Jia]{zheng2021ultra}
Zhuoran Zheng, Wenqi Ren, Xiaochun Cao, Tao Wang, and Xiuyi Jia.
\newblock Ultra-high-definition image {HDR} reconstruction via collaborative
  bilateral learning.
\newblock In \emph{ICCV}, 2021.

\bibitem[Zhu et~al.(2020)Zhu, Su, Lu, Li, Wang, and Dai]{ZhuCoRR}
Xizhou Zhu, Weijie Su, Lewei Lu, Bin Li, Xiaogang Wang, and Jifeng Dai.
\newblock Deformable {DETR:} {D}eformable transformers for end-to-end object
  detection.
\newblock \emph{CoRR}, 2020.

\end{thebibliography}
\end{document}